\documentclass[10pt,twocolumn,letterpaper]{article}

\usepackage[hyphens]{url}
\ifdefined\REVIEW
  \usepackage[review]{cvpr}
\else
  \usepackage[pagenumbers]{cvpr}
\fi
\usepackage{times}
\usepackage{helvet}
\usepackage{courier}
\usepackage{microtype}
\usepackage{graphicx}
\usepackage[numbers,sort&compress]{natbib}
\usepackage{booktabs}
\usepackage{multirow}
\usepackage{makecell}
\usepackage{array}
\usepackage{amsmath}
\usepackage{amsfonts}
\usepackage{bm}
\usepackage[table]{xcolor}
\usepackage{xspace}
\usepackage{pifont}
\usepackage{tcolorbox}

\tcbuselibrary{skins}
\pgfdeclarehorizontalshading{swrrainbow}{100bp}{%
  color(0bp)=(red!80!black); color(25bp)=(red!80!black); color(33bp)=(orange); color(41bp)=(yellow);
  color(50bp)=(green!70!black); color(58bp)=(cyan); color(66bp)=(blue); color(75bp)=(violet); color(100bp)=(violet)}
\newenvironment{swrkeypointrainbow}[1]{%
  \begin{tcolorbox}[enhanced,colback=white,colframe=black!65,boxrule=0.6pt,
    arc=2pt,outer arc=2pt,boxsep=0pt,left=6pt,right=6pt,top=6pt,bottom=6pt,
    before skip=5pt,after skip=5pt,fontupper=\small,
    interior code={\fill[white] (interior.south west) rectangle (interior.north east);
      \shade[shading=swrrainbow,opacity=0.13] (interior.south west) rectangle (interior.north east);}]
  \setlength{\parskip}{0pt}\setlength{\parindent}{0pt}%
  {\normalsize\bfseries #1}\par\nobreak\vspace{2pt}%
  \hrule height 0.35pt\relax\nobreak\vspace{3pt}\noindent
}{\end{tcolorbox}}
\newsavebox\swrtabbox
\newcommand{\swrrainbowbar}[1]{\tikz\shade[shading=swrrainbow] (0,0) rectangle (#1,0.9pt);}

\newcommand{\sref}[1]{\textcolor{cvprblue}{\ref{#1}}}
\makeatletter\newcommand\captionof[1]{\def\@captype{#1}\caption}\makeatother   

\newcommand{\method}{SWR\xspace}
\definecolor{keycol}{HTML}{D4600A}

\newcommand{\swr}{Spectral World Renderer\xspace}
\newcommand{\bench}{SWR-Bench\xspace}

\definecolor{winred}{RGB}{176,24,24}
\definecolor{secblue}{RGB}{21,84,161}
\definecolor{ourscol}{gray}{0.92}

\definecolor{nvgreen}{HTML}{76B900}
\definecolor{tiffany}{HTML}{A9B4EC}\colorlet{rkA}{tiffany}\colorlet{rkB}{tiffany!50}\colorlet{rkC}{tiffany!20}   

\renewcommand{\topfraction}{1.0}
\renewcommand{\dbltopfraction}{0.94}
\renewcommand{\textfraction}{0.0}
\renewcommand{\floatpagefraction}{0.80}

\renewcommand{\dblfloatpagefraction}{0.80}

\title{{\fontsize{13}{15.6}\selectfont Spectral World Renderer: Unified 3D Hyperspectral Unmixing}}

\author{Chia-Ming~Lee\textsuperscript{1,2}\hspace{.62em}Yu-Jou~Hsiao\textsuperscript{3}\hspace{.62em}Ming-Ching~Chang\textsuperscript{1}\hspace{.62em}Xin~Li\textsuperscript{1}\hspace{.62em}Yu-Lun~Liu\textsuperscript{2}\hspace{.62em}Chih-Chung~Hsu\textsuperscript{2,3}\\[1pt]
{\normalsize \textsuperscript{1}University at Albany, State University of New York}\\
{\normalsize \textsuperscript{2}National Yang Ming Chiao Tung University\hspace{.8em}\textsuperscript{3}National Cheng Kung University}\\[0pt]
{\normalsize\color{cvprblue}\url{https://ming053l.github.io/Spectral-World-Renderer/}}
}
\definecolor{cvprblue}{rgb}{0.21,0.49,0.74}
\usepackage[pagebackref,breaklinks,colorlinks,allcolors=cvprblue]{hyperref}
\def\paperID{2687}
\def\confName{CVPR}
\def\confYear{2027}

\makeatletter
\def\abstract{%
   \iftoggle{cvprpagenumbers}{}{\thispagestyle{empty}}%
   \centerline{\large\bf Abstract}%
   \vspace*{5pt}\noindent%
   \it\ignorespaces%
}

\makeatother

\makeatletter
\g@addto@macro\normalsize{\setlength\abovedisplayskip{4pt plus 1pt minus 1pt}\setlength\belowdisplayskip{4pt plus 1pt minus 1pt}\setlength\abovedisplayshortskip{1pt plus 1pt}\setlength\belowdisplayshortskip{2pt plus 1pt}}
\makeatother
\begin{document}
\ifdefined\REVIEW\makeatletter
\expandafter\gdef\csname r@sec:supp_bench\endcsname{{A}{1}{}{}{}}
\expandafter\gdef\csname r@sec:synthetic_benchmark\endcsname{{A}{1}{}{}{}}
\expandafter\gdef\csname r@sec:supp_worlds\endcsname{{A.1}{1}{}{}{}}
\expandafter\gdef\csname r@tab:swr_worlds\endcsname{{S1}{2}{}{}{}}
\expandafter\gdef\csname r@tab:supp_settings\endcsname{{S2}{2}{}{}{}}
\expandafter\gdef\csname r@sec:supp_protocol\endcsname{{A.2}{2}{}{}{}}
\expandafter\gdef\csname r@sec:supp_spectra\endcsname{{B}{3}{}{}{}}
\expandafter\gdef\csname r@sec:spectra\endcsname{{B}{3}{}{}{}}
\expandafter\gdef\csname r@sec:supp_impl\endcsname{{C}{3}{}{}{}}
\expandafter\gdef\csname r@sec:supp_impl_rgb\endcsname{{C.1}{3}{}{}{}}
\expandafter\gdef\csname r@sec:supp_impl_clusters\endcsname{{C.2}{3}{}{}{}}
\expandafter\gdef\csname r@sec:supp_ident\endcsname{{D}{3}{}{}{}}
\expandafter\gdef\csname r@sec:supp_rare\endcsname{{D.1}{3}{}{}{}}
\expandafter\gdef\csname r@sec:capacity_exploration\endcsname{{D.2}{3}{}{}{}}
\expandafter\gdef\csname r@tab:spectra_worlds\endcsname{{S3}{4}{}{}{}}
\expandafter\gdef\csname r@tab:spectra_realism\endcsname{{S4}{4}{}{}{}}
\expandafter\gdef\csname r@fig:rare_classes\endcsname{{S1}{4}{}{}{}}
\expandafter\gdef\csname r@sec:supp_decoder\endcsname{{D.3}{4}{}{}{}}
\expandafter\gdef\csname r@sec:supp_analysis\endcsname{{D.3}{4}{}{}{}}
\expandafter\gdef\csname r@tab:capacity_exploration\endcsname{{S5}{5}{}{}{}}
\expandafter\gdef\csname r@sec:supp_real\endcsname{{E}{5}{}{}{}}
\expandafter\gdef\csname r@sec:s_real\endcsname{{E}{5}{}{}{}}
\expandafter\gdef\csname r@tab:supp_real_datasets\endcsname{{S6}{5}{}{}{}}
\expandafter\gdef\csname r@sec:supp_eff\endcsname{{F}{5}{}{}{}}
\expandafter\gdef\csname r@sec:s_efficiency\endcsname{{F}{5}{}{}{}}
\expandafter\gdef\csname r@sec:supp_apps\endcsname{{G}{5}{}{}{}}
\expandafter\gdef\csname r@fig:nht_features\endcsname{{S2}{6}{}{}{}}
\expandafter\gdef\csname r@tab:s_main_full\endcsname{{S7}{6}{}{}{}}
\expandafter\gdef\csname r@sec:supp_fpq\endcsname{{G.1}{6}{}{}{}}
\expandafter\gdef\csname r@fig:tetra_supp\endcsname{{S3}{7}{}{}{}}
\expandafter\gdef\csname r@fig:qualitative\endcsname{{S4}{8}{}{}{}}
\expandafter\gdef\csname r@fig:qualitative_nespof\endcsname{{S5}{9}{}{}{}}
\expandafter\gdef\csname r@fig:qualitative_apple\endcsname{{S6}{10}{}{}{}}
\expandafter\gdef\csname r@fig:supp_fpq\endcsname{{S7}{10}{}{}{}}
\expandafter\gdef\csname r@sec:app_sensor\endcsname{{G.2}{10}{}{}{}}
\expandafter\gdef\csname r@fig:app_sensor\endcsname{{S8}{11}{}{}{}}
\expandafter\gdef\csname r@fig:app_stats\endcsname{{S9}{11}{}{}{}}
\expandafter\gdef\csname r@sec:app_stats\endcsname{{G.3}{11}{}{}{}}
\expandafter\gdef\csname r@sec:app_anomaly\endcsname{{G.4}{12}{}{}{}}
\expandafter\gdef\csname r@sec:app_edit\endcsname{{G.5}{12}{}{}{}}
\expandafter\gdef\csname r@sec:supp_gallery\endcsname{{H}{12}{}{}{}}
\expandafter\gdef\csname r@fig:app_anomaly\endcsname{{S10}{13}{}{}{}}
\expandafter\gdef\csname r@fig:app_edit\endcsname{{S11}{13}{}{}{}}
\expandafter\gdef\csname r@fig:swr_hsi_1\endcsname{{S12}{14}{}{}{}}
\expandafter\gdef\csname r@fig:swr_hsi_2\endcsname{{S13}{15}{}{}{}}
\expandafter\gdef\csname r@fig:swr_multiview_1\endcsname{{S14}{16}{}{}{}}
\expandafter\gdef\csname r@fig:swr_multiview_2\endcsname{{S15}{17}{}{}{}}
\expandafter\gdef\csname r@fig:swr_multiview_3\endcsname{{S16}{18}{}{}{}}
\expandafter\gdef\csname r@fig:swr_multiview_4\endcsname{{S17}{19}{}{}{}}
\makeatother\fi
\twocolumn[{%
\renewcommand\twocolumn[1][]{#1}%
\maketitle
\vspace{-11.5mm}
\begin{center}
    \vspace{-2mm}\includegraphics[width=\textwidth]{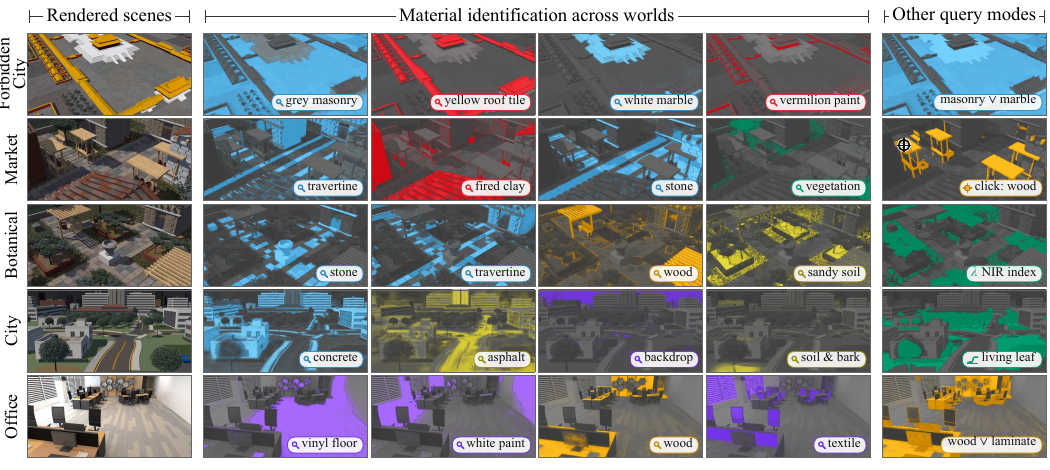}
    \vspace{-8mm}
        \captionof{figure}{\textbf{Material identification across \bench\ worlds.} \emph{Input:} multi-view hyperspectral images from generated worlds with known composition (left). \emph{Representation:} Gaussians with material abundances and spectra. \emph{Queries:} a material name mapped from labelled training views, a clicked pixel (crosshair), a reference spectrum, a near-infrared spectral index or a combination. Tiles show selections on test views.}
    \label{fig:teaser}
\end{center}
}]

\begin{abstract}
Hyperspectral images (HSI) preserve material evidence lost in RGB, yet 3D material identification from multi-view HSI faces two obstacles. Spectral Gaussian splatting reconstructs spectra accurately but leaves unmixing to per-pixel post-processing, while multi-view hyperspectral capture is costly and real datasets lack pixel-level material labels and sub-pixel proportions. We propose Spectral World Renderer (SWR), a material-aware explicit representation that unifies 3D spectral reconstruction and unmixing: it fits a Gaussian scaffold from RGB and decodes ray--Gaussian intersections into abundances over a shared endmember dictionary. Additionally, appearance guidance prevents spectrally similar materials from merging when intensity absorbs their brightness differences. Moreover, SWR requires no material-label supervision during training. To address the data gap, we further construct the SWR Benchmark (SWR-Bench): a coding-agent pipeline generates eight indoor and outdoor worlds from measured reflectance, aligning HSI with both annotations. SWR achieves the highest mIoU in all eight worlds and the highest mean PSNR on each real dataset. On an 11-class urban world, it trains at least $3.67\times$ faster than unmixing baselines; rendering averages 34.9 joint spectral and label views/s over four worlds on a single RTX~4090.\vspace{-1mm}
\end{abstract}

\setlength{\parskip}{1.5pt plus 0.3pt minus 0.2pt}
\par
\section{Introduction}\vspace{-1mm}
\begin{figure*}[t!]\vspace*{\dimexpr-\textfloatsep-3mm\relax}\end{figure*}

Robotic sorting, crop monitoring and building inspection require knowing the materials on a surface as the camera moves (Figure~\ref{fig:teaser}). Green paint and foliage, for example, can look alike in RGB despite having different spectral signatures. Foundation models recognize objects from RGB and support open-vocabulary 3D queries~\citep{CLIP,SAM,DINOv2,lerf}, but object identity does not establish material composition. RGB material labels~\citep{bell2015minc,upchurch2022dms,liang2022mcubes} likewise omit sub-pixel mixing.

\begin{swrkeypointrainbow}{Metamerism: Spectra Beyond RGB}
Different reflectance spectra can yield identical tristimulus values for a fixed illuminant and observer: metamerism~\citep{cie2020metamerism}. HSI retains distinctions lost in RGB integration.
\end{swrkeypointrainbow}

Hyperspectral imaging instead records many bands, preserving evidence about material reflectance~\citep{green1998aviris,ghamisi2017advances,kokaly2017usgs,ecostress}. Multi-view capture lifts it to 3D: views jointly constrain each surface point despite occlusion and shading, so materials persist as the camera moves. Spectral unmixing writes spectra as non-negative endmember mixtures~\citep{Bioucas2012B}. However, two obstacles hinder material identification from multi-view HSI. The first is data: capture requires costly calibration and registration~\citep{nespof,hsnerf2024}, and real datasets~\citep{nespof,hsnerf2024,hs3dnerf} lack pixel-level material labels and sub-pixel proportions (Table~\ref{tab:datasets}).

The second is methodological: spectral radiance fields and Gaussian splatting~\citep{hsnerf2024,spectralnerf,HyperGS,hgs,msgs,sinha2024spectralgaussians,mssplat_neural} primarily evaluate reconstruction fidelity and leave unmixing to a separate step. UnMix-NeRF~\citep{unmixnerf} integrates unmixing into a radiance field, but its implicit representation is costly to train and render for precise analysis. For deployment on edge devices, Gaussian splatting offers compact, fast-rendering and queryable primitives. Existing spectral Gaussian methods~\citep{3dgs,HyperGS,hgs} reconstruct multi-view spectra accurately but identify materials only through Vertex Component Analysis (VCA) and per-pixel Non-Negative Least Squares (NNLS) unmixing, which slows rendering~\citep{vca,lawson1974nnls}. Unmixing directly within Gaussian splatting therefore remains largely open for material-aware 3D analysis.

\begin{figure*}[t]
    \centering\vspace{-5.5mm}
    \includegraphics[width=\textwidth]{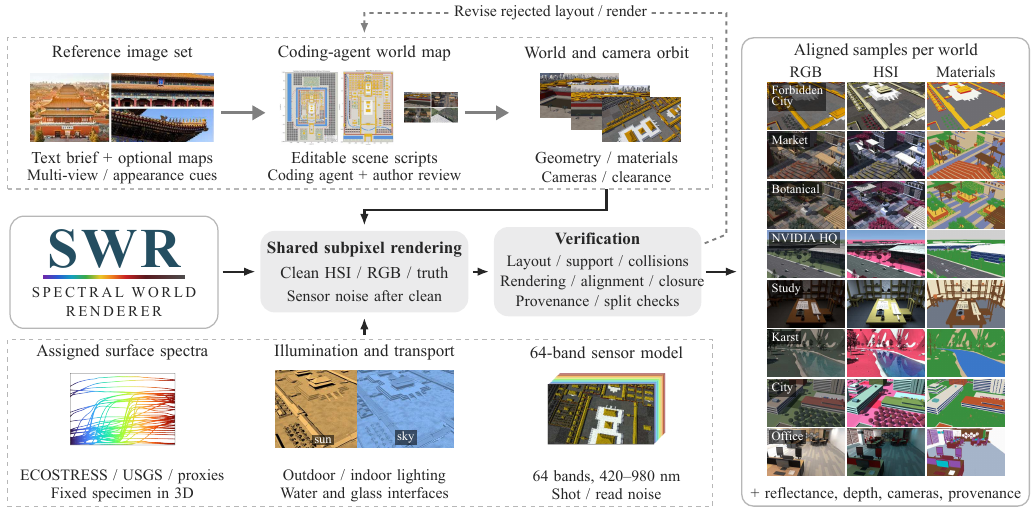}
    \vspace{-6.5mm}
    \caption{\textbf{Agent-driven reference-guided world generation pipeline.} A coding agent converts reference photographs and scene descriptions into executable geometry, materials and cameras; verification ensures physical plausibility. Measured spectra define material appearances. Shared sub-pixel samples yield aligned RGB and hyperspectral observations, material labels and proportions. Right: eight generated worlds.}
    \label{fig:data_pipeline}
    \vspace{-7mm}
\end{figure*}

\looseness=-1
To address these two obstacles, we first construct SWR-Bench for the missing multi-view HSI data: a carefully curated coding-agent pipeline generates eight physically plausible indoor and outdoor worlds, rendering hyperspectral images from measured reflectance with pixel-level material labels and sub-pixel proportions (Figure~\ref{fig:data_pipeline}). We then propose Spectral World Renderer (SWR). Inspired by Scaffold-GS~\citep{lu2024scaffoldgs}, SWR first fits a 3DGS scaffold from RGB, which admits recent Gaussian optimization methods~\citep{ren2025fastgs,condor2026nht} for quality, efficiency and storage. SWR then reconstructs HSI on this scaffold: features at ray intersections are decoded into abundances over a shared endmember dictionary. This unifies high-fidelity spectral reconstruction and fast, queryable material identification on the RGB scaffold. To further prevent spectrally similar materials from merging into one endmember as intensity absorbs their brightness differences, we introduce appearance-based guidance. SWR achieves the highest mIoU in all eight SWR-Bench worlds and the highest mean PSNR on each real dataset. Our main contributions are summarized as follows:
\begingroup
\setlength{\topsep}{1pt}
\setlength{\partopsep}{0pt}
\setlength{\leftmargini}{1em}
\begin{itemize}
\setlength{\itemsep}{0pt}
\setlength{\parsep}{0pt}
\setlength{\parskip}{0pt}
\item \textbf{Agentic benchmark with known composition.} An agentic pipeline builds \bench, indoor and outdoor worlds rendering multi-view HSI from measured spectra with aligned material labels and proportions.
\item \textbf{Material-aware explicit representation.} Compact Gaussian features predict local abundances at each ray--Gaussian intersection before alpha compositing. Shared visibility weights render spectra and material maps, keeping mixtures queryable on primitives across viewpoints.
\item \textbf{Appearance-guided material allocation.} Multi-view spectral-shape and brightness clusters guide abundance assignments without a material-label fitting loss, improving the separation of materials with similar spectral shapes.
\end{itemize}
\endgroup

\vspace{-4mm}\section{Related Work}\vspace{-2mm}
\begin{table}[t]
\centering
\footnotesize
\setlength{\tabcolsep}{1.15pt}
\caption{\textbf{Ground-truth coverage of multi-view datasets.} ID: material labels; Prop.: geometric area proportions; Gen.: on-demand world generation. \bench pairs images with shared-sample proportions and spectral variability.}
\label{tab:datasets}
\vspace{-2mm}
\sbox\swrtabbox{\begin{tabular}{lccccccc}
Dataset & Bands & nm & \makecell{Scenes\\$\times$views} & Capture & ID & Prop. & Gen. \\
\midrule
Replica~\cite{straub2019replica} & 3 & RGB & 18$\times$-- & indoor & \ding{55} & \ding{55} & \ding{55} \\
Hypersim~\cite{roberts2021hypersim} & 3 & RGB & 461$\times$168 & indoor & \ding{55} & \ding{55} & \ding{55} \\
NeSpoF~\cite{nespof} & 21 & 450--650 & 4$\times$18--30 & tabletop & \ding{55} & \ding{55} & \ding{55} \\
HS-NeRF~\cite{hsnerf2024} & 128/141 & 371--1087 & 8$\times$48--433 & turntable & \ding{55} & \ding{55} & \ding{55} \\
HSI-SC~\cite{hs3dnerf} & 204 & 397--1004 & 3$\times$64 & turntable & \ding{55} & \ding{55} & \ding{55} \\
\rowcolor{ourscol}\bench & 64 & 420--980 & 8$\times$96--240 & in/outdoor & \ding{51} & \ding{51} & \ding{51} \\
\end{tabular}}%
\swrrainbowbar{\wd\swrtabbox}\par\nointerlineskip\vspace{1pt}\usebox\swrtabbox\par\nointerlineskip\vspace{1pt}\swrrainbowbar{\wd\swrtabbox}
\vspace{-7mm}
\end{table}

\noindent\textbf{Hyperspectral 3D reconstruction.}
Radiance fields use neural or grid models \citep{nerf,mipnerf360,muller2022instantngp}; 3DGS composites Gaussians \citep{3dgs}, later improved in quality and speed \citep{yu2024mipsplatting,huang20242dgs,lu2024scaffoldgs,ren2025fastgs,hahlbohm2026fastergs}. In parallel, foundation-model features \citep{CLIP,SAM} answer object queries \citep{lerf,feature3dgs,langsplat,sam3dgs,ye2024gaussiangrouping} and select materials in RGB \citep{fischer2026sama}, and neural harmonic textures vary appearance within a primitive \citep{condor2026nht}. Hyperspectral fields render radiance per wavelength \citep{spectralnerf,hsnerf2024,nespof,hs3dnerf}; Gaussian versions store per-band quantities or compact latents decoded to spectra \citep{HyperGS,hgs,msgs,mssplat_neural,sinha2024spectralgaussians}. Complementary RGB inverse-rendering methods recover reflectance and illumination \citep{nerd,NeRFactor,gsir} or reflection components \citep{RefNeRF}. For material selection, \method\ decodes spatially varying features into per-intersection abundances and intensity on an explicit representation.

\noindent\textbf{Spectral unmixing.}
Unmixing explains each spectrum by endmembers and abundances under the abundance non-negativity (ANC) and sum-to-one (ASC) constraints \citep{974727,Bioucas2012B,FCLS}, found by geometric extraction \citep{winter1999nfindr,vca}, non-negative matrix factorization or autoencoders \citep{Lee1999NMF,BHUAE,8628241}, with spatial regularization \citep{Bioucas2012A} or spectral-variability and nonlinear models \citep{ELMM,heylen2014review}. These 2D abundance maps serve remote sensing and material analysis \citep{PLAZA2009S110,ghamisi2017advances}. In 3D, UnMix-NeRF integrates unmixing into a radiance field with a linear mixing model \citep{unmixnerf}. Our method likewise learns endmembers and abundances jointly, but uses explicit Gaussians with spatially varying features for material identification and selection.

\noindent\textbf{Programmatic world generation.}
Procedural rules \citep{raistrick2023infinigen,raistrick2024infinigenindoors,joshi2025infinigenarticulated}, procedural code editable by vision-language models \citep{raistrick2026procfunc}, language-model layouts such as LayoutGPT~\citep{feng2023layoutgpt} and Holodeck~\citep{yang2024holodeck} and scene code \citep{hu2024scenecraft,li2026rcwm} build 3D worlds; SAGE~\citep{xia2026sage} and SPREAD~\citep{li2026spread} further enforce physical validity with agentic simulation checks and physics-aware diffusion, producing scenes that are difficult to capture. These systems target RGB appearance. A hyperspectral world must also be spectrally realistic: materials need measured reflectance with within-class variation, illumination needs spectral irradiance, and observations need sensor band responses and noise. SWR builds executable worlds under these requirements and jointly renders spectra and material proportions to evaluate unmixing.

\par\medskip\noindent\begin{minipage}{\columnwidth}
    \centering
    \includegraphics[width=\columnwidth]{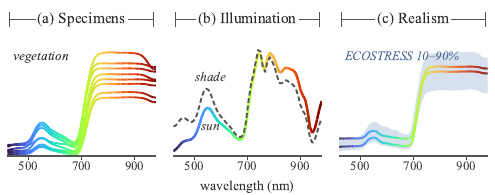}
    \vspace{-7mm}
    \captionof{figure}{\textbf{Material spectra vary with specimen and illumination.} (a) Measured grass specimens have different spectra. (b) Sunlight and shade change the spectral shape of the same grass. (c) Both broadleaf spectra lie within independent leaf measurements.}
    \label{fig:swr_spectra}
\end{minipage}\par\medskip

\noindent\textbf{Synthetic benchmarks and spectral simulation.}
Synthetic and scanned datasets supply geometry, semantics or intrinsic images \citep{straub2019replica,roberts2021hypersim,greff2022kubric}; game and agent benchmarks add G-buffers and checkable scene code \citep{huang2026gwr,gao2026threedcodebench,ye2026code4scene,shen2026fourdcodebench}. Spectral renderers and remote-sensing simulators trace light per wavelength \citep{pharr2016pbrt,wilkie2014hero,mitsuba3,goodenough2012dirsig}, and ray-traced vegetation scenes give known abundances for 2D unmixing \citep{dobigeon2014vegetation}. As Table~\ref{tab:datasets} summarizes, these datasets lack paired multi-view spectra and area proportions, which \bench samples from the same sub-pixel points while varying specimens and illumination.

\begin{figure*}[t]
    \centering\vspace{-5.5mm}
    \includegraphics[width=\textwidth]{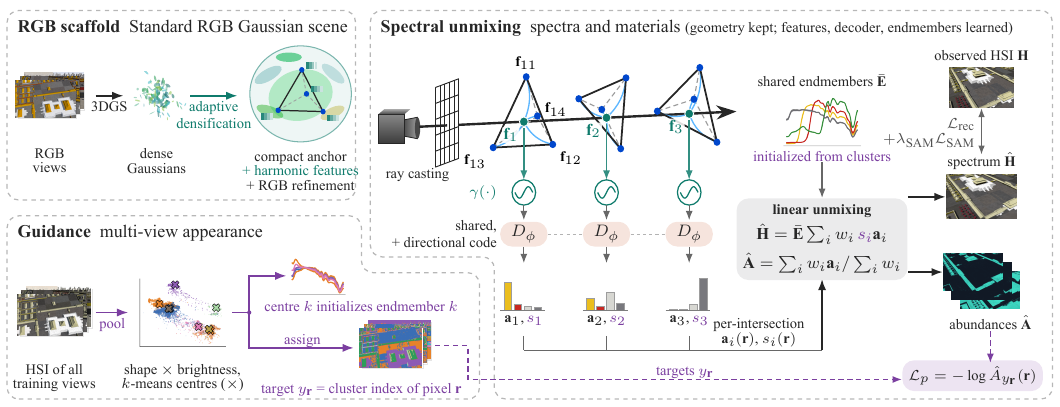}\vspace{-2mm}
    \caption{\textbf{SWR reconstruction and unmixing pipeline.} An RGB scaffold first fits geometry and harmonic features; with geometry frozen, spectral mixtures are then learned. A shared decoder predicts intersection abundances and intensity over scene-wide endmembers. Shared visibility weights jointly render spectra and abundances. Appearance clusters initialize the dictionary and regularize abundance assignments.}
    \label{fig:pipeline}\vspace{-6mm}
\end{figure*}

\vspace{-4.5mm}\section{Spectral World Renderer}
\label{sec:method}\vspace{-1mm}

\looseness=-1
We first build known-composition worlds for evaluation (Sec.~\ref{sec:swr}), then present local Gaussian unmixing (Sec.~\ref{sec:abundance_gs}) and appearance guidance (Sec.~\ref{sec:appearance_prior}). Fitting uses no material-label loss; validation labels select $K$, and labelled training views map components to classes for evaluation and support material identification with diverse queries.

\looseness=-1
A scene is observed from $V$ training cameras. Camera $v$ provides RGB $\mathbf I_v\in[0,1]^{H\times W\times3}$ and a co-registered hyperspectral cube $\mathbf H_v\in\mathbb R^{H\times W\times L}$ with $L$ bands at resolution $H\times W$. The ray $\mathbf r=\mathbf r_v(p)$ of pixel $p$ indexes its observed values $\mathbf I(\mathbf r)\in\mathbb R^3$ and $\mathbf H(\mathbf r)\in\mathbb R^L$. Novel views yield spectra $\hat{\mathbf H}(\mathbf r)$ and endmember abundances $\hat{\mathbf A}(\mathbf r)\in\mathbb R^K$ over a shared dictionary of $K$ spectra.

\input{sec/3_1_swr_bench}
\vspace{-2mm}\subsection{Pre-Compositing Unmixing}
\label{sec:abundance_gs}\vspace{-1mm}

A rendered pixel mixes materials at two stages: within a Gaussian, whose footprint can span several materials, and across Gaussians blended by alpha compositing. Unmixing the composited spectrum cannot attribute materials to individual Gaussians, so we decode a local mixture at each ray--Gaussian intersection before compositing, keeping per-primitive abundances for queries (Figure~\ref{fig:pipeline}).

\looseness=-1
\begin{figure*}[t!]\vspace*{\dimexpr-\textfloatsep-5mm\relax}\end{figure*}
\noindent\textbf{RGB anchor and harmonic features.}
Following the anchor-based design of Scaffold-GS~\citep{lu2024scaffoldgs}, we first fit and refine an RGB Gaussian scaffold, then learn spectra and materials with geometry fixed. Co-registered RGB views, supplied or derived from visible bands, let us directly leverage RGB geometry optimization without a dedicated spectral autoencoder~\citep{HyperGS}. FastGS densification and pruning~\citep{ren2025fastgs} retain primitives supported across views, reducing the features to fit and intersections to decode. The anchor has $N$ Gaussians $G_i$ with mean $\boldsymbol\mu_i$, rotation $\mathbf R_i$, scale $\boldsymbol\sigma_i$ and opacity $o_i$; after fitting, we drop colours and fold the rasterizer's screen-space variance into the scales. To retain local detail, each Gaussian carries a neural harmonic texture (NHT)~\citep{condor2026nht} with four feature vectors $\mathbf f_{im}\in\mathbb R^d$, $m=1,\ldots,4$, with $d=12$. A ray $\mathbf r$ queries $G_i$ at the point of maximal Gaussian response along the ray, whose tetrahedral barycentric coordinates $\beta_{im}(\mathbf r)$ interpolate the corner features:
\begin{equation}
 \mathbf f_i(\mathbf r)=\sum_{m=1}^{4}\beta_{im}(\mathbf r)\mathbf f_{im}\in\mathbb R^d.
 \label{eq:interp}
\end{equation}
Consequently, rays through one Gaussian can decode different mixtures across material boundaries. Before spectral fitting, we refine features and geometry with the NHT RGB shader, without further densification, and then freeze the geometry. Primitive storage is band-independent: 11 geometry/opacity values and 48 features, plus $LK$ values for the shared $K$-endmember dictionary. HyperGS~\citep{HyperGS} pretrains a spectral autoencoder and DD-HGS~\citep{hgs} adds per-band parameters and a diffusion denoiser; on real captures, SWR models are 6--16$\times$ and 10--31$\times$ smaller (Appendix~\sref{sec:supp_eff}).

\noindent\textbf{Pre-compositing material decoder.}
The interpolated feature from Eq.~\ref{eq:interp}, encoded as $\gamma(\mathbf f)=[\sin\mathbf f;\cos\mathbf f]\in\mathbb R^{2d}$~\citep{tancik2020fourier}, enters a shared MLP $D_\phi$ with a directional code $\mathbf Y(\mathbf d_i(\mathbf r))\in\mathbb R^9$: real spherical harmonics of degree $\le2$~\citep{ramamoorthi2001sh}, where $\mathbf d_i$ is the unit Gaussian-to-camera direction. 
\begin{equation}
 \mathbf z_i(\mathbf r)=D_\phi\!\left(\gamma(\mathbf f_i(\mathbf r)),\mathbf Y(\mathbf d_i(\mathbf r))\right)\in\mathbb R^{K+1}.
 \label{eq:abund}
\end{equation}
Superscripts index decoder outputs: the first $K$ entries yield abundances $\mathbf a_i(\mathbf r)=\operatorname{softmax}(\mathbf z_i^{1:K}(\mathbf r))$; the last gives intensity $s_i(\mathbf r)=\operatorname{softplus}(z_i^{K+1}(\mathbf r))$. The MLP has three width-128 SiLU hidden layers; a zero-initialized linear map adds the directional code to layer one.

Across intersections, these mixtures share a scene-wide dictionary $\bar{\mathbf E}=[\bar{\mathbf e}_1,\ldots,\bar{\mathbf e}_K]\in\mathbb R^{L\times K}$. The parameters $\boldsymbol\eta_k\in\mathbb R^L$ define each endmember through
\begin{equation}
 \bar{\mathbf e}_k=\frac{\operatorname{sigmoid}(\boldsymbol\eta_k)}{\|\operatorname{sigmoid}(\boldsymbol\eta_k)\|_2}\in\mathbb R^L_{\ge0}.
 \label{eq:endmember}
\end{equation}
With endmember scale fixed by normalization, $s_i$ scales the resulting radiance mixture~\citep{Bioucas2012B}:
\begin{equation}
 \mathbf h_i(\mathbf r)=s_i(\mathbf r)\bar{\mathbf E}\mathbf a_i(\mathbf r)\in\mathbb R^L_{\ge0}.
 \label{eq:hit}
\end{equation}

\looseness=-1
\noindent\textbf{Joint spectral and abundance rendering.}
We then composite both outputs with shared visibility weights. For Gaussians with ray-dependent opacity $\alpha_i(\mathbf r)$ (from $o_i$ and the Gaussian response), transmittance $T_i(\mathbf r)=\prod_{i'<i}(1-\alpha_{i'}(\mathbf r))$ gives visibility weights $w_i(\mathbf r)=T_i(\mathbf r)\alpha_i(\mathbf r)$~\citep{3dgs}:
\begin{equation}
 \hat{\mathbf H}(\mathbf r)=\sum_iw_i(\mathbf r)\mathbf h_i(\mathbf r),\quad
 \hat{\mathbf A}(\mathbf r)=\frac{\sum_iw_i(\mathbf r)\mathbf a_i(\mathbf r)}{\sum_iw_i(\mathbf r)}.
 \label{eq:composite}
\end{equation}
Both $\mathbf a_i$ and $\hat{\mathbf A}$ satisfy ANC and ASC on non-empty rays~\citep{FCLS}. However, $\hat{\mathbf H}=\bar{\mathbf E}\sum_iw_is_i\mathbf a_i$, while $\hat{\mathbf A}$ uses visibility weights alone, excluding $s_i$. Brightness thus raises radiance but not abundance, though these proportions need not equal area fractions. Averaging per-intersection abundances over training rays enables the primitive queries in Sec.~\ref{sec:application}. Unlike the volumetric abundances of UnMix-NeRF~\citep{unmixnerf}, these mixtures stay on explicit primitives, and SWR identifies materials more accurately (0.58 vs.\ 0.16 mean mIoU, Table~\ref{tab:bench}).

\vspace{-2mm}\subsection{Multi-View Appearance Guidance}
\label{sec:appearance_prior}\vspace{-1.5mm}
\begin{table*}[t]
\centering\vspace{-5.5mm}
\caption{\textbf{Spectral reconstruction and material identification on \bench.} SWR leads in mIoU and endmember accuracy across all worlds. Rare: classes below 3\,\% of test pixels; Spectra: distinct measured specimens; EM: class-matched endmember angle in degrees.}\vspace{-2mm}
\label{tab:bench}
\footnotesize\setlength{\tabcolsep}{3.0pt}
\sbox\swrtabbox{\begin{tabular}{l cccc cccc cccc >{\columncolor{ourscol}}c>{\columncolor{ourscol}}c>{\columncolor{ourscol}}c>{\columncolor{ourscol}}c}
 & & & & & \multicolumn{4}{c}{UnMix-NeRF~\cite{unmixnerf}} & \multicolumn{4}{c}{3DGS~\cite{3dgs} + VCA / NNLS} & \multicolumn{4}{c}{\swr} \\
\cmidrule(lr){6-9}\cmidrule(lr){10-13}\cmidrule(lr){14-17}
Scene & Setting & Classes & Rare & Spectra & PSNR$\uparrow$ & SAM$\downarrow$ & mIoU$\uparrow$ & EM$\downarrow$ & PSNR$\uparrow$ & SAM$\downarrow$ & mIoU$\uparrow$ & EM$\downarrow$ & PSNR$\uparrow$ & SAM$\downarrow$ & mIoU$\uparrow$ & EM$\downarrow$ \\
\midrule
Forbidden City & outdoor & 9 & 3 & 47 & \cellcolor{rkB}26.13 & \cellcolor{rkC}7.37 & \cellcolor{rkB}0.19 & \cellcolor{rkB}27.4 & \cellcolor{rkC}25.03 & \cellcolor{rkB}5.05 & \cellcolor{rkC}0.17 & \cellcolor{rkC}30.7 & \cellcolor{rkA}28.37 & \cellcolor{rkA}4.45 & \cellcolor{rkA}0.55 & \cellcolor{rkA}4.4 \\
Market & outdoor & 6 & 0 & 29 & \cellcolor{rkB}31.87 & \cellcolor{rkC}3.61 & \cellcolor{rkC}0.13 & \cellcolor{rkC}30.9 & \cellcolor{rkC}30.02 & \cellcolor{rkA}1.85 & \cellcolor{rkB}0.24 & \cellcolor{rkB}26.9 & \cellcolor{rkA}33.63 & \cellcolor{rkB}1.95 & \cellcolor{rkA}0.78 & \cellcolor{rkA}2.4 \\
Botanical & outdoor & 6 & 0 & 29 & \cellcolor{rkB}30.24 & \cellcolor{rkC}5.99 & \cellcolor{rkC}0.10 & \cellcolor{rkC}38.8 & \cellcolor{rkC}29.57 & \cellcolor{rkB}3.12 & \cellcolor{rkB}0.22 & \cellcolor{rkB}27.1 & \cellcolor{rkA}32.88 & \cellcolor{rkA}2.97 & \cellcolor{rkA}0.71 & \cellcolor{rkA}2.4 \\
NVIDIA HQ & outdoor & 8 & 2 & 27 & \cellcolor{rkC}18.29 & \cellcolor{rkC}14.98 & \cellcolor{rkC}0.08 & \cellcolor{rkC}43.0 & \cellcolor{rkB}22.27 & \cellcolor{rkB}12.31 & \cellcolor{rkB}0.13 & \cellcolor{rkB}24.4 & \cellcolor{rkA}25.30 & \cellcolor{rkA}8.25 & \cellcolor{rkA}0.31 & \cellcolor{rkA}3.7 \\
Study & indoor & 5 & 2 & 5 & \cellcolor{rkA}33.78 & \cellcolor{rkC}2.81 & \cellcolor{rkC}0.20 & \cellcolor{rkC}22.9 & \cellcolor{rkC}32.53 & \cellcolor{rkA}1.52 & \cellcolor{rkB}0.30 & \cellcolor{rkB}14.9 & \cellcolor{rkB}33.14 & \cellcolor{rkB}2.01 & \cellcolor{rkA}0.52 & \cellcolor{rkA}0.9 \\
Karst & outdoor & 5 & 2 & 10 & \cellcolor{rkC}26.91 & \cellcolor{rkC}3.29 & \cellcolor{rkB}0.38 & \cellcolor{rkB}25.8 & \cellcolor{rkB}27.19 & \cellcolor{rkB}2.26 & \cellcolor{rkC}0.35 & \cellcolor{rkC}28.8 & \cellcolor{rkA}28.10 & \cellcolor{rkA}2.15 & \cellcolor{rkA}0.73 & \cellcolor{rkA}2.9 \\
City & outdoor & 11 & 5 & 56 & \cellcolor{rkB}27.00 & \cellcolor{rkC}4.40 & \cellcolor{rkC}0.10 & \cellcolor{rkC}30.4 & \cellcolor{rkC}25.61 & \cellcolor{rkB}4.21 & \cellcolor{rkB}0.21 & \cellcolor{rkB}29.1 & \cellcolor{rkA}28.18 & \cellcolor{rkA}3.53 & \cellcolor{rkA}0.52 & \cellcolor{rkA}3.1 \\
Office & indoor & 8 & 3 & 35 & \cellcolor{rkB}25.33 & \cellcolor{rkC}6.65 & \cellcolor{rkC}0.11 & \cellcolor{rkB}18.8 & \cellcolor{rkC}25.01 & \cellcolor{rkB}5.03 & \cellcolor{rkB}0.17 & \cellcolor{rkC}41.3 & \cellcolor{rkA}27.54 & \cellcolor{rkA}4.31 & \cellcolor{rkA}0.53 & \cellcolor{rkA}3.3 \\
\end{tabular}}%
\swrrainbowbar{\wd\swrtabbox}\par\nointerlineskip\vspace{1pt}\usebox\swrtabbox\par\nointerlineskip\vspace{1pt}\swrrainbowbar{\wd\swrtabbox}\vspace{-5mm}
\end{table*}

\begin{figure}[t]
\centering
\includegraphics[width=\columnwidth]{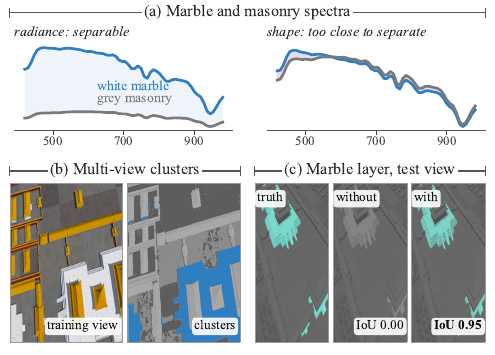}
\vspace{-6mm}
\caption{\textbf{Appearance guidance separates similar materials.} Forbidden City: (a) White marble is four times brighter than grey masonry but only $3.8^\circ$ apart in shape. (b) Shape-and-brightness clusters pool training views. (c) Cluster assignments keep the two materials distinct in the test view.}
\label{fig:appearance_prior}
\vspace{-5mm}
\end{figure}

\noindent\textbf{Assignment ambiguity.}
Unmixing after reconstruction, as in 3DGS with VCA and NNLS, depends heavily on endmember initialization: geometric extraction assumes pure pixels and is sensitive to noise~\citep{vca}, and dominant materials pull both endmembers and abundances, so small objects and rare classes are hard to recover. HSI unmixing counters this with SUnSAL-TV~\citep{Bioucas2012A}, ELMM~\citep{drumetz2016elmm} or MESMA~\citep{roberts1998mesma}, all within single 2D images. Learned dictionaries face the same risk, since reconstruction alone does not decide which endmember explains which surface. Since Eq.~\ref{eq:hit} scales each mixture by $s_i$, materials with similar shapes but different brightness, such as marble and masonry (Figure~\ref{fig:appearance_prior}(a)), can share one endmember~\citep{drumetz2016elmm,ELMM}; spectra stay accurate, so the reconstruction loss cannot detect the merge. We therefore introduce appearance guidance, an unsupervised prior from observed spectra, through cluster initialization and $\mathcal L_p$.

\looseness=-1
\noindent\textbf{Multi-view appearance clusters.}
Because one material spans several spectra across specimens and lighting (Figure~\ref{fig:swr_spectra}), we allow $K>C$~\citep{bateson2000bundles,roberts1998mesma,somers2011variability}; extra slots alone, however, are again absorbed by frequent materials. Clusters of observed appearance therefore allocate them: shape separates most materials, and brightness matters only for nearly parallel spectra. Without material labels, we pool training-view foreground pixels and describe each $3{\times}3$-smoothed spectrum $\tilde{\mathbf H}(\mathbf r)$ by its shape and brightness:
\begin{equation}
 \mathbf q_{\mathbf r}=\left[\frac{\tilde{\mathbf H}(\mathbf r)}{\|\tilde{\mathbf H}(\mathbf r)\|_2},\;\kappa\,b(\tilde{\mathbf H}(\mathbf r))\right]\in\mathbb R^{L+1},
 \label{eq:appearance_descriptor}
\end{equation}
where $b$ is the log norm of the spectrum, standardized robustly over the training views and clipped. The bounded weight $\kappa$, set per scene from the spread of shapes, keeps brightness secondary. Farthest-point seeding and spherical $k$-means group the descriptors into $K$ clusters; as in clustering-based self-supervision~\citep{caron2018deepcluster}, cluster indices provide pixel pseudo-labels $y_{\mathbf r}\in\{1,\ldots,K\}$, and cluster centres initialize the corresponding endmembers, so each appearance mode starts with its own endmember.

\looseness=-1
\noindent\textbf{Persistent assignment regularization.}
Initialization alone may drift back to an equally accurate merged solution, so, as in pseudo-label training~\citep{lee2013pseudo,sohn2020fixmatch}, we keep the clusters as targets, penalizing low rendered abundance of each pixel's assigned endmember:
\begin{equation}
 \mathcal L_p=-\frac{1}{|\mathcal B_p|}\sum_{\mathbf r\in\mathcal B_p}\log\hat A_{y_{\mathbf r}}(\mathbf r),
 \label{eq:assignment_prior}
\end{equation}
where $\hat A_{y_{\mathbf r}}(\mathbf r)$ is entry $y_{\mathbf r}$ of $\hat{\mathbf A}(\mathbf r)$ and $\mathcal B_p$ holds the pseudo-labelled pixels of the batch whose rays hit the scene. Because Gaussians are shared across views, they couple these pixel constraints, lifting 2D clusters into 3D assignments~\citep{zhi2021semanticnerf,siddiqui2023panoptic,ye2024gaussiangrouping}. Endmembers stay fixed for 3k updates while the weight decays to a non-zero floor (Sec.~\ref{sec:loss}), so guidance persists as spectra and intensities adapt.

\vspace{-2mm}\subsection{Optimization}
\label{sec:loss}\vspace{-1.5mm}
We first fit the anchor and refine geometry and features on RGB with $\mathcal L_{\mathrm{RGB}}=0.8\mathcal L_1+0.2\mathcal L_{\mathrm{D\text{-}SSIM}}$~\citep{3dgs}. With this initialization, we then freeze geometry and update features, decoder and endmembers using a uniform pixel batch $\mathcal B$ sampled within one view:
\begin{align}
 \mathcal L&=\mathcal L_{\mathrm{rec}}+\lambda_{\mathrm{SAM}}\mathcal L_{\mathrm{SAM}}+\lambda_p(t)\mathcal L_p,\nonumber\\
 \mathcal L_{\mathrm{rec}}&=\textstyle\frac{1}{|\mathcal B|L}\sum_{\mathbf r\in\mathcal B}\|\hat{\mathbf H}(\mathbf r)-\mathbf H(\mathbf r)\|_1,\nonumber\\
 \mathcal L_{\mathrm{SAM}}&=\textstyle\frac{1}{|\mathcal B_+|}\sum_{\mathbf r\in\mathcal B_+}\mathrm{SAM}(\hat{\mathbf H}(\mathbf r),\mathbf H(\mathbf r)),
 \label{eq:terms}
\end{align}
The reconstruction term fits band values, SAM penalizes spectral-shape error and $\mathcal L_p$ guides assignments. Material allocation is therefore explicitly optimized. Here $\mathcal B_+\subseteq\mathcal B$ contains rays with non-zero target spectra, and SAM is the spectral angle~\citep{kruse1993sips} in radians. We use $\lambda_{\mathrm{SAM}}=.05$ and, at update $t$, \mbox{$\lambda_p(t)=\max(.005,.05(1-t/3000))$}.

\flushbottom
\setlength{\floatsep}{12pt}
\setlength{\textfloatsep}{20pt}
\setlength{\dblfloatsep}{12pt}
\setlength{\dbltextfloatsep}{20pt}
\vspace{-2mm}\section{Experiments}
\label{sec:experiments}\vspace{-2mm}

\label{sec:setup}

\begin{figure*}[t]
    \centering
    \vspace{-5.5mm}\includegraphics[width=0.95\textwidth]{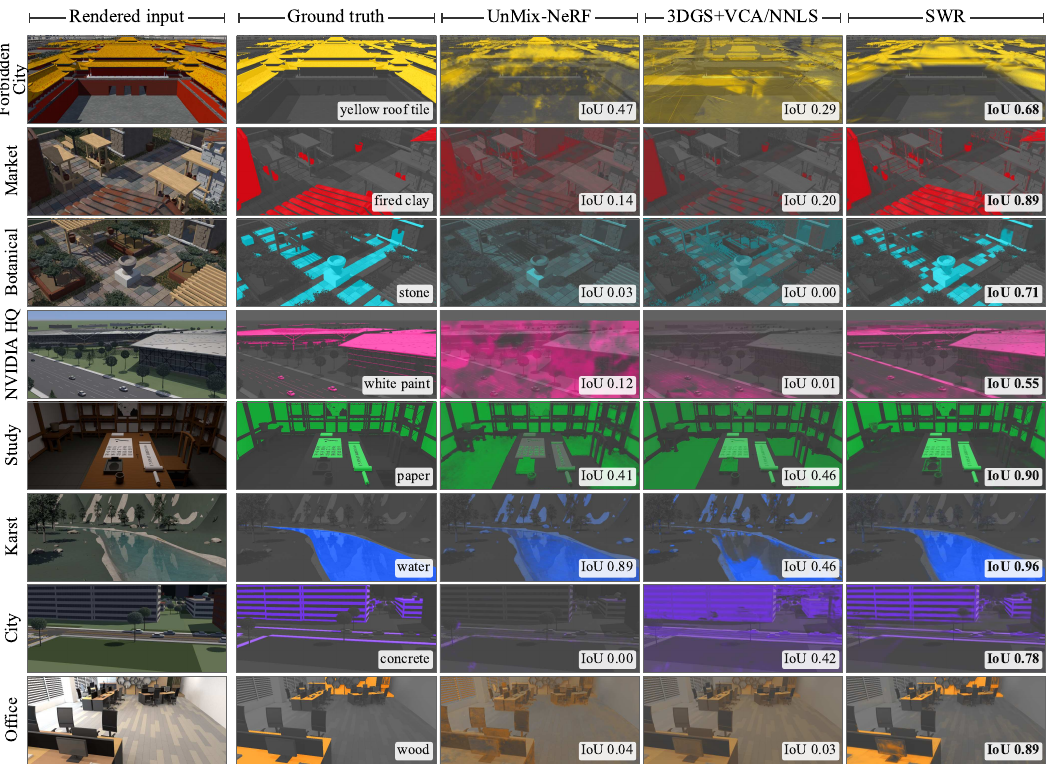}
    \vspace{0pt}
    \caption{\textbf{Material abundance maps across eight worlds.} Rows tint one material by its abundance. UnMix-NeRF integrates abundances predicted at volume samples; 3DGS+VCA/NNLS unmixes rendered spectra per pixel with VCA endmembers and NNLS; SWR composites abundances decoded at each ray--Gaussian intersection with the spectra's visibility weights. IoU thresholds class-summed abundance at $0.5$.}
    \label{fig:bench_qual}
    \vspace{-5mm}
\end{figure*}

We evaluate spectral fidelity and material identification separately, then isolate the design choices with controls.

\noindent\textbf{Data and baselines.} For material identification on \bench, we compare UnMix-NeRF \citep{unmixnerf} and an $L$-channel 3DGS \citep{3dgs} unmixed afterwards with VCA \citep{vca} endmembers from training-view renders and NNLS \citep{lawson1974nnls}. For spectral fidelity, we additionally compare HyperGS \citep{HyperGS} and DD-HGS \citep{hgs}. Real data span 21 to 204 bands: four multispectral NeSpoF scenes \citep{nespof}, the 204-band apple of HSI-SC-NeRF \citep{hs3dnerf} (HSI-SC apple) and five HS-NeRF scenes \citep{hsnerf2024}.

\looseness=-1
\noindent\textbf{Evaluation protocol.} Methods train on noisy $480{\times}270$ views; PSNR and the spectral angle SAM~\citep{kruse1993sips} (mean and 95th percentile, in degrees) are evaluated against clean $960{\times}540$ foreground ground truth. For mIoU, labelled training pixels map each endmember to the most frequent class among pixels it dominates; with this map frozen, each test pixel takes the class of its dominant endmember. Appendix~\sref{sec:supp_bench} defines the endmember angle error (EM) using minimum-total-angle matching between pure-pixel class means and endmembers, with equal weight for all material classes.

\begin{swrkeypointrainbow}{Balancing Reconstruction and Identification}
In Table~\ref{tab:bench} and Figure~\ref{fig:bench_qual}, the baselines reconstruct spectra competitively, even with the lowest SAM on Market and Study, yet reach only 0.08--0.38 mIoU with endmember errors of at least 14.9$^\circ$. Accurate spectra thus do not imply correct materials; SWR attains both, with 0.9--4.4$^\circ$ endmember error.
\end{swrkeypointrainbow}

\noindent\textbf{Implementation details.} The appendix lists the training settings of all methods in Table~\sref{tab:supp_settings}. Worlds use 64--180 training and 16--32 test views (Table~\sref{tab:swr_worlds}). Validation mIoU selects the endmember count $K$ per world from the class count, 16, 24 and 32: $K{=}24$ for Market, NVIDIA HQ and Study, $K{=}32$ elsewhere. Real captures use $K{=}16$; unmixing baselines on \bench\ use the known class count. 

\begin{table}[t]
\centering
\caption{\textbf{Spectral reconstruction on real datasets.} Two-run means.}
\label{tab:main}
\vspace{-1.5mm}
\begingroup
\fontsize{7.2}{8.4}\selectfont\setlength{\tabcolsep}{1pt}
\renewcommand{\arraystretch}{1.0}
\sbox\swrtabbox{\begin{tabular*}{\columnwidth}{@{\extracolsep{\fill}}l ccc ccc ccc@{}}
 & \multicolumn{3}{c}{NeSpoF} & \multicolumn{3}{c}{HSI-SC apple} & \multicolumn{3}{c}{HS-NeRF} \\
\cmidrule(lr){2-4}\cmidrule(lr){5-7}\cmidrule(l){8-10}
Method & PSNR & SAM & SAM$_{\mathrm{p95}}$ & PSNR & SAM & SAM$_{\mathrm{p95}}$ & PSNR & SAM & SAM$_{\mathrm{p95}}$ \\
\midrule
UnMix & 19.23 & 10.35 & 19.07 & 14.17 & 10.21 & 35.03 & 18.74 & 16.42 & 39.95 \\
3DGS & \cellcolor{rkC}27.48 & \cellcolor{rkB}4.38 & \cellcolor{rkB}10.01 & 23.45 & \cellcolor{rkC}2.67 & 10.99 & \cellcolor{rkB}27.48 & \cellcolor{rkA}6.24 & \cellcolor{rkA}15.84 \\
HyperGS & \cellcolor{rkB}27.83 & 4.88 & 10.23 & \cellcolor{rkC}25.10 & 2.69 & \cellcolor{rkC}8.42 & 27.34 & 7.71 & 19.59 \\
DD-HGS & 24.44 & \cellcolor{rkC}4.45 & \cellcolor{rkC}10.19 & \cellcolor{rkB}25.25 & \cellcolor{rkB}1.92 & \cellcolor{rkB}7.14 & \cellcolor{rkC}27.40 & \cellcolor{rkB}6.51 & \cellcolor{rkB}17.05 \\
\rowcolor{ourscol}SWR & \cellcolor{rkA}29.38 & \cellcolor{rkA}3.89 & \cellcolor{rkA}8.14 & \cellcolor{rkA}30.52 & \cellcolor{rkA}1.34 & \cellcolor{rkA}3.86 & \cellcolor{rkA}27.50 & \cellcolor{rkC}6.80 & \cellcolor{rkC}17.12 \\
\end{tabular*}}%
\swrrainbowbar{\wd\swrtabbox}\par\nointerlineskip\vspace{1pt}\usebox\swrtabbox\par\nointerlineskip\vspace{1pt}\swrrainbowbar{\wd\swrtabbox}
\endgroup
\vspace{-2mm}
\end{table}

\begin{table}[t]
\centering
\footnotesize\setlength{\tabcolsep}{3pt}
\caption{\textbf{Effect of RGB anchor refinement.}}
\label{tab:rgb_stage}
\vspace{-2mm}
\sbox\swrtabbox{\begin{tabular*}{\columnwidth}{@{\extracolsep{\fill}}l rrr rr@{}}
 & \multicolumn{3}{c}{Market} & \multicolumn{2}{c}{HSI-SC apple} \\
\cmidrule(lr){2-4}\cmidrule(l){5-6}
Variant & mIoU$\uparrow$ & PSNR$\uparrow$ & SAM$\downarrow$ & PSNR$\uparrow$ & SAM$\downarrow$ \\
\midrule
Anchor only & 0.66 & 30.73 & 3.20 & \cellcolor{rkC}28.44 & \cellcolor{rkC}1.67 \\
Geometry only & \cellcolor{rkC}0.70 & \cellcolor{rkC}31.55 & \cellcolor{rkC}2.76 & 27.42 & 1.91 \\
Features only & \cellcolor{rkB}0.77 & \cellcolor{rkB}33.16 & \cellcolor{rkB}2.04 & \cellcolor{rkA}31.32 & \cellcolor{rkA}1.29 \\
Refinement (full) & \cellcolor{rkA}0.78 & \cellcolor{rkA}33.63 & \cellcolor{rkA}1.95 & \cellcolor{rkB}30.99 & \cellcolor{rkB}1.34 \\
\end{tabular*}}%
\swrrainbowbar{\wd\swrtabbox}\par\nointerlineskip\vspace{1pt}\usebox\swrtabbox\par\nointerlineskip\vspace{1pt}\swrrainbowbar{\wd\swrtabbox}
\vspace{-4.5mm}
\end{table}

\begin{table}[t]
\centering
\caption{\textbf{Ablation of decoder components} at $K{=}C$ without guidance, two seeds. Mix: endmember mixture, Eq.~\ref{eq:hit} (\ding{55}: direct spectra); Hit: decoding before compositing (\ding{55}: after); Tet.: four corner features, Eq.~\ref{eq:interp} (\ding{55}: one); Enc.: encoding $\gamma$; Int.: intensity $s_i$ (\ding{55}: $s_i{=}1$). (a) full Spectral World Renderer, (h) NHT~\cite{condor2026nht}.}
\label{tab:nht_tradeoff}
\vspace{-2mm}
\begingroup
\fontsize{7.5}{8.5}\selectfont\setlength{\tabcolsep}{1.6pt}
\sbox\swrtabbox{\begin{tabular*}{\columnwidth}{@{\extracolsep{\fill}}c ccccc cc ccc@{}}
 & & & & & & \multicolumn{2}{c}{NeSpoF scene3} & \multicolumn{3}{c}{Market} \\
\cmidrule(lr){7-8}\cmidrule(l){9-11}
 & Mix & Hit & Tet. & Enc. & Int. & PSNR$\uparrow$ & SAM$\downarrow$ & PSNR$\uparrow$ & SAM$\downarrow$ & mIoU$\uparrow$ \\
\midrule
\rowcolor{ourscol}(a) & \ding{51} & \ding{51} & \ding{51} & \ding{51} & \ding{51} & \cellcolor{rkB}29.62 & \cellcolor{rkC}4.22 & 33.72 & 1.87 & \cellcolor{rkA}0.47 \\
(b) & \ding{51} & \ding{51} & \ding{55} & \ding{51} & \ding{51} & 29.24 & 4.30 & 32.09 & 2.08 & \cellcolor{rkC}0.34 \\
(c) & \ding{51} & \ding{51} & \ding{51} & \ding{55} & \ding{51} & \cellcolor{rkC}29.61 & \cellcolor{rkB}4.21 & 33.65 & 1.89 & 0.30 \\
(d) & \ding{51} & \ding{55} & \ding{51} & \ding{51} & \ding{51} & 29.16 & \cellcolor{rkA}4.17 & \cellcolor{rkB}34.37 & \cellcolor{rkB}1.57 & \cellcolor{rkB}0.37 \\
(e) & \ding{51} & \ding{51} & \ding{51} & \ding{51} & \ding{55} & 29.54 & 4.37 & 33.70 & 2.14 & 0.22 \\
(f) & \ding{55} & \ding{51} & \ding{51} & \ding{51} & -- & \cellcolor{rkA}29.69 & 4.28 & \cellcolor{rkC}33.74 & \cellcolor{rkC}1.72 & -- \\
(g) & \ding{55} & \ding{51} & \ding{55} & \ding{51} & -- & 29.26 & 4.34 & 32.11 & 1.90 & -- \\
(h) & \ding{55} & \ding{55} & \ding{51} & \ding{51} & -- & 29.21 & \cellcolor{rkA}4.17 & \cellcolor{rkA}34.38 & \cellcolor{rkA}1.41 & -- \\
\end{tabular*}}%
\swrrainbowbar{\wd\swrtabbox}\par\nointerlineskip\vspace{1pt}\usebox\swrtabbox\par\nointerlineskip\vspace{1pt}\swrrainbowbar{\wd\swrtabbox}
\endgroup
\vspace{-4mm}
\end{table}

\begin{figure}[t]
\centering\vspace{3mm}
\includegraphics[width=\columnwidth]{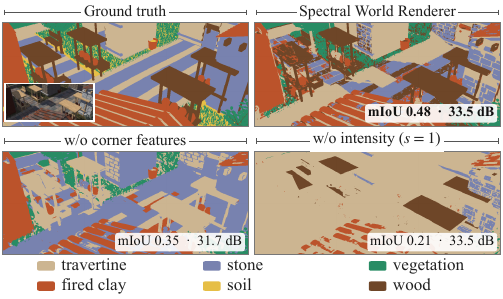}
\vspace{-7mm}
\caption{\textbf{Effects of decoder design.}}
\label{fig:feature_field}
\vspace{-2mm}
\end{figure}
\begin{table}[t]
\centering
\footnotesize\setlength{\tabcolsep}{2.3pt}
\caption{\textbf{Training and rendering cost.}}
\label{tab:representation}
\vspace{-2mm}
\sbox\swrtabbox{\begin{tabular}{@{}l c ccccc@{}}
 & City & \multicolumn{5}{c}{FPS$\uparrow$ at $960\times540\times64$} \\
\cmidrule(l){3-7}
Method & train$\downarrow$ & Market & Bot. & Study & Karst & Mean \\
\midrule
UnMix-NeRF & \cellcolor{rkB}24\,min 17\,s & 0.18 & 0.21 & 0.21 & 0.12 & 0.18 \\
3DGS+VCA/NNLS & 85\,min 2\,s & \cellcolor{rkC}2.53 & \cellcolor{rkC}2.51 & \cellcolor{rkC}4.70 & \cellcolor{rkC}4.73 & \cellcolor{rkC}3.62 \\
HyperGS+VCA/NNLS & \cellcolor{rkC}36\,min 50\,s & \cellcolor{rkB}2.82 & \cellcolor{rkB}2.85 & \cellcolor{rkB}5.59 & \cellcolor{rkB}5.62 & \cellcolor{rkB}4.22 \\
DD-HGS+VCA/NNLS & 5\,h 10\,min & 1.99 & 1.94 & 3.04 & 3.10 & 2.52 \\
\rowcolor{ourscol}\method & \cellcolor{rkA}6\,min 37\,s & \cellcolor{rkA}36.5 & \cellcolor{rkA}39.3 & \cellcolor{rkA}36.3 & \cellcolor{rkA}27.4 & \cellcolor{rkA}34.9 \\
\end{tabular}}%
\swrrainbowbar{\wd\swrtabbox}\par\nointerlineskip\vspace{1pt}\usebox\swrtabbox\par\nointerlineskip\vspace{1pt}\swrrainbowbar{\wd\swrtabbox}
\vspace{-9mm}
\end{table}

\vspace{-1.5mm}\subsection{Reconstruction and Material Identification}\vspace{-.5mm}
\label{sec:bench}

\looseness=-1
\noindent\textbf{Identification and dictionary capacity.} Worlds differ in difficulty: they contain 5--11 classes, up to five of them rare (Table~\ref{tab:bench}), so the dictionary size that covers their spectral variation also differs. With $K$ selected per world on validation views, SWR attains the highest mIoU and the lowest endmember error in every world, more than doubling the mean mIoU of the strongest baseline. UnMix-NeRF is competitive only on the five-class worlds and falls to 0.08--0.19 mIoU on worlds with more classes.

\looseness=-1
\noindent\textbf{Fidelity--identification gap.} Measured-specimen NNLS reaches 0.84--0.94 mIoU on clean radiance and 0.32--0.77 with noise (Appendix~\sref{sec:supp_bench}), separating observation ambiguity from learning errors.

\begin{figure}[t]
    \centering
    \includegraphics[width=\columnwidth]{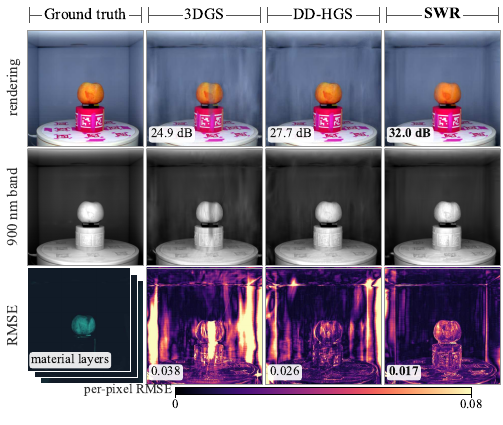}
    \vspace{-8.5mm}
    \caption{\textbf{Reconstruction of the 204-band HSI-SC apple.} Beyond RGB, the 900\,nm band records near-infrared reflectance that colour images cannot capture; SWR reconstructs it without the streaks of the baselines and with the lowest RMSE.}
    \label{fig:real}
    \vspace{-3.5mm}
\end{figure}

\par\medskip
\begin{figure}[t]
\centering
\includegraphics[width=\columnwidth]{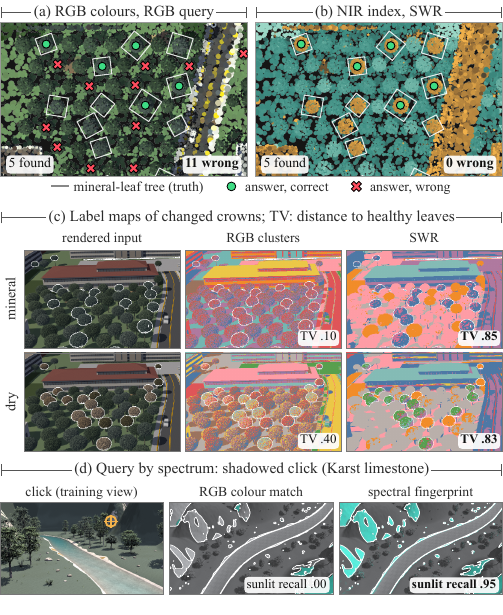}
\vspace{-8.5mm}
\caption{\textbf{Material queries beyond RGB.} City: boxed crowns use mineral leaves. (a,b) Spectral queries isolate them; RGB retrieves ordinary trees. (c) Mineral and dry versus healthy crowns. (d) With known irradiance, shadowed limestone retrieves sunlit Karst rock.}
\label{fig:application}
\vspace{-5mm}
\end{figure}

\vspace{-1.5mm}\subsection{Real Captures and Efficiency}\vspace{-.5mm}
\label{sec:real}
\noindent\textbf{Spectral reconstruction.} With band-independent features, SWR gains 1.55\,dB mean PSNR on NeSpoF and 5.27\,dB on the HSI-SC apple over the strongest baseline (Table~\ref{tab:main}; one view in Figure~\ref{fig:real}). HS-NeRF is nearly tied (27.50/27.48\,dB), with lower mean and tail SAM for 3DGS.

\noindent\textbf{Efficiency.} With spectra and abundances from one pass and no post-hoc NNLS, SWR trains on City at least 3.67$\times$ faster than the unmixing pipelines and renders at least 8$\times$ more views/s (Table~\ref{tab:representation}), so material inspection is interactive.

\vspace{-1.5mm}\subsection{Representation and Assignment Controls}\vspace{-.5mm}
\label{sec:decoder_diagnostics}
\noindent\textbf{RGB anchor refinement.} Refining on RGB before spectral fitting improves both fidelity and identification (Table~\ref{tab:rgb_stage}), mainly through the features; geometry alone helps little.

\noindent\textbf{Decoder components.} In Table~\ref{tab:nht_tradeoff}, fidelity and identification rely on different components. Spatially varying features help both, letting one Gaussian span a material boundary. The encoding and intensity barely change PSNR but keep mixtures from absorbing shading. Decoding after compositing or predicting spectra directly gains little fidelity yet loses per-primitive materials (Figure~\ref{fig:feature_field}).

\noindent\textbf{Appearance guidance and capacity.} The two are complementary: a dictionary larger than the class count covers within-class variation, while guidance assigns the extra endmembers to surfaces (Table~\sref{tab:capacity_exploration}).

\vspace{-1.5mm}\subsection{Multi-View Queries Beyond RGB}\vspace{-.5mm}
\label{sec:application}

\begin{swrkeypointrainbow}{Queryable Materials in 3D}
A clicked spectrum or a library spectrum queried by material name is matched to each Gaussian, and the explicit representation renders the selection quickly from any view, enabling material identification beyond RGB, even for approximate metamers.
\end{swrkeypointrainbow}

\looseness=-1
\noindent\textbf{From image labels to surface queries.} Unlike pixel-map classification and clustering~\citep{melgani2004svm,tarabalka2009clustering,chen2016deep,hong2022spectralformer,zhang2016sssc}, shared Gaussian features make spectral distinctions queryable from novel cameras. On shared candidates, RGB retrieves eleven ordinary trees and the mineral query none (Figure~\ref{fig:application}(a)).

\noindent\textbf{Queries under known illumination.} Shadow alters spectra, as shadowed surfaces see mainly skylight; whitening by the known sun/sky irradiance removes this effect: a shadowed click retrieves the same material in sunlight with 77\,\% mean recall over four worlds, against 10\,\% for RGB (Figure~\ref{fig:application}(d)).

\par
\setlength{\parskip}{4pt}
\section{Conclusion}
\label{sec:conclusion}
\looseness=-1
Our contributions are twofold. First, an agentic pipeline builds \bench, eight multi-view HSI worlds with composition labels, addressing the lack of annotated data. Second, \swr\ unifies spectral reconstruction and unmixing in one compact Gaussian representation that trains in minutes and renders interactively. It identifies materials best in all eight worlds, leads mean PSNR on every real dataset, and quickly retrieves materials that look alike in RGB, such as mineral and living leaves, advancing 3D material perception beyond RGB.

\ifdefined\REVIEW\else\section*{Acknowledgement}
This study was supported in part by the National Science and Technology Council (NSTC), Taiwan, under grants 115-2627-M-006-001, 115-2218-E-A49-023, and 115-2634-F-A49-014. We thank the National Center for High-performance Computing (NCHC) of National Applied Research Laboratories (NARLabs) in Taiwan for providing computational and storage resources.
\fi

\clearpage
\setlength{\parskip}{0pt}
\section*{LLM Usage Statement}
We used large language model assistants (Anthropic Claude and OpenAI Codex) in this work. They assisted in (i) drafting and editing the manuscript text, captions and tables, (ii) writing the plotting scripts that produce the figures, and (iii) writing and debugging training, evaluation and timing code and launching the scripted experiments. The authors directed the research, checked the reported results against the experiment outputs, and take full responsibility for the content of this paper.

\bibliographystyle{ieeenat_fullname}
\bibliography{main}

\begin{thebibliography}{102}
\providecommand{\natexlab}[1]{#1}
\providecommand{\url}[1]{\texttt{#1}}
\expandafter\ifx\csname urlstyle\endcsname\relax
  \providecommand{\doi}[1]{doi: #1}\else
  \providecommand{\doi}{doi: \begingroup \urlstyle{rm}\Url}\fi

\bibitem[Acito et~al.(2011)Acito, Diani, and Corsini]{acito2011signal}
Nicola Acito, Marco Diani, and Giovanni Corsini.
\newblock Signal-dependent noise modeling and model parameter estimation in
  hyperspectral images.
\newblock \emph{IEEE Transactions on Geoscience and Remote Sensing},
  49\penalty0 (8):\penalty0 2957--2971, 2011.

\bibitem[{Anthropic}(2024)]{anthropic2024mcp}
{Anthropic}.
\newblock Model context protocol.
\newblock \url{https://modelcontextprotocol.io}, 2024.

\bibitem[{ASTM International}(2003)]{astmg173}
{ASTM International}.
\newblock {ASTM G173-03}: Standard tables for reference solar spectral
  irradiances: Direct normal and hemispherical on 37$^\circ$ tilted surface,
  2003.

\bibitem[Barron et~al.(2022)Barron, Mildenhall, Verbin, Srinivasan, and
  Hedman]{mipnerf360}
Jonathan~T. Barron, Ben Mildenhall, Dor Verbin, Pratul~P. Srinivasan, and Peter
  Hedman.
\newblock {Mip-NeRF} 360: Unbounded anti-aliased neural radiance fields.
\newblock In \emph{CVPR}, pages 5470--5479, 2022.

\bibitem[Bateson et~al.(2000)Bateson, Asner, and Wessman]{bateson2000bundles}
C.~Ann Bateson, Gregory~P. Asner, and Carol~A. Wessman.
\newblock Endmember bundles: A new approach to incorporating endmember
  variability into spectral mixture analysis.
\newblock \emph{IEEE Transactions on Geoscience and Remote Sensing},
  38\penalty0 (2):\penalty0 1083--1094, 2000.

\bibitem[Bell et~al.(2015)Bell, Upchurch, Snavely, and Bala]{bell2015minc}
Sean Bell, Paul Upchurch, Noah Snavely, and Kavita Bala.
\newblock Material recognition in the wild with the {Materials in Context}
  database.
\newblock In \emph{CVPR}, pages 3479--3487, 2015.

\bibitem[Bioucas-Dias et~al.(2012)Bioucas-Dias, Plaza, Dobigeon, Parente, Du,
  Gader, and Chanussot]{Bioucas2012B}
José~M. Bioucas-Dias, Antonio Plaza, Nicolas Dobigeon, Mario Parente, Qian Du,
  Paul Gader, and Jocelyn Chanussot.
\newblock Hyperspectral unmixing overview: Geometrical, statistical, and sparse
  regression-based approaches.
\newblock \emph{IEEE Journal of Selected Topics in Applied Earth Observations
  and Remote Sensing}, 5\penalty0 (2):\penalty0 354--379, 2012.

\bibitem[Boss et~al.(2021)Boss, Braun, Jampani, Barron, Liu, and Lensch]{nerd}
Mark Boss, Raphael Braun, Varun Jampani, Jonathan~T. Barron, Ce Liu, and
  Hendrik~P.A. Lensch.
\newblock {NeRD}: Neural reflectance decomposition from image collections.
\newblock In \emph{ICCV}, pages 12684--12694, 2021.

\bibitem[Caron et~al.(2018)Caron, Bojanowski, Joulin, and
  Douze]{caron2018deepcluster}
Mathilde Caron, Piotr Bojanowski, Armand Joulin, and Matthijs Douze.
\newblock Deep clustering for unsupervised learning of visual features.
\newblock In \emph{ECCV}, pages 139--156, 2018.

\bibitem[Cen et~al.(2025)Cen, Fang, Yang, Xie, Zhang, Shen, and Tian]{sam3dgs}
Jiazhong Cen, Jiemin Fang, Chen Yang, Lingxi Xie, Xiaopeng Zhang, Wei Shen, and
  Qi Tian.
\newblock Segment any {3D} {Gaussians}.
\newblock In \emph{AAAI}, pages 1971--1979, 2025.

\bibitem[Chen et~al.(2024)Chen, Narayanan, Ottou, Missaoui, Muriki, Pradalier,
  and Chen]{hsnerf2024}
Gerry Chen, Sunil~Kumar Narayanan, Thomas~Gautier Ottou, Benjamin Missaoui,
  Harsh Muriki, C{\'e}dric Pradalier, and Yongsheng Chen.
\newblock Hyperspectral neural radiance fields.
\newblock \emph{arXiv preprint arXiv:2403.14839}, 2024.

\bibitem[Chen et~al.(2016)Chen, Jiang, Li, Jia, and Ghamisi]{chen2016deep}
Yushi Chen, Hanlu Jiang, Chunyang Li, Xiuping Jia, and Pedram Ghamisi.
\newblock Deep feature extraction and classification of hyperspectral images
  based on convolutional neural networks.
\newblock \emph{IEEE Transactions on Geoscience and Remote Sensing},
  54\penalty0 (10):\penalty0 6232--6251, 2016.

\bibitem[{CIE}(2018)]{cie2018colorimetry}
{CIE}.
\newblock Colorimetry, 4th edition.
\newblock Technical Report CIE 015:2018, International Commission on
  Illumination, 2018.

\bibitem[Condor et~al.(2026)Condor, Moenne-Loccoz, Nimier-David, Didyk, Gojcic,
  and Wu]{condor2026nht}
Jorge Condor, Nicolas Moenne-Loccoz, Merlin Nimier-David, Piotr Didyk, Zan
  Gojcic, and Qi Wu.
\newblock Neural harmonic textures for high-quality primitive based neural
  reconstruction.
\newblock In \emph{ECCV}, pages 322--341, 2026.

\bibitem[Dobigeon et~al.(2014)Dobigeon, Tits, Somers, Altmann, and
  Coppin]{dobigeon2014vegetation}
Nicolas Dobigeon, Laurent Tits, Ben Somers, Yoann Altmann, and Pol Coppin.
\newblock A comparison of nonlinear mixing models for vegetated areas using
  simulated and real hyperspectral data.
\newblock \emph{IEEE Journal of Selected Topics in Applied Earth Observations
  and Remote Sensing}, 7\penalty0 (6):\penalty0 1869--1878, 2014.

\bibitem[Drumetz et~al.(2016)Drumetz, Veganzones, Henrot, Phlypo, Chanussot,
  and Jutten]{drumetz2016elmm}
Lucas Drumetz, Miguel-Angel Veganzones, Simon Henrot, Ronald Phlypo, Jocelyn
  Chanussot, and Christian Jutten.
\newblock Blind hyperspectral unmixing using an extended linear mixing model to
  address spectral variability.
\newblock \emph{IEEE Transactions on Image Processing}, 25\penalty0
  (8):\penalty0 3890--3905, 2016.

\bibitem[Feng et~al.(2023)Feng, Zhu, Fu, Jampani, Akula, He, Basu, Wang, and
  Wang]{feng2023layoutgpt}
Weixi Feng, Wanrong Zhu, Tsu-Jui Fu, Varun Jampani, Arjun Akula, Xuehai He,
  Sugato Basu, Xin~Eric Wang, and William~Yang Wang.
\newblock {LayoutGPT}: Compositional visual planning and generation with large
  language models.
\newblock In \emph{NeurIPS}, 2023.

\bibitem[Fischer et~al.(2026)Fischer, Georgiev, Groueix, Kim, Ritschel, and
  Deschaintre]{fischer2026sama}
Michael Fischer, Iliyan Georgiev, Thibault Groueix, Vladimir~G. Kim, Tobias
  Ritschel, and Valentin Deschaintre.
\newblock {SAMa}: Material-aware {3D} selection and segmentation.
\newblock In \emph{International Conference on 3D Vision (3DV)}, pages
  1812--1822, 2026.

\bibitem[Foi et~al.(2008)Foi, Trimeche, Katkovnik, and
  Egiazarian]{foi2008noise}
Alessandro Foi, Mejdi Trimeche, Vladimir Katkovnik, and Karen Egiazarian.
\newblock Practical {Poissonian-Gaussian} noise modeling and fitting for
  single-image raw-data.
\newblock \emph{IEEE Transactions on Image Processing}, 17\penalty0
  (10):\penalty0 1737--1754, 2008.

\bibitem[Gao et~al.(2026)Gao, Shu, Ye, Xiong, Makadia, Guo, Itti, and
  Chen]{gao2026threedcodebench}
Yipeng Gao, Lei Shu, Genzhi Ye, Xi Xiong, Ameesh Makadia, Meiqi Guo, Laurent
  Itti, and Jindong Chen.
\newblock {3DCodeBench}: Benchmarking agentic procedural {3D} modeling via
  code.
\newblock arXiv preprint arXiv:2606.01057, 2026.

\bibitem[Ghamisi et~al.(2017)Ghamisi, Yokoya, Li, Liao, Liu, Plaza, Rasti, and
  Plaza]{ghamisi2017advances}
Pedram Ghamisi, Naoto Yokoya, Jun Li, Wenzhi Liao, Sicong Liu, Javier Plaza,
  Behnood Rasti, and Antonio Plaza.
\newblock Advances in hyperspectral image and signal processing: A
  comprehensive overview of the state of the art.
\newblock \emph{IEEE Geoscience and Remote Sensing Magazine}, 5\penalty0
  (4):\penalty0 37--78, 2017.

\bibitem[Goodenough and Brown(2012)]{goodenough2012dirsig}
Adam~A. Goodenough and Scott~D. Brown.
\newblock {DIRSIG} 5: core design and implementation.
\newblock In \emph{Algorithms and Technologies for Multispectral,
  Hyperspectral, and Ultraspectral Imagery XVIII, Proc. SPIE 8390}, page
  83900H, 2012.

\bibitem[Green et~al.(1998)Green, Eastwood, Sarture, Chrien, Aronsson,
  Chippendale, Faust, Pavri, Chovit, Solis, et~al.]{green1998aviris}
Robert~O. Green, Michael~L. Eastwood, Charles~M. Sarture, Thomas~G. Chrien,
  Mikael Aronsson, Bruce~J. Chippendale, Jessica~A. Faust, Betina~E. Pavri,
  Christopher~J. Chovit, Manuel Solis, et~al.
\newblock Imaging spectroscopy and the airborne visible/infrared imaging
  spectrometer ({AVIRIS}).
\newblock \emph{Remote Sensing of Environment}, 65\penalty0 (3):\penalty0
  227--248, 1998.

\bibitem[Greff et~al.(2022)Greff, Belletti, Beyer, Doersch, Du, Duckworth,
  Fleet, Gnanapragasam, Golemo, Herrmann, et~al.]{greff2022kubric}
Klaus Greff, Francois Belletti, Lucas Beyer, Carl Doersch, Yilun Du, Daniel
  Duckworth, David~J. Fleet, Dan Gnanapragasam, Florian Golemo, Charles
  Herrmann, et~al.
\newblock Kubric: A scalable dataset generator.
\newblock In \emph{CVPR}, pages 3749--3761, 2022.

\bibitem[Gr{\"u}n et~al.(2025)Gr{\"u}n, Meyer, Weiherer, Egger, Stamminger, and
  Franke]{msgs}
Josef Gr{\"u}n, Lukas Meyer, Maximilian Weiherer, Bernhard Egger, Marc
  Stamminger, and Linus Franke.
\newblock Towards integrating multi-spectral imaging with {Gaussian} splatting.
\newblock In \emph{Vision, Modeling, and Visualization (VMV)}, 2025.

\bibitem[Hahlbohm et~al.(2026)Hahlbohm, Franke, Eisemann, and
  Magnor]{hahlbohm2026fastergs}
Florian Hahlbohm, Linus Franke, Martin Eisemann, and Marcus Magnor.
\newblock {Faster-GS}: Analyzing and improving {Gaussian} splatting
  optimization.
\newblock In \emph{CVPR}, pages 18946--18957, 2026.

\bibitem[Heinz and Chang(2001)]{FCLS}
Daniel~C. Heinz and Chein-I Chang.
\newblock Fully constrained least squares linear spectral mixture analysis
  method for material quantification in hyperspectral imagery.
\newblock \emph{IEEE Transactions on Geoscience and Remote Sensing},
  39\penalty0 (3):\penalty0 529--545, 2001.

\bibitem[Heylen et~al.(2014)Heylen, Parente, and Gader]{heylen2014review}
Rob Heylen, Mario Parente, and Paul Gader.
\newblock A review of nonlinear hyperspectral unmixing methods.
\newblock \emph{IEEE Journal of Selected Topics in Applied Earth Observations
  and Remote Sensing}, 7\penalty0 (6):\penalty0 1844--1868, 2014.

\bibitem[Hong et~al.(2019)Hong, Yokoya, Chanussot, and Zhu]{ELMM}
Danfeng Hong, Naoto Yokoya, Jocelyn Chanussot, and Xiao~Xiang Zhu.
\newblock An augmented linear mixing model to address spectral variability for
  hyperspectral unmixing.
\newblock \emph{IEEE Transactions on Image Processing}, 28\penalty0
  (4):\penalty0 1923--1938, 2019.

\bibitem[Hong et~al.(2022)Hong, Han, Yao, Gao, Zhang, Plaza, and
  Chanussot]{hong2022spectralformer}
Danfeng Hong, Zhu Han, Jing Yao, Lianru Gao, Bing Zhang, Antonio Plaza, and
  Jocelyn Chanussot.
\newblock {SpectralFormer}: Rethinking hyperspectral image classification with
  transformers.
\newblock \emph{IEEE Transactions on Geoscience and Remote Sensing},
  60:\penalty0 1--15, 2022.

\bibitem[Horler et~al.(1983)Horler, Dockray, and Barber]{horler1983rededge}
D.~N.~H. Horler, M. Dockray, and J. Barber.
\newblock The red edge of plant leaf reflectance.
\newblock \emph{International Journal of Remote Sensing}, 4\penalty0
  (2):\penalty0 273--288, 1983.

\bibitem[Hu et~al.(2024)Hu, Iscen, Jain, Kipf, Yue, Ross, Schmid, and
  Fathi]{hu2024scenecraft}
Ziniu Hu, Ahmet Iscen, Aashi Jain, Thomas Kipf, Yisong Yue, David~A. Ross,
  Cordelia Schmid, and Alireza Fathi.
\newblock {SceneCraft}: An {LLM} agent for synthesizing {3D} scenes as
  {Blender} code.
\newblock In \emph{ICML}, pages 19252--19282, 2024.

\bibitem[Huang et~al.(2024)Huang, Yu, Chen, Geiger, and Gao]{huang20242dgs}
Binbin Huang, Zehao Yu, Anpei Chen, Andreas Geiger, and Shenghua Gao.
\newblock {2D} {Gaussian} splatting for geometrically accurate radiance fields.
\newblock In \emph{ACM SIGGRAPH 2024 Conference Papers}, pages 32:1--32:11,
  2024.

\bibitem[Huang et~al.(2026)Huang, Wang, Tan, Yu, Zhang, Zheng, Liu, Chuang, and
  Zhang]{huang2026gwr}
Zheng-Hui Huang, Zhixiang Wang, Jiaming Tan, Ruihan Yu, Yidan Zhang, Bo Zheng,
  Yu-Lun Liu, Yung-Yu Chuang, and Kaipeng Zhang.
\newblock Generative world renderer.
\newblock arXiv preprint arXiv:2604.02329, 2026.

\bibitem[{International Commission on Illumination}(2020)]{cie2020metamerism}
{International Commission on Illumination}.
\newblock Metamerism.
\newblock International Lighting Vocabulary, CIE S 017:2020, entry 17-23-006,
  2020.

\bibitem[Iordache et~al.(2012)Iordache, Bioucas-Dias, and Plaza]{Bioucas2012A}
Marian-Daniel Iordache, José~M. Bioucas-Dias, and Antonio Plaza.
\newblock Total variation spatial regularization for sparse hyperspectral
  unmixing.
\newblock \emph{IEEE Transactions on Geoscience and Remote Sensing},
  50\penalty0 (11):\penalty0 4484--4502, 2012.

\bibitem[Jakob et~al.(2022)Jakob, Speierer, Roussel, Nimier-David, Vicini,
  Zeltner, Nicolet, Crespo, Leroy, and Zhang]{mitsuba3}
Wenzel Jakob, S{\'e}bastien Speierer, Nicolas Roussel, Merlin Nimier-David,
  Delio Vicini, Tizian Zeltner, Baptiste Nicolet, Miguel Crespo, Vincent Leroy,
  and Ziyi Zhang.
\newblock Mitsuba 3 renderer.
\newblock https://mitsuba-renderer.org, 2022.

\bibitem[Joshi et~al.(2025)Joshi, Han, Nugent, Gonzalez Saez-Diez, Zuo, Liu,
  Wen, Alexandropoulos, Kayan, Calveri, Sun, Liu, Shao, Raistrick, and
  Deng]{joshi2025infinigenarticulated}
Abhishek Joshi, Beining Han, Jack Nugent, Max Gonzalez Saez-Diez, Yiming Zuo,
  Jonathan Liu, Hongyu Wen, Stamatis Alexandropoulos, Karhan Kayan, Anna
  Calveri, Tao Sun, Gaowen Liu, Yi Shao, Alexander Raistrick, and Jia Deng.
\newblock Procedural generation of articulated simulation-ready assets.
\newblock \emph{arXiv preprint arXiv:2505.10755}, 2025.

\bibitem[Kerbl et~al.(2023)Kerbl, Kopanas, Leimk{\"u}hler, and Drettakis]{3dgs}
Bernhard Kerbl, Georgios Kopanas, Thomas Leimk{\"u}hler, and George Drettakis.
\newblock {3D} {Gaussian} splatting for real-time radiance field rendering.
\newblock \emph{ACM Transactions on Graphics}, 42\penalty0 (4):\penalty0
  139:1--139:14, 2023.

\bibitem[Kerr et~al.(2023)Kerr, Kim, Goldberg, Kanazawa, and Tancik]{lerf}
Justin Kerr, Chung~Min Kim, Ken Goldberg, Angjoo Kanazawa, and Matthew Tancik.
\newblock {LERF}: Language embedded radiance fields.
\newblock In \emph{ICCV}, pages 19729--19739, 2023.

\bibitem[Keshava and Mustard(2002)]{974727}
N. Keshava and J.F. Mustard.
\newblock Spectral unmixing.
\newblock \emph{IEEE Signal Processing Magazine}, 19\penalty0 (1):\penalty0
  44--57, 2002.

\bibitem[Kim et~al.(2023)Kim, Jin, Cho, and Baek]{nespof}
Youngchan Kim, Wonjoon Jin, Sunghyun Cho, and Seung-Hwan Baek.
\newblock Neural spectro-polarimetric fields.
\newblock In \emph{ACM SIGGRAPH Asia 2023 Conference Papers}, pages
  109:1--109:11, 2023.

\bibitem[Kirillov et~al.(2023)Kirillov, Mintun, Ravi, Mao, Rolland, Gustafson,
  Xiao, Whitehead, Berg, Lo, Doll{\'a}r, and Girshick]{SAM}
Alexander Kirillov, Eric Mintun, Nikhila Ravi, Hanzi Mao, Chloe Rolland, Laura
  Gustafson, Tete Xiao, Spencer Whitehead, Alexander~C. Berg, Wan-Yen Lo, Piotr
  Doll{\'a}r, and Ross Girshick.
\newblock Segment anything.
\newblock In \emph{ICCV}, pages 4015--4026, 2023.

\bibitem[Kokaly et~al.(2017)Kokaly, Clark, Swayze, Livo, Hoefen, Pearson, Wise,
  Benzel, Lowers, Driscoll, and Klein]{kokaly2017usgs}
Raymond~F. Kokaly, Roger~N. Clark, Gregg~A. Swayze, K.~Eric Livo, Todd~M.
  Hoefen, Neil~C. Pearson, Richard~A. Wise, William~M. Benzel, Heather~A.
  Lowers, Rhonda~L. Driscoll, and Anna~J. Klein.
\newblock {USGS} spectral library version 7.
\newblock U.S. Geological Survey Data Series 1035, 2017.

\bibitem[Kruse et~al.(1993)Kruse, Lefkoff, Boardman, Heidebrecht, Shapiro,
  Barloon, and Goetz]{kruse1993sips}
Fred~A. Kruse, A.~B. Lefkoff, Joseph~W. Boardman, K.~B. Heidebrecht, A.~T.
  Shapiro, P.~J. Barloon, and Alexander F.~H. Goetz.
\newblock The spectral image processing system ({SIPS}): Interactive
  visualization and analysis of imaging spectrometer data.
\newblock \emph{Remote Sensing of Environment}, 44\penalty0 (2--3):\penalty0
  145--163, 1993.

\bibitem[Ku et~al.(2026)Ku, Jubery, Krishnamurthy, and
  Ganapathysubramanian]{hs3dnerf}
Kibon Ku, Talukder~Z. Jubery, Adarsh Krishnamurthy, and Baskar
  Ganapathysubramanian.
\newblock {HS-3D-NeRF}: {3D} surface and hyperspectral reconstruction from
  stationary hyperspectral images using multi-channel {NeRFs}.
\newblock arXiv preprint arXiv:2602.16950, 2026.

\bibitem[Lawson and Hanson(1974)]{lawson1974nnls}
Charles~L. Lawson and Richard~J. Hanson.
\newblock \emph{Solving Least Squares Problems}.
\newblock Prentice-Hall, 1974.

\bibitem[Lee and Seung(1999)]{Lee1999NMF}
Daniel~D. Lee and H.~Sebastian Seung.
\newblock Learning the parts of objects by non-negative matrix factorization.
\newblock \emph{Nature}, 401\penalty0 (6755):\penalty0 788--791, 1999.

\bibitem[Lee(2013)]{lee2013pseudo}
Dong-Hyun Lee.
\newblock Pseudo-label: The simple and efficient semi-supervised learning
  method for deep neural networks.
\newblock In \emph{ICML Workshop on Challenges in Representation Learning},
  2013.

\bibitem[Li et~al.(2026{\natexlab{a}})Li, Shao, Li, Jiao, Bai, Zhou, Shen, Gu,
  and Yu]{li2026spread}
Minzhang Li, Kuixiang Shao, Xuebing Li, Yuyang Jiao, Yinuo Bai, Hengan Zhou,
  Sixian Shen, Jiayuan Gu, and Jingyi Yu.
\newblock {SPREAD}: Spatial-physical {RE}asoning via geometry aware diffusion.
\newblock In \emph{CVPR}, pages 31008--31018, 2026{\natexlab{a}}.

\bibitem[Li et~al.(2024)Li, Liu, Liu, Zhang, Zeng, and Liu]{spectralnerf}
Ru Li, Jia Liu, Guanghui Liu, Shengping Zhang, Bing Zeng, and Shuaicheng Liu.
\newblock {SpectralNeRF}: Physically based spectral rendering with neural
  radiance field.
\newblock In \emph{AAAI}, pages 3154--3162, 2024.

\bibitem[Li et~al.(2026{\natexlab{b}})Li, Liao, and Zhu]{li2026rcwm}
Zhiqi Li, Yuxuan Liao, and Bo Zhu.
\newblock Recursive code world models: Building complex worlds through
  recursive scene programs.
\newblock arXiv preprint arXiv:2609.11499, 2026{\natexlab{b}}.

\bibitem[Liang et~al.(2022)Liang, Wakaki, Nobuhara, and
  Nishino]{liang2022mcubes}
Yupeng Liang, Ryosuke Wakaki, Shohei Nobuhara, and Ko Nishino.
\newblock Multimodal material segmentation.
\newblock In \emph{CVPR}, pages 19800--19808, 2022.

\bibitem[Liang et~al.(2024)Liang, Zhang, Feng, Shan, and Jia]{gsir}
Zhihao Liang, Qi Zhang, Ying Feng, Ying Shan, and Kui Jia.
\newblock {GS-IR}: {3D} {Gaussian} splatting for inverse rendering.
\newblock In \emph{CVPR}, pages 21644--21653, 2024.

\bibitem[Lu et~al.(2024)Lu, Yu, Xu, Xiangli, Wang, Lin, and
  Dai]{lu2024scaffoldgs}
Tao Lu, Mulin Yu, Linning Xu, Yuanbo Xiangli, Limin Wang, Dahua Lin, and Bo
  Dai.
\newblock {Scaffold-GS}: Structured {3D} {Gaussians} for view-adaptive
  rendering.
\newblock In \emph{CVPR}, pages 20654--20664, 2024.

\bibitem[Meerdink et~al.(2019)Meerdink, Hook, Roberts, and Abbott]{ecostress}
Susan~K. Meerdink, Simon~J. Hook, Dar~A. Roberts, and Elsa~A. Abbott.
\newblock The {ECOSTRESS} spectral library version 1.0.
\newblock \emph{Remote Sensing of Environment}, 230:\penalty0 111196, 2019.

\bibitem[Melgani and Bruzzone(2004)]{melgani2004svm}
Farid Melgani and Lorenzo Bruzzone.
\newblock Classification of hyperspectral remote sensing images with support
  vector machines.
\newblock \emph{IEEE Transactions on Geoscience and Remote Sensing},
  42\penalty0 (8):\penalty0 1778--1790, 2004.

\bibitem[Meola et~al.(2011)Meola, Eismann, Moses, and Ash]{meola2011modeling}
Joseph Meola, Michael~T. Eismann, Randolph~L. Moses, and Joshua~N. Ash.
\newblock Modeling and estimation of signal-dependent noise in hyperspectral
  imagery.
\newblock \emph{Applied Optics}, 50\penalty0 (21):\penalty0 3829, 2011.

\bibitem[Meyer et~al.(2026)Meyer, Gr{\"u}n, Weiherer, Egger, Stamminger, and
  Franke]{mssplat_neural}
Lukas Meyer, Josef Gr{\"u}n, Maximilian Weiherer, Bernhard Egger, Marc
  Stamminger, and Linus Franke.
\newblock Multi-spectral {Gaussian} splatting with neural color representation.
\newblock \emph{Computer Graphics Forum}, 45\penalty0 (2), 2026.

\bibitem[Mildenhall et~al.(2020)Mildenhall, Srinivasan, Tancik, Barron,
  Ramamoorthi, and Ng]{nerf}
Ben Mildenhall, Pratul~P. Srinivasan, Matthew Tancik, Jonathan~T. Barron, Ravi
  Ramamoorthi, and Ren Ng.
\newblock {NeRF}: Representing scenes as neural radiance fields for view
  synthesis.
\newblock In \emph{ECCV}, pages 405--421, 2020.

\bibitem[M{\"u}ller et~al.(2022)M{\"u}ller, Evans, Schied, and
  Keller]{muller2022instantngp}
Thomas M{\"u}ller, Alex Evans, Christoph Schied, and Alexander Keller.
\newblock Instant neural graphics primitives with a multiresolution hash
  encoding.
\newblock \emph{ACM Transactions on Graphics}, 41\penalty0 (4):\penalty0
  102:1--102:15, 2022.

\bibitem[Narayanan et~al.(2026)Narayanan, Zhao, Gan, and Chen]{hgs}
Sunil~Kumar Narayanan, Lingjun Zhao, Lu Gan, and Yongsheng Chen.
\newblock Diffusion-denoised hyperspectral {Gaussian} splatting.
\newblock In \emph{International Conference on 3D Vision (3DV)}, pages
  1218--1227, 2026.

\bibitem[Nascimento and Bioucas-Dias(2005)]{vca}
Jos{\'e} M.~P. Nascimento and Jos{\'e}~M. Bioucas-Dias.
\newblock Vertex component analysis: A fast algorithm to unmix hyperspectral
  data.
\newblock \emph{IEEE Transactions on Geoscience and Remote Sensing},
  43\penalty0 (4):\penalty0 898--910, 2005.

\bibitem[Oquab et~al.(2024)Oquab, Darcet, Moutakanni, Vo, Szafraniec, Khalidov,
  Fernandez, HAZIZA, Massa, El-Nouby, Assran, Ballas, Galuba, Howes, Huang, Li,
  Misra, Rabbat, Sharma, Synnaeve, Xu, Jegou, Mairal, Labatut, Joulin, and
  Bojanowski]{DINOv2}
Maxime Oquab, Timoth{\'e}e Darcet, Th{\'e}o Moutakanni, Huy~V. Vo, Marc
  Szafraniec, Vasil Khalidov, Pierre Fernandez, Daniel HAZIZA, Francisco Massa,
  Alaaeldin El-Nouby, Mido Assran, Nicolas Ballas, Wojciech Galuba, Russell
  Howes, Po-Yao Huang, Shang-Wen Li, Ishan Misra, Michael Rabbat, Vasu Sharma,
  Gabriel Synnaeve, Hu Xu, Herve Jegou, Julien Mairal, Patrick Labatut, Armand
  Joulin, and Piotr Bojanowski.
\newblock {DINO}v2: Learning robust visual features without supervision.
\newblock \emph{Transactions on Machine Learning Research}, 2024.

\bibitem[Otsu(1979)]{otsu1979threshold}
Nobuyuki Otsu.
\newblock A threshold selection method from gray-level histograms.
\newblock \emph{IEEE Transactions on Systems, Man, and Cybernetics}, 9\penalty0
  (1):\penalty0 62--66, 1979.

\bibitem[Palsson et~al.(2022)Palsson, Sveinsson, and Ulfarsson]{BHUAE}
Burkni Palsson, Johannes~R. Sveinsson, and Magnus~O. Ulfarsson.
\newblock Blind hyperspectral unmixing using autoencoders: A critical
  comparison.
\newblock \emph{IEEE Journal of Selected Topics in Applied Earth Observations
  and Remote Sensing}, 15:\penalty0 1340--1372, 2022.

\bibitem[Perez et~al.(2025)Perez, Rojas, Hinojosa, Rueda-Chac{\'o}n, and
  Ghanem]{unmixnerf}
Fabian Perez, Sara Rojas, Carlos Hinojosa, Hoover Rueda-Chac{\'o}n, and Bernard
  Ghanem.
\newblock {UnMix-NeRF}: Spectral unmixing meets neural radiance fields.
\newblock In \emph{ICCV}, pages 26284--26293, 2025.

\bibitem[Pharr et~al.(2016)Pharr, Jakob, and Humphreys]{pharr2016pbrt}
Matt Pharr, Wenzel Jakob, and Greg Humphreys.
\newblock \emph{Physically Based Rendering: From Theory to Implementation}.
\newblock Morgan Kaufmann, 3 edition, 2016.

\bibitem[Plaza et~al.(2009)Plaza, Benediktsson, Boardman, Brazile, Bruzzone,
  Camps-Valls, Chanussot, Fauvel, Gamba, Gualtieri, Marconcini, Tilton, and
  Trianni]{PLAZA2009S110}
Antonio Plaza, Jon~Atli Benediktsson, Joseph~W. Boardman, Jason Brazile,
  Lorenzo Bruzzone, Gustavo Camps-Valls, Jocelyn Chanussot, Mathieu Fauvel,
  Paolo Gamba, Anthony Gualtieri, Mattia Marconcini, James~C. Tilton, and
  Giovanna Trianni.
\newblock Recent advances in techniques for hyperspectral image processing.
\newblock \emph{Remote Sensing of Environment}, 113:\penalty0 S110--S122, 2009.

\bibitem[Qin et~al.(2024)Qin, Li, Zhou, Wang, and Pfister]{langsplat}
Minghan Qin, Wanhua Li, Jiawei Zhou, Haoqian Wang, and Hanspeter Pfister.
\newblock {LangSplat}: {3D} language {Gaussian} splatting.
\newblock In \emph{CVPR}, pages 20051--20060, 2024.

\bibitem[Radford et~al.(2021)Radford, Kim, Hallacy, Ramesh, Goh, Agarwal,
  Sastry, Askell, Mishkin, Clark, Krueger, and Sutskever]{CLIP}
Alec Radford, Jong~Wook Kim, Chris Hallacy, Aditya Ramesh, Gabriel Goh,
  Sandhini Agarwal, Girish Sastry, Amanda Askell, Pamela Mishkin, Jack Clark,
  Gretchen Krueger, and Ilya Sutskever.
\newblock Learning transferable visual models from natural language
  supervision.
\newblock In \emph{ICML}, pages 8748--8763, 2021.

\bibitem[Raistrick et~al.(2023)Raistrick, Lipson, Ma, Mei, Wang, Zuo, Kayan,
  Wen, Han, Wang, Newell, Law, Goyal, Yang, and Deng]{raistrick2023infinigen}
Alexander Raistrick, Lahav Lipson, Zeyu Ma, Lingjie Mei, Mingzhe Wang, Yiming
  Zuo, Karhan Kayan, Hongyu Wen, Beining Han, Yihan Wang, Alejandro Newell, Hei
  Law, Ankit Goyal, Kaiyu Yang, and Jia Deng.
\newblock Infinite photorealistic worlds using procedural generation.
\newblock In \emph{CVPR}, pages 12630--12641, 2023.

\bibitem[Raistrick et~al.(2024)Raistrick, Mei, Kayan, Yan, Zuo, Han, Wen,
  Parakh, Alexandropoulos, Lipson, Ma, and Deng]{raistrick2024infinigenindoors}
Alexander Raistrick, Lingjie Mei, Karhan Kayan, David Yan, Yiming Zuo, Beining
  Han, Hongyu Wen, Meenal Parakh, Stamatis Alexandropoulos, Lahav Lipson, Zeyu
  Ma, and Jia Deng.
\newblock Infinigen indoors: Photorealistic indoor scenes using procedural
  generation.
\newblock In \emph{CVPR}, pages 21783--21794, 2024.

\bibitem[Raistrick et~al.(2026)Raistrick, Kayan, Nugent, Yan, Mei, Parakh, Wen,
  Li, Zuo, Liang, and Deng]{raistrick2026procfunc}
Alexander Raistrick, Karhan Kayan, Jack Nugent, David Yan, Lingjie Mei, Meenal
  Parakh, Hongyu Wen, Dylan Li, Yiming Zuo, Erich Liang, and Jia Deng.
\newblock {ProcFunc}: Function-oriented abstractions for procedural {3D}
  generation in {Python}.
\newblock \emph{arXiv preprint arXiv:2604.26943}, 2026.

\bibitem[Ramamoorthi and Hanrahan(2001)]{ramamoorthi2001sh}
Ravi Ramamoorthi and Pat Hanrahan.
\newblock An efficient representation for irradiance environment maps.
\newblock In \emph{ACM SIGGRAPH 2001 Conference Proceedings}, pages 497--500,
  2001.

\bibitem[Ren et~al.(2026)Ren, Wen, Fang, and Lu]{ren2025fastgs}
Shiwei Ren, Tianci Wen, Yongchun Fang, and Biao Lu.
\newblock {FastGS}: Training {3D} {Gaussian} splatting in 100 seconds.
\newblock In \emph{CVPR}, pages 26094--26103, 2026.

\bibitem[Roberts et~al.(1998)Roberts, Gardner, Church, Ustin, Scheer, and
  Green]{roberts1998mesma}
Dar~A. Roberts, Michael Gardner, Rick Church, Susan Ustin, Gary Scheer, and
  Robert~O. Green.
\newblock Mapping chaparral in the {Santa Monica Mountains} using multiple
  endmember spectral mixture models.
\newblock \emph{Remote Sensing of Environment}, 65\penalty0 (3):\penalty0
  267--279, 1998.

\bibitem[Roberts et~al.(2021)Roberts, Ramapuram, Ranjan, Kumar, Bautista,
  Paczan, Webb, and Susskind]{roberts2021hypersim}
Mike Roberts, Jason Ramapuram, Anurag Ranjan, Atulit Kumar, Miguel~Angel
  Bautista, Nathan Paczan, Russ Webb, and Joshua~M. Susskind.
\newblock Hypersim: A photorealistic synthetic dataset for holistic indoor
  scene understanding.
\newblock In \emph{ICCV}, pages 10912--10922, 2021.

\bibitem[Shen et~al.(2026)Shen, Kova{\v{c}}i{\v{c}}, Kulits, Wang, Li,
  Tenenbaum, Yuille, Chen, and Wu]{shen2026fourdcodebench}
Ruihong Shen, {\v{Z}}iga Kova{\v{c}}i{\v{c}}, Peter Kulits, Xingrui Wang,
  Zizhang Li, Joshua~B. Tenenbaum, Alan Yuille, Jieneng Chen, and Jiajun Wu.
\newblock {4DCodeBench}: Benchmarking agents on inverse graphics of dynamic
  scenes.
\newblock arXiv preprint arXiv:2610.03715, 2026.

\bibitem[Siddiqui et~al.(2023)Siddiqui, Porzi, Rota~Bul{\`o}, M{\"u}ller,
  Nie{\ss}ner, Dai, and Kontschieder]{siddiqui2023panoptic}
Yawar Siddiqui, Lorenzo Porzi, Samuel Rota~Bul{\`o}, Norman M{\"u}ller,
  Matthias Nie{\ss}ner, Angela Dai, and Peter Kontschieder.
\newblock Panoptic lifting for {3D} scene understanding with neural fields.
\newblock In \emph{CVPR}, pages 9043--9052, 2023.

\bibitem[Sinha et~al.(2025)Sinha, Graf, and
  Weinmann]{sinha2024spectralgaussians}
Saptarshi~Neil Sinha, Holger Graf, and Michael Weinmann.
\newblock {SpectralGaussians}: Semantic, spectral {3D} {Gaussian} splatting for
  multi-spectral scene representation, visualization and analysis.
\newblock \emph{ISPRS Journal of Photogrammetry and Remote Sensing},
  227:\penalty0 789--803, 2025.

\bibitem[Sohn et~al.(2020)Sohn, Berthelot, Li, Zhang, Carlini, Cubuk, Kurakin,
  Zhang, and Raffel]{sohn2020fixmatch}
Kihyuk Sohn, David Berthelot, Chun-Liang Li, Zizhao Zhang, Nicholas Carlini,
  Ekin~D. Cubuk, Alex Kurakin, Han Zhang, and Colin Raffel.
\newblock {FixMatch}: Simplifying semi-supervised learning with consistency and
  confidence.
\newblock In \emph{NeurIPS}, 2020.

\bibitem[Somers et~al.(2011)Somers, Asner, Tits, and
  Coppin]{somers2011variability}
Ben Somers, Gregory~P. Asner, Laurent Tits, and Pol Coppin.
\newblock Endmember variability in spectral mixture analysis: A review.
\newblock \emph{Remote Sensing of Environment}, 115\penalty0 (7):\penalty0
  1603--1616, 2011.

\bibitem[Straub et~al.(2019)Straub, Whelan, Ma, Chen, Wijmans, Green, Engel,
  Mur-Artal, Ren, Verma, et~al.]{straub2019replica}
Julian Straub, Thomas Whelan, Lingni Ma, Yufan Chen, Erik Wijmans, Simon Green,
  Jakob~J. Engel, Raul Mur-Artal, Carl Ren, Shobhit Verma, et~al.
\newblock The {Replica} dataset: A digital replica of indoor spaces.
\newblock \emph{arXiv preprint arXiv:1906.05797}, 2019.

\bibitem[Su et~al.(2019)Su, Li, Plaza, Marinoni, Gamba, and
  Chakravortty]{8628241}
Yuanchao Su, Jun Li, Antonio Plaza, Andrea Marinoni, Paolo Gamba, and Somdatta
  Chakravortty.
\newblock {DAEN}: Deep autoencoder networks for hyperspectral unmixing.
\newblock \emph{IEEE Transactions on Geoscience and Remote Sensing},
  57\penalty0 (7):\penalty0 4309--4321, 2019.

\bibitem[Tancik et~al.(2020)Tancik, Srinivasan, Mildenhall, Fridovich-Keil,
  Raghavan, Singhal, Ramamoorthi, Barron, and Ng]{tancik2020fourier}
Matthew Tancik, Pratul~P. Srinivasan, Ben Mildenhall, Sara Fridovich-Keil,
  Nithin Raghavan, Utkarsh Singhal, Ravi Ramamoorthi, Jonathan~T. Barron, and
  Ren Ng.
\newblock Fourier features let networks learn high frequency functions in low
  dimensional domains.
\newblock In \emph{NeurIPS}, 2020.

\bibitem[Tarabalka et~al.(2009)Tarabalka, Benediktsson, and
  Chanussot]{tarabalka2009clustering}
Yuliya Tarabalka, J{\'o}n~Atli Benediktsson, and Jocelyn Chanussot.
\newblock Spectral--spatial classification of hyperspectral imagery based on
  partitional clustering techniques.
\newblock \emph{IEEE Transactions on Geoscience and Remote Sensing},
  47\penalty0 (8):\penalty0 2973--2987, 2009.

\bibitem[Thirgood et~al.(2025)Thirgood, Mendez, Ling, Storey, and
  Hadfield]{HyperGS}
Christopher Thirgood, Oscar Mendez, Erin Ling, Jon Storey, and Simon Hadfield.
\newblock {HyperGS}: Hyperspectral {3D} {Gaussian} splatting.
\newblock In \emph{CVPR}, pages 5970--5979, 2025.

\bibitem[Tucker(1979)]{tucker1979ndvi}
Compton~J. Tucker.
\newblock Red and photographic infrared linear combinations for monitoring
  vegetation.
\newblock \emph{Remote Sensing of Environment}, 8\penalty0 (2):\penalty0
  127--150, 1979.

\bibitem[Upchurch and Niu(2022)]{upchurch2022dms}
Paul Upchurch and Ransen Niu.
\newblock A dense material segmentation dataset for indoor and outdoor scene
  parsing.
\newblock In \emph{ECCV}, pages 450--466, 2022.

\bibitem[Verbin et~al.(2025)Verbin, Hedman, Mildenhall, Zickler, Barron, and
  Srinivasan]{RefNeRF}
Dor Verbin, Peter Hedman, Ben Mildenhall, Todd Zickler, Jonathan~T. Barron, and
  Pratul~P. Srinivasan.
\newblock {Ref-NeRF}: Structured view-dependent appearance for neural radiance
  fields.
\newblock \emph{IEEE Transactions on Pattern Analysis and Machine
  Intelligence}, 47\penalty0 (11):\penalty0 9426--9437, 2025.

\bibitem[Wilkie et~al.(2014)Wilkie, Nawaz, Droske, Weidlich, and
  Hanika]{wilkie2014hero}
Alexander Wilkie, Sehera Nawaz, Marc Droske, Andrea Weidlich, and Johannes
  Hanika.
\newblock Hero wavelength spectral sampling.
\newblock \emph{Computer Graphics Forum}, 33\penalty0 (4):\penalty0 123--131,
  2014.

\bibitem[Winter(1999)]{winter1999nfindr}
Michael~E. Winter.
\newblock {N-FINDR}: An algorithm for fast autonomous spectral end-member
  determination in hyperspectral data.
\newblock In \emph{Imaging Spectrometry V, Proc. SPIE 3753}, pages 266--275,
  1999.

\bibitem[Xia et~al.(2026)Xia, Li, Li, Ma, Xu, Liu, Cui, Lin, Ma, Wang, Song,
  and Wei]{xia2026sage}
Hongchi Xia, Xuan Li, Zhaoshuo Li, Qianli Ma, Jiashu Xu, Ming-Yu Liu, Yin Cui,
  Tsung-Yi Lin, Wei-Chiu Ma, Shenlong Wang, Shuran Song, and Fangyin Wei.
\newblock {SAGE}: Scalable agentic {3D} scene generation for embodied {AI}.
\newblock In \emph{CVPR}, pages 22358--22368, 2026.

\bibitem[Yang et~al.(2024)Yang, Sun, Weihs, VanderBilt, Herrasti, Han, Wu,
  Haber, Krishna, Liu, Callison-Burch, Yatskar, Kembhavi, and
  Clark]{yang2024holodeck}
Yue Yang, Fan-Yun Sun, Luca Weihs, Eli VanderBilt, Alvaro Herrasti, Winson Han,
  Jiajun Wu, Nick Haber, Ranjay Krishna, Lingjie Liu, Chris Callison-Burch,
  Mark Yatskar, Aniruddha Kembhavi, and Christopher Clark.
\newblock Holodeck: Language guided generation of {3D} embodied {AI}
  environments.
\newblock In \emph{CVPR}, pages 16227--16237, 2024.

\bibitem[Ye et~al.(2024)Ye, Danelljan, Yu, and Ke]{ye2024gaussiangrouping}
Mingqiao Ye, Martin Danelljan, Fisher Yu, and Lei Ke.
\newblock {Gaussian Grouping}: Segment and edit anything in {3D} scenes.
\newblock In \emph{ECCV}, pages 162--179, 2024.

\bibitem[Ye et~al.(2026)Ye, Mantri, Chen, Zhang, Zheng, Li, Chen, Huang, and
  Qin]{ye2026code4scene}
Xiaokang Ye, Siddhant~Hitesh Mantri, Zimeng Chen, Edward Zhang, Zhaoxu Zheng,
  Yuanheng Li, Yizhao Chen, Tianyang Huang, and Lianhui Qin.
\newblock {Code4Scene}: Benchmarking coding agents for constructing and editing
  {3D} scenes.
\newblock arXiv preprint arXiv:2609.36777, 2026.

\bibitem[Yu et~al.(2024)Yu, Chen, Huang, Sattler, and
  Geiger]{yu2024mipsplatting}
Zehao Yu, Anpei Chen, Binbin Huang, Torsten Sattler, and Andreas Geiger.
\newblock {Mip-Splatting}: Alias-free {3D} {Gaussian} splatting.
\newblock In \emph{CVPR}, pages 19447--19456, 2024.

\bibitem[Zhang et~al.(2016)Zhang, Zhai, Zhang, and Li]{zhang2016sssc}
Hongyan Zhang, Han Zhai, Liangpei Zhang, and Pingxiang Li.
\newblock Spectral--spatial sparse subspace clustering for hyperspectral remote
  sensing images.
\newblock \emph{IEEE Transactions on Geoscience and Remote Sensing},
  54\penalty0 (6):\penalty0 3672--3684, 2016.

\bibitem[Zhang et~al.(2021)Zhang, Srinivasan, Deng, Debevec, Freeman, and
  Barron]{NeRFactor}
Xiuming Zhang, Pratul~P. Srinivasan, Boyang Deng, Paul Debevec, William~T.
  Freeman, and Jonathan~T. Barron.
\newblock {NeRFactor}: Neural factorization of shape and reflectance under an
  unknown illumination.
\newblock \emph{ACM Transactions on Graphics}, 40\penalty0 (6):\penalty0
  237:1--237:18, 2021.

\bibitem[Zhi et~al.(2021)Zhi, Laidlow, Leutenegger, and
  Davison]{zhi2021semanticnerf}
Shuaifeng Zhi, Tristan Laidlow, Stefan Leutenegger, and Andrew~J. Davison.
\newblock In-place scene labelling and understanding with implicit scene
  representation.
\newblock In \emph{ICCV}, pages 15838--15847, 2021.

\bibitem[Zhou et~al.(2024)Zhou, Chang, Jiang, Fan, Zhu, Xu, Chari, You, Wang,
  and Kadambi]{feature3dgs}
Shijie Zhou, Haoran Chang, Sicheng Jiang, Zhiwen Fan, Zehao Zhu, Dejia Xu,
  Pradyumna Chari, Suya You, Zhangyang Wang, and Achuta Kadambi.
\newblock {Feature 3DGS}: Supercharging {3D} {Gaussian} splatting to enable
  distilled feature fields.
\newblock In \emph{CVPR}, pages 21676--21685, 2024.

\end{thebibliography}

\ifdefined\REVIEW
\else
\clearpage
\twocolumn[{\centering{\Large\bf Supplementary Material\par}\vspace{1.5em}}]
\renewcommand{\method}{Spectral World Renderer\xspace}
\setcounter{section}{0}\setcounter{figure}{0}\setcounter{table}{0}\setcounter{equation}{0}
\renewcommand{\thesection}{\Alph{section}}\renewcommand{\thefigure}{S\arabic{figure}}\renewcommand{\thetable}{S\arabic{table}}\renewcommand{\theequation}{S\arabic{equation}}
\setcounter{totalnumber}{5}\setcounter{dbltopnumber}{3}\setcounter{topnumber}{3}
\renewcommand{\topfraction}{0.94}\renewcommand{\dbltopfraction}{0.94}\renewcommand{\textfraction}{0.05}
\renewcommand{\floatpagefraction}{0.75}
\renewcommand{\dblfloatpagefraction}{0.75}
\setlength{\floatsep}{6pt plus 2pt minus 2pt}\setlength{\dblfloatsep}{6pt plus 2pt minus 2pt}\setlength{\textfloatsep}{8pt plus 2pt minus 2pt}

\newcommand{\tocdots}{\leaders\hbox to 0.6em{\hss.\hss}\hfill}
\newcommand{\tocsec}[2]{\noindent\hbox to\columnwidth{\hbox to 1.2em{\textbf{\ref*{#1}}\hss}\textbf{#2}\tocdots\hbox to 1.6em{\hss\textbf{\pageref*{#1}}}}\par\vspace{1pt}}
\newcommand{\tocsub}[2]{\noindent\hbox to\columnwidth{\hspace{1.2em}\hbox to 2.4em{\ref*{#1}\hss}#2\tocdots\hbox to 1.6em{\hss\pageref*{#1}}}\par}
\noindent{\large\bfseries Contents}\par\medskip
{\small
\tocsec{sec:supp_bench}{\bench\ Worlds and Protocol}
\tocsub{sec:supp_worlds}{Worlds and Rendering}
\tocsub{sec:supp_protocol}{Protocol and Metrics}
\smallskip
\tocsec{sec:supp_spectra}{Spectral Sources and Validation}
\smallskip
\tocsec{sec:supp_impl}{Implementation Details}
\tocsub{sec:supp_impl_rgb}{RGB Scaffold and Harmonic Features}
\tocsub{sec:supp_impl_clusters}{Appearance Clusters}
\smallskip
\tocsec{sec:supp_ident}{Material Identification and Design Analysis}
\tocsub{sec:supp_rare}{Rare Classes}
\tocsub{sec:capacity_exploration}{Dictionary Capacity and Appearance Guidance}
\tocsub{sec:supp_decoder}{Decoder Components}
\smallskip
\tocsec{sec:supp_real}{Real Captures}
\smallskip
\tocsec{sec:supp_eff}{Efficiency}
\smallskip
\tocsec{sec:supp_apps}{Material Queries and Applications}
\tocsub{sec:supp_fpq}{Spectral Material Queries}
\tocsub{sec:app_sensor}{Simulated Multispectral Sensors}
\tocsub{sec:app_stats}{3D Material Statistics}
\tocsub{sec:app_anomaly}{Spectra Unexplained by the Dictionary}
\tocsub{sec:app_edit}{Material Editing}
\smallskip
\tocsec{sec:supp_gallery}{Additional Visualizations}
}\par\medskip
\noindent Figures, tables and equations numbered with S belong to this supplement; other numbers refer to the main paper.\par\medskip

\providecommand{\bench}{SWR-Bench\xspace}
\providecommand{\swr}{Spectral World Renderer\xspace}

\section{\bench\ Worlds and Protocol}\label{sec:supp_bench}
\label{sec:synthetic_benchmark}

\noindent This section describes the worlds and their rendering (Sec.~\ref{sec:supp_worlds}) and the evaluation protocol and metrics (Sec.~\ref{sec:supp_protocol}); Sec.~\ref{sec:supp_spectra} details the measured spectra.

\subsection{Worlds and Rendering}\label{sec:supp_worlds}

\begin{table*}[t]
\centering
\caption{\textbf{The eight worlds of \bench.} $C$: material classes; Specimens: distinct measured specimens (the total removes duplicates across worlds; Table~\ref{tab:spectra_worlds}). Views are split by camera. Noise: electrons $N_e$ at the reference radiance and read noise $\sigma_{\mathrm{read}}$; Noise PSNR compares noisy and clean test views (dB). City variants share the City cameras and splits.}
\label{tab:swr_worlds}
\vspace{-2mm}
\scriptsize\setlength{\tabcolsep}{3pt}
\sbox\swrtabbox{\begin{tabular*}{\textwidth}{@{\extracolsep{\fill}}l l l l cc rrr cc@{}}
 & & & & & & \multicolumn{3}{c}{Views} & & \\
\cmidrule(lr){7-9}
World & Content & Cameras & Transport & $C$ & Specimens & Train & Val & Test & $N_e\,/\,\sigma_{\mathrm{read}}$ & \makecell{Noise\\PSNR} \\
\midrule
Forbidden City & palace: golden roofs, courtyards, water & 3 orbits, 45--430\,m & two-term & 9 & 47 & 135 & 21 & 24 & 5k\,/\,2 & 45.4 \\
Market & market square: stalls, pottery, walls & 3 rings, inward/downward & two-term & 6 & 29 & 135 & 21 & 24 & 5k\,/\,2 & 45.6 \\
Botanical & botanical garden: planters, trees, pergolas & 3 rings, inward/downward & two-term & 6 & 29 & 135 & 21 & 24 & 5k\,/\,2 & 46.4 \\
NVIDIA HQ & offices: glazed facades, 300 trees & spiral, facade, campus routes & extended & 8 & 27 & 180 & 28 & 32 & 5k\,/\,2 & 46.7 \\
Study & scholar's study: shelves, paper, ink, ceramics & eye-level and elevated orbits & path traced & 5 & 5 & 64 & 16 & 16 & 3k\,/\,3 & 44.8 \\
Karst & river valley: limestone cliffs, clear water & river banks, cliff, high loop & two-term + water & 5 & 10 & 128 & 32 & 32 & 5k\,/\,2 & 44.3 \\
City & campus block: roads, 297 trees, vehicles & 4 orbits, 16--95\,m & two-term & 11 & 56 & 180 & 28 & 32 & 5k\,/\,2 & 45.2 \\
Office & office: 35 LED downlights, daylight & eye-level and elevated loops & path traced & 8 & 35 & 64 & 16 & 16 & 3k\,/\,3 & 46.0 \\
\midrule
Total & & & & & 125 & 1{,}021 & 183 & 200 & & \\
City variants & metamer, stress, morning, afternoon & as City & two-term & 11--12 & 56--60 & \multicolumn{3}{c}{as City} & as City & -- \\
\end{tabular*}}%
\swrrainbowbar{\wd\swrtabbox}\par\nointerlineskip\vspace{1pt}\usebox\swrtabbox\par\nointerlineskip\vspace{1pt}\swrrainbowbar{\wd\swrtabbox}
\vspace{1mm}
\end{table*}

\begin{table}[t]
\centering
\caption{\textbf{Training configuration.} On \bench, all methods use the same noisy $480{\times}270$ training views, normalized by $H_\mathrm{ref}$ and clamped to $[0,1]$ at loading (the released Study and Office cubes are not clipped; Sec.~\ref{sec:supp_worlds}), with a shared stereo point cloud; evaluation renders at $960{\times}540$. HyperGS and DD-HGS~\citep{HyperGS,hgs} use VCA/NNLS~\citep{vca,lawson1974nnls} for unmixing.}
\label{tab:supp_settings}
\footnotesize\setlength{\tabcolsep}{3pt}
\sbox\swrtabbox{\begin{tabular*}{\columnwidth}{@{\extracolsep{\fill}}l l@{}}
Setting & Configuration \\
\midrule
\multicolumn{2}{@{}l}{\textit{RGB geometry and feature refinement}} \\
RGB scaffold & 30,000 steps \\
Harmonic features & 5,000 refinement steps, no densification \\
\midrule
\multicolumn{2}{@{}l}{\textit{Appearance clustering and spectral fitting}} \\
Spectrum sampling & $\le$2,304 per view, after $3{\times}3$ box filter \\
Foreground & spectrum norm $>0.03$ \\
Optimizer & Adam, 30,000 steps \\
Batch & 2,048 pixels from one camera \\
Learning rate & $10^{-3}$, cosine decay to $10^{-5}$ \\
Endmembers & fixed for 3,000 steps, then $10^{-3}$ \\
Assignment weight & $\lambda_p(t)=\max(0.005,\,0.05(1-t/3000))$ \\
Evaluation weights & EMA, decay 0.99 \\
Decoder ablation & no guidance; $K{=}C$ (\bench), 16 (real) \\
\midrule
\multicolumn{2}{@{}l}{\textit{Numerical safeguards}} \\
Abundance norm. & denominator $\ge10^{-8}$ \\
$\mathcal L_p$ support & rays with $\sum_i w_i>10^{-6}$ \\
\midrule
\multicolumn{2}{@{}l}{\textit{Baselines}} \\
3DGS+VCA/NNLS & $L$-channel 3DGS, 30,000 steps; $K{=}C$ \\
HyperGS & 2,000-step AE pretraining + 30,000 steps \\
DD-HGS & 60,000 steps with diffusion denoiser \\
UnMix-NeRF & official schedule, 30,000 steps \\
\end{tabular*}}%
\swrrainbowbar{\wd\swrtabbox}\par\nointerlineskip\vspace{1pt}\usebox\swrtabbox\par\nointerlineskip\vspace{1pt}\swrrainbowbar{\wd\swrtabbox}
\end{table}

\noindent\textbf{Scene construction.} Each world is defined by an executable scene program specifying geometry, material assignments, illumination and cameras. Reference photographs, maps and text guide the construction. Spectral specimens remain fixed on surfaces across viewpoints, ensuring a common material identity for all observations. Table~\ref{tab:swr_worlds} summarizes the scene configurations and camera splits.

\noindent\textbf{Shared samples.} Spectral images, RGB and ground truth share 16 sub-pixel samples per pixel, so labels and area proportions describe the first surfaces along the sample rays. Radiance reflected or transmitted by water and glazing is therefore attributed to the water or glazing class, and interreflected light to the surface on which it is observed; both are also stored separately.

\noindent\textbf{Rendering.} All worlds share 64 Gaussian bands centred from 420 to 980\,nm (20\,nm full width at half maximum) at $960\times540$. Outdoor worlds use Lambertian surfaces under ASTM G173 sunlight and skylight~\citep{astmg173}; NVIDIA HQ adds glossy highlights, glass reflection and an analytic sky, and Karst adds refracting water with pure-water absorption. The Study and Office are path traced in Mitsuba~\citep{mitsuba3} with interreflection, sampling reflectance at the band centres. RGB is noiseless: CIE 1931 colour matching gives XYZ, which is adapted to D65 by the Bradford transform (70\,\% partial adaptation indoors) and converted to sRGB with one fixed exposure per world.

\noindent\textbf{Noise.} Every world is provided clean and noisy, with shot and read noise added per band to the clean $960\times540$ band radiance $H_{\mathrm{clean}}$. With $N_e$ electrons at the reference radiance $H_\mathrm{ref}$ and read noise $\sigma_{\mathrm{read}}$ as listed in Table~\ref{tab:swr_worlds}, the observation is $H=(H_\mathrm{ref}/N_e)\,[\mathrm{Poisson}(N_eH_{\mathrm{clean}}/H_\mathrm{ref})+\mathcal N(0,\sigma_{\mathrm{read}}^2)]$. Outdoors, $H_\mathrm{ref}$ is the maximum clean radiance over all training pixels and bands, and negative values of the released noisy cubes are clipped to zero; in the Study and Office, $H_\mathrm{ref}$ is the 99.5th percentile of the first training view, excluding emitters, and the released cubes are not clipped. For training, all methods normalize the views by $H_\mathrm{ref}$ and clamp them to $[0,1]$ (Table~\ref{tab:supp_settings}).

\noindent\textbf{Classes and City variants.} Finer labels are merged into one class when they play the same material role with similar spectra, for example the marbles and granite of Market and Botanical, the near-neutral grey paving and brick of Forbidden City, and bark and soil; spectrally distinct labels remain separate even when rare, such as painted eaves and yellow road paint, whereas white road paint (0.3\,\% of City pixels) is merged into asphalt. Four City variants change one factor each: in \emph{metamer}, one broadleaf specimen is replaced by a foliage-coloured mineral and two car finishes by other measured materials, each matched in RGB colour to the original and grouped into one additional class; in \emph{stress}, 87 trees carry dry leaves; the remaining two use a morning or an afternoon sun. Every object keeps its instance and specimen.

\subsection{Protocol and Metrics}\label{sec:supp_protocol}
\noindent\textbf{Evaluation views.} Training views are $2\times2$ box averages of the noisy cubes, normalized by $H_\mathrm{ref}$, at $480\times270$; evaluation uses clean $960\times540$ foreground radiance. All methods share camera splits, classes and the stereo point cloud. Ground-truth consistency is verified by full primary-ray coverage, at least 99.9\,\% agreement between instance and material maps, and exact recomposition of outdoor radiance from the stored shading coefficients.

\noindent\textbf{Metrics.} PSNR and the spectral angle SAM are computed against clean radiance $\mathbf H_{\mathrm{clean}}$ on the ground-truth support. For non-zero spectra, $\mathrm{SAM}(\hat{\mathbf H},\mathbf H_{\mathrm{clean}})=\arccos(\hat{\mathbf H}^\top\mathbf H_{\mathrm{clean}}/(\|\hat{\mathbf H}\|_2\|\mathbf H_{\mathrm{clean}}\|_2))$; reported angles use degrees.
\begin{itemize}\setlength{\itemsep}{0pt}
\item \emph{mIoU.} Endmembers are mapped many-to-one to classes by majority vote on training views. With the endmember-to-class map $\pi$ frozen, each test pixel takes the class of its dominant endmember, $c=\pi(\arg\max_k\hat A_k)$, not the class of largest summed abundance; endmembers that dominate no training pixel stay unassigned.
\item \emph{Endmember angle error (EM).} Each class is paired with a distinct endmember by the linear-sum (Hungarian) assignment of minimum total angle to the mean clean spectrum of its pure training pixels (area proportion at least 0.9). With $K\ge C$, the remaining $K-C$ endmembers do not enter EM, and the endmember-to-class map plays no role.
\item \emph{Measured-specimen NNLS.} Every second test pixel along each axis is unmixed by non-negative least squares against all measured specimens of the world, each under sunlight and skylight, and labelled by the largest class-summed contribution; Karst is scored on its opaque classes. This reference is accurate on clean radiance and remains competitive under noise, and gives the reference IoU of Figure~\ref{fig:rare_classes}. It assumes that each pixel is a mixture of first-surface specimens under sunlight and skylight, and is therefore evaluated on Forbidden City, Market, Botanical, Karst and City. It is undefined for the path-traced Study and Office, whose radiance includes interreflection, and for NVIDIA HQ, whose radiance includes glass-reflected sources, glossy highlights and atmospheric haze along the camera ray. The distant-haze class of Forbidden City is not atmospheric: it consists of distant surfaces with assigned specimens and is therefore covered by the dictionary.
\end{itemize}

\providecommand{\bench}{SWR-Bench\xspace}

\section{Spectral Sources and Validation}\label{sec:supp_spectra}
\label{sec:spectra}

\begin{table*}[t]
\centering
\caption{\textbf{Materials and measured spectra in \bench.} Per world: material classes, specimens drawn from ECOSTRESS (ECO) and USGS, and the classes with their specimen counts. Specimens shared between worlds are counted in each world; the eight worlds use 125 distinct specimens.}
\label{tab:spectra_worlds}
\footnotesize\setlength{\tabcolsep}{4pt}\renewcommand{\arraystretch}{1.1}
\sbox\swrtabbox{\begin{tabular*}{\textwidth}{@{\extracolsep{\fill}}l c c c c p{0.60\textwidth}@{}}
World & $C$ & ECO & USGS & Total & Material classes (specimens) \\
\midrule
Forbidden City & 9 & 27 & 20 & 47 & yellow roof tile (5), vermilion paint (3), vegetation (8), white marble (4), grey masonry (9), painted eaves (4), soil \& bark (7), water (3), distant haze (4) \\
Market & 6 & 29 & 0 & 29 & \multirow{2}{0.60\textwidth}{travertine (1), stone: white/pink marble, granite (7), vegetation (8), fired clay (4), sandy soil (4), wood (5)} \\
Botanical & 6 & 29 & 0 & 29 & \\
NVIDIA HQ & 8 & 21 & 6 & 27 & white paint (3), glazing (1), vegetation (8), dark panels/frames (2), concrete (3), asphalt (2), soil \& bark (7), wood (1) \\
Study & 5 & 3 & 2 & 5 & aged wood (1), paper (1), ink (1), inkstone (1), ceramic (1) \\
Karst & 5 & 10 & 0 & 10 & limestone (2), vegetation (5), soil (1), wood/bark (2), water (absorption model) \\
City & 11 & 33 & 23 & 56 & concrete (7), soil \& bark (6), vegetation (7), fired clay (8), asphalt (4), yellow road paint (2), dark glazing (2), painted facade (6), metal roofing (4), vehicle (8), backdrop (2) \\
Office & 8 & 11 & 24 & 35 & wood (6), dark laminate (5), textile (5), vinyl floor (3), white paint (6), vegetation (4), metal (4), display (2) \\
\midrule
City-metamer & 12 & 35 & 24 & 59 & City materials plus substitutes for one broadleaf specimen and two car finishes (3) \\
City-stress & 11 & 36 & 24 & 60 & City materials with vegetation expanded from 7 to 11 specimens \\
\end{tabular*}}%
\swrrainbowbar{\wd\swrtabbox}\par\nointerlineskip\vspace{1pt}\usebox\swrtabbox\par\nointerlineskip\vspace{1pt}\swrrainbowbar{\wd\swrtabbox}
\end{table*}

\noindent\textbf{Sources and selection.} Reflectance comes from ECOSTRESS~\citep{ecostress} and the USGS Spectral Library~7~\citep{kokaly2017usgs}, as shown in Table~\ref{tab:spectra_worlds}. A library record is accepted as a specimen only if it is a reflectance measurement that covers 400--1000\,nm with gaps of at most 20\,nm and values in $[0,1]$, and if its metadata match the assigned material class, such as photosynthetic grass for foliage or bark for trunks. Accepted specimens are interpolated on a 1\,nm grid without smoothing or extrapolation and integrated against the 64 band responses of Sec.~\ref{sec:supp_worlds}, truncated at $\pm20$\,nm. Several specimens per class and separate spectra under sunlight and skylight give each material a range of appearances, as shown in Figure~3 of the main paper.

\begin{table}[t]
\centering
\caption{\textbf{Validation of the assigned reflectance spectra.} Pairs: checked pairs of specimen and material type; a specimen assigned to two material types is checked against both, so the 125 distinct specimens of the eight worlds form 154 pairs. Range: reflectance within $[0,1]$. Ref.: eligible pairs within $3^\circ$ of independent library records of their material type, with the median angle in degrees. Leaf: red-edge and NDVI checks passed.}
\label{tab:spectra_realism}
\footnotesize\setlength{\tabcolsep}{3pt}
\sbox\swrtabbox{\begin{tabular*}{\columnwidth}{@{\extracolsep{\fill}}l r c c r c@{}}
Scope & Pairs & Range & Ref. & Median & Leaf \\
\midrule
Forbidden City & 47 & 47/47 & 22/24 & 1.28 & 8/8 \\
Market & 29 & 29/29 & 17/21 & 2.01 & 8/8 \\
Botanical & 29 & 29/29 & 17/21 & 2.01 & 8/8 \\
NVIDIA HQ & 27 & 27/27 & 17/19 & 1.35 & 8/8 \\
Study & 5 & 5/5 & 1/2 & 6.09 & -- \\
Karst & 10 & 10/10 & 9/9 & 0.60 & 5/5 \\
City & 56 & 56/56 & 20/30 & 2.07 & 7/7 \\
Office & 35 & 35/35 & 8/18 & 3.46 & 4/4 \\
\midrule
Six outdoor worlds & 119 & 119/119 & 47/61 & 1.47 & 17/17 \\
All eight worlds & 154 & 154/154 & 55/78 & 2.00 & 21/21 \\
\quad natural & 45 & 45/45 & 37/40 & 1.25 & 21/21 \\
\quad man-made & 109 & 109/109 & 18/38 & 3.46 & -- \\
\end{tabular*}}%
\swrrainbowbar{\wd\swrtabbox}\par\nointerlineskip\vspace{1pt}\usebox\swrtabbox\par\nointerlineskip\vspace{1pt}\swrrainbowbar{\wd\swrtabbox}
\end{table}

\noindent\textbf{Realism.} Rendered natural materials agree closely with independent library records of the same material type, excluding the specimens used in \bench, as reported in Table~\ref{tab:spectra_realism}: nearly all eligible natural-material assignments lie within $3^\circ$ of a reference, and the vegetation spectra pass red-edge and NDVI checks~\citep{horler1983rededge,tucker1979ndvi}. The dry and mineral substitutes in the City variants lack these vegetation signatures. Man-made materials are more varied, and pigment proxies for paints deviate most.

\noindent\textbf{Proxies.} Some objects use a measured specimen of a related material, such as pigment powders for paints and road markings, cotton for paper, or granite for an inkstone.

\section{Implementation Details}
\label{sec:supp_impl}

\noindent Table~\ref{tab:supp_settings} lists the training settings of \method\ and of all baselines. This section details the RGB scaffold and harmonic features (Sec.~\ref{sec:supp_impl_rgb}) and the appearance clusters (Sec.~\ref{sec:supp_impl_clusters}); Sec.~\ref{sec:supp_eff} reports the computational cost.

\subsection{RGB Scaffold and Harmonic Features}\label{sec:supp_impl_rgb}
\noindent\textbf{RGB scaffold.} The scaffold uses FastGS densification and pruning~\citep{ren2025fastgs}, which clones and splits Gaussians that cover many high-error pixels and prunes by opacity and multi-view importance. After fitting, colours are discarded and the rasterizer's screen-space variance is folded into the scales,
\begin{equation}
    \boldsymbol{\sigma}_i^2\leftarrow\boldsymbol{\sigma}_i^2+0.3\,\mathrm{med}_{v}(\delta_{iv}/F_v)^2.
\end{equation}
The square and update are componentwise. Here $\delta_{iv}$ is the depth of Gaussian $i$ in training camera $v$, $F_v$ the focal length in pixels of the $480\times270$ training views, and the median runs over the training cameras that see the Gaussian; the term transfers the $0.3$\,px$^2$ low-pass variance of the RGB rasterizer to the 3D kernel. The harmonic features are then refined for 5k steps without densification. The RGB objective combines mean absolute error $\mathcal L_1$ and structural dissimilarity $\mathcal L_{\mathrm{D\text{-}SSIM}}$. Each Gaussian stores three position coordinates, a four-value rotation quaternion, three scales and one opacity, plus four 12-dimensional feature vectors. The encoding $\gamma(\mathbf f)=[\sin\mathbf f;\cos\mathbf f]$ is applied elementwise.

\noindent\textbf{Tetrahedral interpolation.} The canonical tetrahedron is regular with inradius 1 in the frame of the Gaussian scaled by $\boldsymbol\sigma_i$, with vertices $(\pm\sqrt6,-\sqrt2,-1)$, $(0,2\sqrt2,-1)$ and $(0,0,3)$. The barycentric coordinates of the point of maximal response are affine in its position and are not clamped, so they extrapolate linearly outside the tetrahedron.

\noindent\textbf{RGB anchor refinement.} Table~4 of the main paper uses single seed-42 runs. \emph{Anchor only} passes the 30k-step RGB scaffold with randomly initialized harmonic features directly to spectral fitting; \emph{Geometry only} and \emph{Features only} run the 5k RGB refinement steps with the other group frozen; \emph{Refinement (full)} updates both. Spectral fitting and evaluation are identical across rows, so the apple scores differ slightly from the two-run means of Table~3.

\subsection{Appearance Clusters}\label{sec:supp_impl_clusters}
\noindent With the sampling settings in Table~\ref{tab:supp_settings}, each smoothed pixel spectrum $\tilde{\mathbf H}(\mathbf r)$ is described by the appearance descriptor $\mathbf q_{\mathbf r}$ of Eq.~8 of the main paper, whose first $L$ entries $\mathbf q_{\mathbf r}^{1:L}$ form the unit spectrum. The standardized log norm is $b=\mathrm{clip}\big((\log\|\tilde{\mathbf H}(\mathbf r)\|_2-\nu)/\max(0.1,\,1.4826\,\mathrm{MAD}),-4,4\big)$, where $\nu$ and MAD are the median and median absolute deviation of $\log\|\tilde{\mathbf H}(\mathbf r)\|_2$. The weight $\kappa=\mathrm{clip}(0.5\,\mathrm{med}_{\mathbf r}\|\mathbf q_{\mathbf r}^{1:L}-\bar{\mathbf q}\|_2,0.05,0.2)$, with $\bar{\mathbf q}$ the normalized mean of the unit spectra, adapts to each scene. Farthest-point seeding and 20 Lloyd iterations of spherical $k$-means give $K$ clusters. The endmember parameters $\boldsymbol\eta_k$ of Eq.~5 of the main paper start from the logit of the unit-spectrum part of cluster centre $k$, clipped to $[10^{-5},1-10^{-5}]$.

\section{Material Identification and Design Analysis}\label{sec:supp_ident}

\noindent This section analyses which classes are hard to identify (Sec.~\ref{sec:supp_rare}), how dictionary size and appearance guidance affect identification (Sec.~\ref{sec:capacity_exploration}), and how decoder components affect fidelity and identification (Sec.~\ref{sec:supp_decoder}).

\subsection{Rare Classes}\label{sec:supp_rare}
\begin{figure}[t]
\centering
\includegraphics[width=\columnwidth]{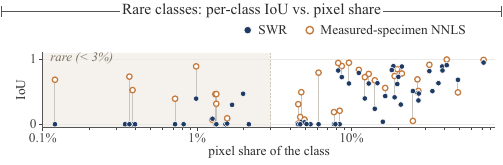}
\vspace{-6mm}
\caption{\textbf{Per-class IoU and pixel share.} Per-class IoU against pixel share for the classes of the eight worlds, with $K$ equal to the class count. Measured-specimen NNLS, defined for five worlds (Sec.~\ref{sec:supp_protocol}), separates many rare classes that \method\ merges.}
\label{fig:rare_classes}
\end{figure}

\noindent Figure~\ref{fig:rare_classes} relates per-class IoU to pixel share at $K{=}C$. Measured-specimen NNLS separates many rare classes merged by SWR, indicating that dictionary learning and material allocation contribute to the errors beyond ambiguity in the observations.

\subsection{Dictionary Capacity and Appearance Guidance}\label{sec:capacity_exploration}
\noindent A material can span several spectral appearances, so the endmember count need not equal the number of material classes. Table~\ref{tab:capacity_exploration}(a) shows that a larger dictionary improves identification over the class-count setting in every world. The validation-selected size is also the best test candidate in all but one world.

Panel~(b) compares methods at the same $K=24$. SWR compares favourably with 3DGS+VCA/NNLS in every world, with a small margin on Study, so capacity alone does not explain the difference in identification. Appearance guidance improves most worlds but reduces mIoU on Study and Karst, so appearance guidance and dictionary size play distinct roles.

\begin{table}[tp]
\centering
\caption{\textbf{Dictionary capacity and material identification} (test mIoU). (a) Dictionary size with appearance guidance; shading marks the $K$ selected on validation views. (b) At matched $K{=}24$, guided and unguided SWR against 3DGS+VCA/NNLS; shading marks the best per world.}
\label{tab:capacity_exploration}
\footnotesize\setlength{\tabcolsep}{3pt}
\textbf{(a) Dictionary capacity}\par\smallskip
\sbox\swrtabbox{\begin{tabular*}{\columnwidth}{@{\extracolsep{\fill}}l c cccc@{}}
 & & \multicolumn{4}{c}{Test mIoU at $K$} \\
\cmidrule(l){3-6}
World & $C$ & $C$ & 16 & 24 & 32 \\
\midrule
Forbidden City & 9 & 0.410 & 0.500 & 0.538 & \cellcolor{rkA}0.552 \\
Market & 6 & 0.340 & 0.611 & \cellcolor{rkA}0.784 & 0.754 \\
Botanical & 6 & 0.315 & 0.628 & 0.684 & \cellcolor{rkA}0.712 \\
NVIDIA HQ & 8 & 0.202 & 0.307 & \cellcolor{rkA}0.311 & 0.315 \\
Study & 5 & 0.402 & 0.516 & \cellcolor{rkA}0.521 & 0.517 \\
Karst & 5 & 0.504 & 0.562 & 0.561 & \cellcolor{rkA}0.725 \\
City & 11 & 0.263 & 0.370 & 0.477 & \cellcolor{rkA}0.519 \\
Office & 8 & 0.437 & 0.500 & 0.524 & \cellcolor{rkA}0.527 \\
\end{tabular*}}%
\swrrainbowbar{\wd\swrtabbox}\par\nointerlineskip\vspace{1pt}\usebox\swrtabbox\par\nointerlineskip\vspace{1pt}\swrrainbowbar{\wd\swrtabbox}
\par\medskip
\textbf{(b) Matched capacity ($K{=}24$)}\par\smallskip
\sbox\swrtabbox{\begin{tabular*}{\columnwidth}{@{\extracolsep{\fill}}l ccc@{}}
 & \multicolumn{2}{c}{SWR} & \\
\cmidrule(lr){2-3}
World & guided & unguided & 3DGS+VCA/NNLS \\
\midrule
Forbidden City & \cellcolor{rkA}0.538 & 0.396 & 0.277 \\
Market & \cellcolor{rkA}0.784 & 0.505 & 0.323 \\
Botanical & \cellcolor{rkA}0.684 & 0.629 & 0.356 \\
NVIDIA HQ & \cellcolor{rkA}0.311 & 0.247 & 0.199 \\
Study & 0.521 & \cellcolor{rkA}0.588 & 0.514 \\
Karst & 0.561 & \cellcolor{rkA}0.694 & 0.395 \\
City & \cellcolor{rkA}0.477 & 0.348 & 0.179 \\
Office & \cellcolor{rkA}0.524 & 0.307 & 0.163 \\
\end{tabular*}}%
\swrrainbowbar{\wd\swrtabbox}\par\nointerlineskip\vspace{1pt}\usebox\swrtabbox\par\nointerlineskip\vspace{1pt}\swrrainbowbar{\wd\swrtabbox}
\end{table}

\noindent\textbf{Components of appearance guidance.} On Market at $K{=}24$, the assignment regularization $\mathcal L_p$ and cluster initialization each improve mIoU substantially over using neither, and their combination is higher than either alone. The two components of appearance guidance are therefore complementary. The adaptive brightness weight $\kappa$ also gives higher mIoU than fixed weights of 0.05 and 0.2.

\subsection{Decoder Components}\label{sec:supp_decoder}
\label{sec:supp_analysis}

\noindent\textbf{Fidelity and identification.} Table~5 of the main paper compares decoder variants at $K{=}C$ without appearance guidance (neither cluster initialization nor $\mathcal L_p$), averaged over two seeds. Replacing the endmember mixture with direct spectra changes PSNR only marginally, whereas replacing the four corner features with one feature per Gaussian lowers it noticeably. Reconstruction quality therefore depends on the harmonic features.

Identification responds differently. Removing the encoding $\gamma$ or the intensity $s_i$ preserves much of Market's PSNR but reduces mIoU, as mixtures absorb shading. Decoding after compositing (row d) slightly raises Market PSNR while clearly lowering mIoU. This comparison illustrates why accurate spectra alone do not establish correct material identification. Pre-compositing decoding also retains a mixture at each ray--Gaussian intersection for direct primitive queries.

\begin{figure*}[t]
\centering
\includegraphics[width=\textwidth]{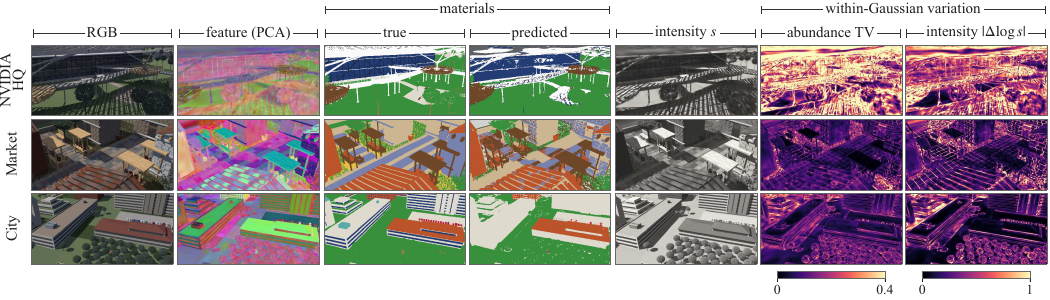}
\caption{\textbf{Harmonic features and their decoding.} Columns show RGB, the per-intersection feature (PCA), true and predicted materials, intensity, and the variation of mixture and intensity within each Gaussian against its mean feature in the same viewing direction (fixed scales).}
\label{fig:nht_features}
\end{figure*}

\noindent\textbf{Feature structure.} Projected on their principal components, the harmonic features follow both material and shading, as shown in Figure~\ref{fig:nht_features}. Within a Gaussian, intensity follows shading, whereas abundances change mainly at material boundaries, so one primitive can span an edge.

\begin{figure*}[t]
\centering
\includegraphics[width=\textwidth]{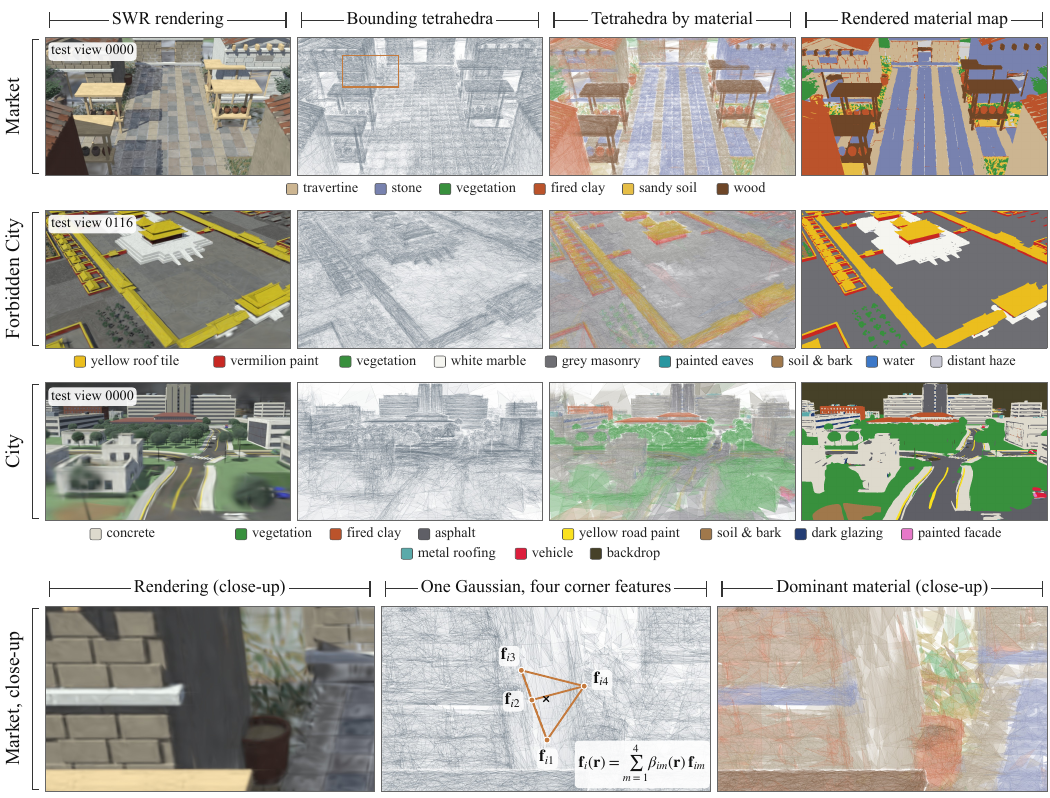}
\vspace{-5mm}
\caption{\textbf{Gaussian tetrahedra and decoded materials.} Tetrahedra follow each Gaussian's scale, rotation and centre; their barycentric coordinates interpolate four corner features according to Eq.~3 of the main paper. Columns show the rendering, opacity- and visibility-weighted tetrahedra, centre-decoded classes, and material maps. Bottom: one Market Gaussian, marked by the amber window.}
\label{fig:tetra_supp}
\end{figure*}

\noindent\textbf{Material structure on explicit primitives.} The primitive visualizations in Figure~\ref{fig:tetra_supp} associate surface regions with their decoded material classes. Class boundaries in the rendered maps align with changes between primitives, so material selections (Sec.~\ref{sec:supp_fpq}) follow the primitives across views.

\section{Real Captures}\label{sec:supp_real}
\label{sec:s_real}

\begin{table}[tp]
\centering
\footnotesize
\setlength{\tabcolsep}{2.6pt}
\caption{\textbf{Real captures and splits.} Test views are spread along each trajectory (about one in seven); NeSpoF also holds out every fifth training view for validation. HS-NeRF sensors: Surface Optics (SO) and BaySpec (BS).}
\label{tab:supp_real_datasets}
\vspace{-2mm}
\sbox\swrtabbox{\begin{tabular}{@{}l l c c c r r@{}}
Dataset & Scene & Bands & nm & Resolution & Train & Test \\
\midrule
\multirow{4}{*}{NeSpoF} & scene1 & \multirow{4}{*}{21} & \multirow{4}{*}{450--650} & \multirow{4}{*}{$512{\times}428$} & 26 & 4 \\
 & scene2 & & & & 21 & 4 \\
 & scene3 & & & & 15 & 3 \\
 & scene4 & & & & 21 & 4 \\
\midrule
HSI-SC & apple & 204 & 397--1004 & $512{\times}512$ & 56 & 8 \\
\midrule
\multirow{4}{*}{\makecell[l]{HS-NeRF\\(SO)}} & Basil & \multirow{4}{*}{128} & \multirow{4}{*}{371--1044} & \multirow{4}{*}{$348{\times}260$} & 41 & 6 \\
 & Rosemary & & & & 37 & 6 \\
 & Harsh1 & & & & 42 & 7 \\
 & Harsh2 & & & & 20 & 3 \\
\makecell[l]{HS-NeRF (BS)} & Fakeplant1 & 141 & 401--1087 & $320{\times}256$ & 50 & 8 \\
\midrule
\rowcolor{ourscol}\multicolumn{5}{@{}l}{Total (10 scenes)} & 329 & 53 \\
\end{tabular}}%
\swrrainbowbar{\wd\swrtabbox}\par\nointerlineskip\vspace{1pt}\usebox\swrtabbox\par\nointerlineskip\vspace{1pt}\swrrainbowbar{\wd\swrtabbox}
\vspace{1mm}
\end{table}

\noindent\textbf{Alignment and scene selection.} HS-NeRF evaluation requires a unique correspondence among hyperspectral cubes, RGB images and COLMAP cameras. Five of the eight scenes admit this correspondence and use the views listed in Table~\ref{tab:supp_real_datasets}; Fakeplant2 and Pinecone, for example, are excluded because their processed RGB backgrounds differ from the raw hyperspectral captures. Figures~\ref{fig:qualitative}--\ref{fig:qualitative_apple} compare the first test view of each evaluated scene; errors concentrate at boundaries and thin structures.

\begin{figure*}[p]
    \centering
    \includegraphics[width=\textwidth,height=0.92\textheight,keepaspectratio]{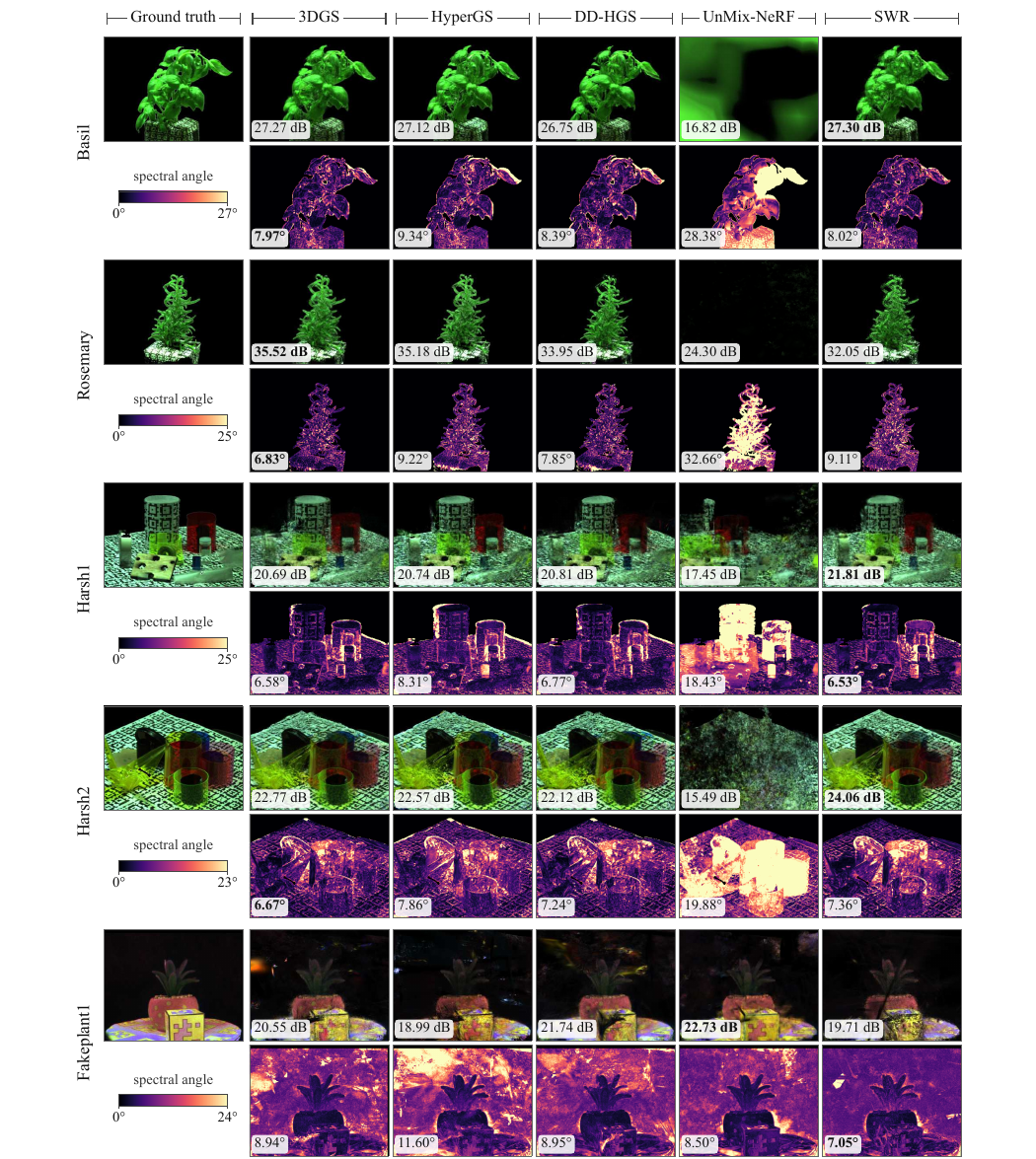}
    \vspace{-2mm}
    \caption{\textbf{Qualitative comparison on HS-NeRF test views}, the first test view of each of the five scenes. Rows show false colour (630, 550 and 460\,nm) and per-pixel spectral angle (left scale). Bold marks the best scores.}
    \label{fig:qualitative}
\end{figure*}

\begin{figure*}[p]
    \centering
    \includegraphics[width=0.98\textwidth]{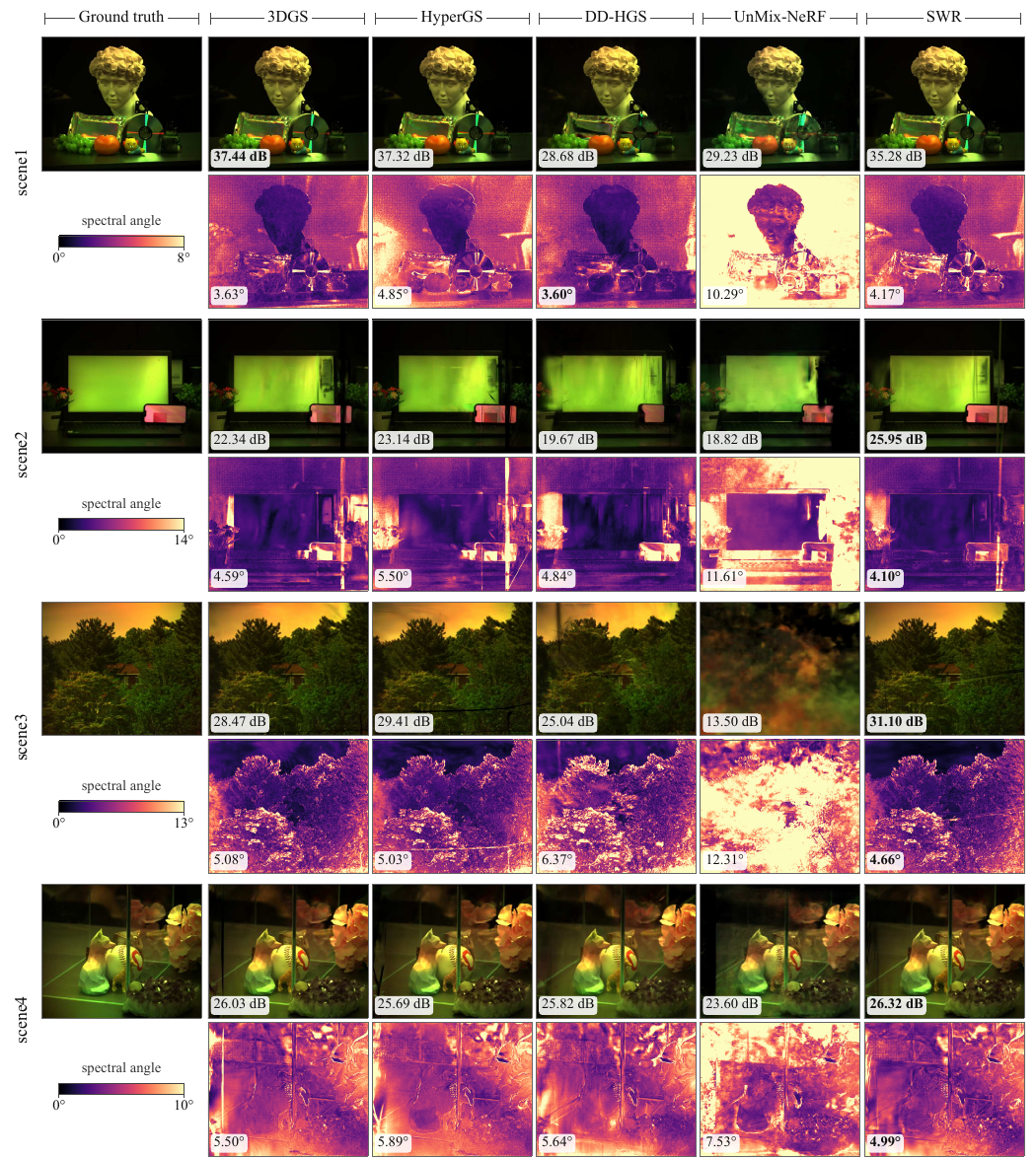}
    \vspace{-2mm}
    \caption{\textbf{Qualitative comparison on NeSpoF test views}, the first test view of each of the four scenes, as in Figure~\ref{fig:qualitative}.}
    \label{fig:qualitative_nespof}
\end{figure*}

\begin{figure*}[t]
    \centering
    \includegraphics[width=0.98\textwidth]{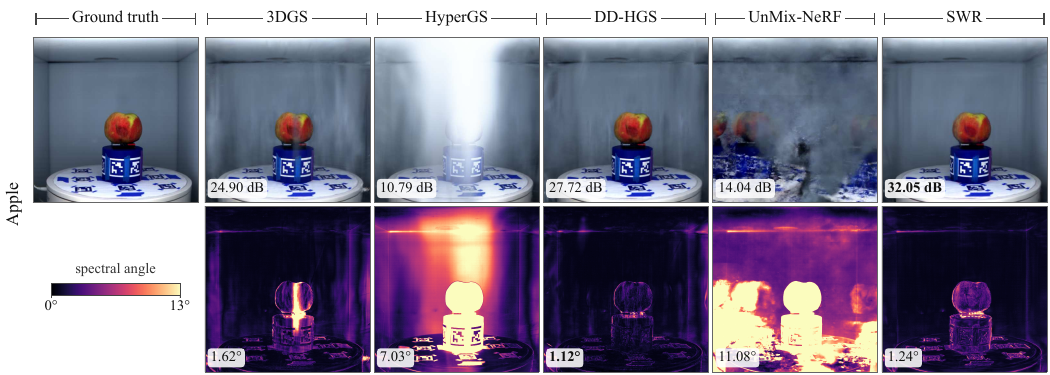}
    \vspace{-2mm}
    \caption{\textbf{Qualitative comparison on the HSI-SC apple}, first test view, as in Figure~\ref{fig:qualitative}.}
    \label{fig:qualitative_apple}
    \vspace{-2mm}
\end{figure*}

\begin{table*}[t]
\centering
\footnotesize
\setlength{\tabcolsep}{2pt}
\caption{\textbf{Spectral reconstruction on real captures, per scene} (test views; the main paper reports the means). Each value is the mean of two runs; baselines are trained for their authors' iteration counts (30k, or 60k for DD-HGS with its spectral denoiser). MS/HS: measured multispectral/hyperspectral; 3DGS stores one channel per band. PSNR is higher-is-better; SAM and its 95th percentile SAM$_{\mathrm{p95}}$ are lower-is-better. Darker shading indicates better ranks.}
\label{tab:s_main_full}
\vspace{5pt}
\sbox\swrtabbox{\begin{tabular}{cc >{\columncolor{ourscol}}c>{\columncolor{ourscol}}c>{\columncolor{ourscol}}c ccc ccc ccc ccc}
\multirow{2}{*}{Source} & \multirow{2}{*}{Scene} & \multicolumn{3}{c}{\method} & \multicolumn{3}{c}{HyperGS} & \multicolumn{3}{c}{DD-HGS} & \multicolumn{3}{c}{3DGS} & \multicolumn{3}{c}{UnMix-NeRF} \\
\cmidrule(lr){3-5}\cmidrule(lr){6-8}\cmidrule(lr){9-11}\cmidrule(lr){12-14}\cmidrule(lr){15-17}
 &  & PSNR & SAM & SAM$_{\mathrm{p95}}$ & PSNR & SAM & SAM$_{\mathrm{p95}}$ & PSNR & SAM & SAM$_{\mathrm{p95}}$ & PSNR & SAM & SAM$_{\mathrm{p95}}$ & PSNR & SAM & SAM$_{\mathrm{p95}}$ \\
\midrule
\multirow{5}{*}{MS} & NeSpoF scene1 & \cellcolor{rkB}30.77 & \cellcolor{rkC}3.83 & \cellcolor{rkB}7.18 & \cellcolor{rkA}32.24 & 4.62 & 9.56 & 25.52 & \cellcolor{rkB}3.76 & \cellcolor{rkC}7.64 & \cellcolor{rkC}30.62 & \cellcolor{rkA}3.60 & \cellcolor{rkA}6.87 & 20.51 & 12.21 & 19.98 \\
 & NeSpoF scene2 & \cellcolor{rkA}28.43 & \cellcolor{rkA}3.62 & \cellcolor{rkA}8.34 & \cellcolor{rkB}26.44 & 4.80 & \cellcolor{rkB}11.02 & 19.99 & \cellcolor{rkB}4.34 & \cellcolor{rkC}11.89 & \cellcolor{rkC}24.99 & \cellcolor{rkC}4.57 & 12.32 & 21.64 & 11.75 & 22.48 \\
 & NeSpoF scene3 & \cellcolor{rkA}30.12 & \cellcolor{rkA}4.17 & \cellcolor{rkA}9.21 & 23.88 & \cellcolor{rkC}5.57 & \cellcolor{rkB}11.34 & \cellcolor{rkC}24.22 & 5.85 & 13.64 & \cellcolor{rkB}25.65 & \cellcolor{rkB}5.42 & \cellcolor{rkC}12.77 & 14.92 & 8.88 & 16.17 \\
 & NeSpoF scene4 & \cellcolor{rkC}28.20 & \cellcolor{rkB}3.92 & \cellcolor{rkB}7.84 & \cellcolor{rkA}28.76 & 4.55 & 8.97 & 28.02 & \cellcolor{rkA}3.86 & \cellcolor{rkA}7.60 & \cellcolor{rkB}28.65 & \cellcolor{rkC}3.94 & \cellcolor{rkC}8.06 & 19.83 & 8.54 & 17.65 \\
\cmidrule(lr){2-17}
 & Mean & \cellcolor{rkA}29.38 & \cellcolor{rkA}3.89 & \cellcolor{rkA}8.14 & \cellcolor{rkB}27.83 & 4.88 & 10.23 & 24.44 & \cellcolor{rkC}4.45 & \cellcolor{rkC}10.19 & \cellcolor{rkC}27.48 & \cellcolor{rkB}4.38 & \cellcolor{rkB}10.01 & 19.23 & 10.35 & 19.07 \\
\midrule
HS & HSI-SC apple & \cellcolor{rkA}30.52 & \cellcolor{rkA}1.34 & \cellcolor{rkA}3.86 & \cellcolor{rkC}25.10 & 2.69 & \cellcolor{rkC}8.42 & \cellcolor{rkB}25.25 & \cellcolor{rkB}1.92 & \cellcolor{rkB}7.14 & 23.45 & \cellcolor{rkC}2.67 & 10.99 & 14.17 & 10.21 & 35.03 \\
\midrule
\multirow{6}{*}{HS} & Basil & \cellcolor{rkA}30.28 & \cellcolor{rkC}6.23 & \cellcolor{rkB}17.34 & \cellcolor{rkC}29.98 & 7.08 & 19.40 & 29.13 & \cellcolor{rkB}6.09 & \cellcolor{rkC}17.59 & \cellcolor{rkB}30.23 & \cellcolor{rkA}5.77 & \cellcolor{rkA}16.24 & 14.86 & 17.20 & 36.10 \\
 & Rosemary & 34.14 & \cellcolor{rkC}8.38 & \cellcolor{rkC}20.62 & \cellcolor{rkA}35.73 & 8.99 & 23.41 & \cellcolor{rkC}34.43 & \cellcolor{rkB}7.30 & \cellcolor{rkB}19.79 & \cellcolor{rkB}35.65 & \cellcolor{rkA}6.51 & \cellcolor{rkA}16.62 & 24.94 & 21.43 & 54.38 \\
 & Harsh1 & \cellcolor{rkA}21.33 & \cellcolor{rkA}6.18 & \cellcolor{rkB}17.43 & 20.05 & 8.05 & 22.07 & \cellcolor{rkC}20.50 & \cellcolor{rkC}6.59 & \cellcolor{rkC}18.39 & \cellcolor{rkB}20.61 & \cellcolor{rkB}6.28 & \cellcolor{rkA}17.29 & 15.94 & 13.79 & 43.32 \\
 & Harsh2 & \cellcolor{rkA}24.78 & \cellcolor{rkC}6.71 & \cellcolor{rkC}17.60 & \cellcolor{rkC}23.31 & 7.19 & 17.61 & 22.94 & \cellcolor{rkB}6.45 & \cellcolor{rkB}16.73 & \cellcolor{rkB}23.53 & \cellcolor{rkA}6.11 & \cellcolor{rkA}15.61 & 15.32 & 22.00 & 49.98 \\
 & Fakeplant1 & 26.98 & \cellcolor{rkB}6.49 & \cellcolor{rkA}12.59 & \cellcolor{rkB}27.65 & 7.22 & 15.44 & \cellcolor{rkA}30.01 & \cellcolor{rkA}6.10 & \cellcolor{rkB}12.76 & \cellcolor{rkC}27.38 & \cellcolor{rkC}6.53 & \cellcolor{rkC}13.45 & 22.62 & 7.67 & 15.96 \\
\cmidrule(lr){2-17}
 & Mean & \cellcolor{rkA}27.50 & \cellcolor{rkC}6.80 & \cellcolor{rkC}17.12 & 27.34 & 7.71 & 19.59 & \cellcolor{rkC}27.40 & \cellcolor{rkB}6.51 & \cellcolor{rkB}17.05 & \cellcolor{rkB}27.48 & \cellcolor{rkA}6.24 & \cellcolor{rkA}15.84 & 18.74 & 16.42 & 39.95 \\
\end{tabular}}%
\swrrainbowbar{\wd\swrtabbox}\par\nointerlineskip\vspace{1pt}\usebox\swrtabbox\par\nointerlineskip\vspace{1pt}\swrrainbowbar{\wd\swrtabbox}
\end{table*}

\section{Efficiency}\label{sec:supp_eff}
\label{sec:s_efficiency}

\noindent\textbf{Representation size.} Per-Gaussian storage in SWR is independent of band count and comparable to that of HyperGS, whereas per-band fields grow with the number of bands. The shared endmember dictionary adds a small fixed cost, and complete model storage also includes the shared decoder. Over the ten real captures of Table~\ref{tab:s_main_full}, complete SWR models are consistently several times smaller than those of HyperGS, and smaller still relative to DD-HGS.

\noindent\textbf{Computational cost.} The complete City pipeline takes 397\,s on one RTX~4090 (Table~6 of the main paper): RGB scaffold fitting 73\,s, harmonic-feature refinement 37\,s, appearance clustering 68\,s and spectral fitting 220\,s. Only appearance clustering and spectral fitting depend on $K$; scaffold fitting and harmonic-feature refinement are shared across dictionary sizes. Per-intersection decoding adds computation relative to a stored spectrum. The fused half-precision kernel reduces this cost through joint evaluation and negligible-weight pruning, reaching 34.9 joint spectral and label views/s, at least eight times the unmixing baselines in Table~6 of the main paper, whose rendering time is dominated by per-pixel NNLS. Half precision skips intersections of weight below $10^{-4}$; its dominant classes match full precision on 99.97\,\% of pixels.

\providecommand{\bench}{SWR-Bench\xspace}

\section{Material Queries and Applications}\label{sec:supp_apps}

\noindent Because \method\ renders full spectra and stores abundances on explicit Gaussians, a trained model can be queried and analysed directly. Sec.~\ref{sec:supp_fpq} retrieves materials from clicked and library spectra. Secs.~\ref{sec:app_sensor}--\ref{sec:app_edit} illustrate four further uses on City without retraining, each compared with the ground truth of \bench\ on the test views: simulated sensors, 3D material statistics, detection of unexplained spectra and material editing. Sensor simulation and material statistics use the City model of Table~2; detection and editing use a model whose dictionary size equals the class count, trained on the unaltered world.

\subsection{Spectral Material Queries}\label{sec:supp_fpq}

\begin{figure*}[t]
\centering
\includegraphics[width=\textwidth]{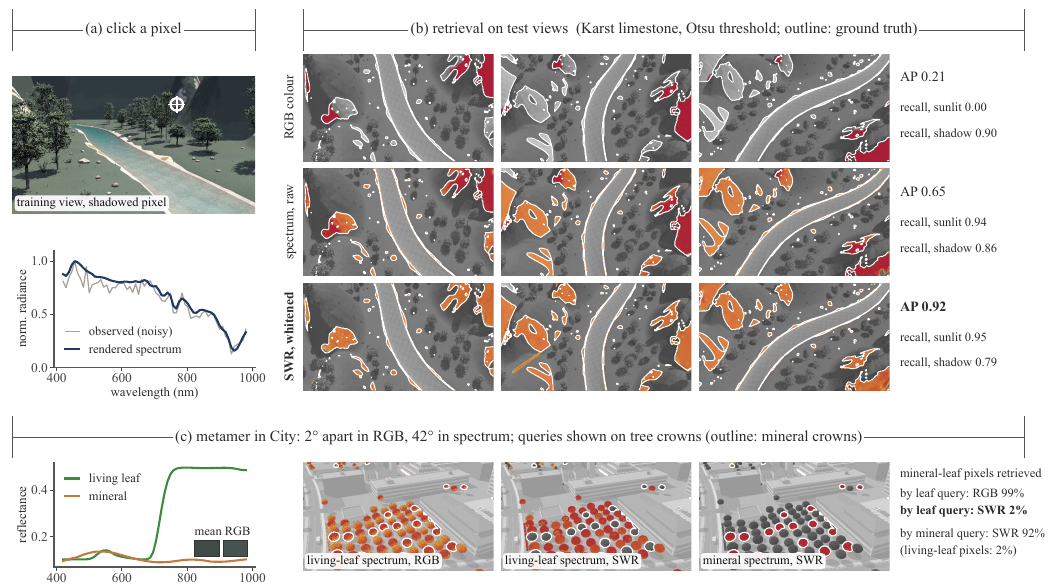}
\vspace{-6mm}
\caption{\textbf{Material retrieval from clicked and library spectra.} (a) A rendered spectrum clicked on shadowed limestone in a Karst training view defines the query. (b) Test-view selections use Otsu thresholds on RGB, raw spectral and whitened spectral angles; white outlines mark ground truth. Selected pixels are coloured from yellow (angle near the threshold) through orange to red (closest match); unselected pixels are grey. AP and recall cover all test views. Whitening retrieves sunlit rock missed by RGB. (c) In the City metamer variant, spectral queries distinguish mineral crowns (outlined) from living foliage, with the same selection colours as in (b); RGB confuses them.}
\label{fig:supp_fpq}
\end{figure*}

\noindent\textbf{Illumination-aware spectral matching.} The query is a spectrum: the rendered spectrum at a clicked pixel or a library reflectance $\boldsymbol\rho$. Each Gaussian carries its visibility-weighted mean spectrum over training rays with non-zero visibility, $\bar{\mathbf h}_i=\sum_{\mathbf r}w_i(\mathbf r)\mathbf h_i(\mathbf r)/\sum_{\mathbf r}w_i(\mathbf r)$. Let $\boldsymbol\Phi_{\mathrm d},\boldsymbol\Phi_{\mathrm s}\in\mathbb R^L$ denote the direct and sky irradiance of the world integrated over the band responses; both are known for \bench, and the cone and the whitening below use them. Under a two-component sun/sky approximation, surface radiance is modelled as proportional to $\boldsymbol\rho\odot(\xi\boldsymbol\Phi_{\mathrm d}+\boldsymbol\Phi_{\mathrm s})$, where $\odot$ denotes bandwise multiplication and $\xi\in[0,1]$ the direct-shading coefficient. We reduce this modelled variation using two comparisons:
\begin{itemize}\setlength{\itemsep}{0pt}
\item a library spectrum is compared by its angle to the non-negative cone spanned by $\boldsymbol\rho\odot\boldsymbol\Phi_{\mathrm d}$ and $\boldsymbol\rho\odot\boldsymbol\Phi_{\mathrm s}$, which spans the illumination mixtures in this model;
\item a clicked spectrum is compared after illumination whitening: the root-mean-square (RMS) log-ratio of the two spectra once the constant vector and the two principal directions of $\log(\xi\boldsymbol\Phi_{\mathrm d}+\boldsymbol\Phi_{\mathrm s})$, $\xi\in[0,1]$, are projected out.
\end{itemize}
The constant direction removes overall intensity; the two principal directions approximate sun/sky colour variation. Both use the known irradiance and assume this transport model; unknown lighting is not estimated. Visibility weights render the per-Gaussian angles, and Otsu's method~\citep{otsu1979threshold} estimates the selection threshold from the log-angle histogram.

\noindent\textbf{Evaluation.} Market, City, Botanical and Karst use the models of Table~2. Clicks are three sunlit and three shadowed pixels inside a single class in training views. Library queries use the assigned specimen of each class, another specimen of the class and ECOSTRESS spectra~\citep{ecostress} that do not occur in the world. Each query is scored by AP against the class mask on all test views, and the four worlds are weighted equally. With Otsu thresholds, a shadowed click retrieves a larger part of the sunlit region of its class after whitening than with raw spectra, and RGB retrieves little of it. Click queries operate on rendered pixels; per-Gaussian queries give nearly the same AP.

\noindent\textbf{Approximate RGB metamerism.} The City mineral substitute is close to foliage in RGB but differs strongly in spectrum; the two angles are reported in Figure~\ref{fig:supp_fpq}(c). This is an approximate camera-colour match rather than an exact CIE metameric match. A living-leaf spectrum separates the two leaf types and largely excludes mineral crowns, whereas RGB ranks mineral crowns as more leaf-like than living foliage, as Figure~\ref{fig:supp_fpq}(c) shows. Both targets and distractors are tree crowns, reducing object-category cues in this test.

\noindent\textbf{From spectra to a 3D selection.} Training-view aggregation gives each Gaussian its mean spectrum $\bar{\mathbf h}_i$. A query scores these spectra once, and the resulting scores remain attached to the same primitives as the camera moves. Visibility weights then render the selection from each new viewpoint. This reuses a scene-level spectral comparison across cameras instead of per-image classification.

\noindent\textbf{Altered leaves.} In the City variants, \method\ separates mineral and dry leaves from healthy foliage more clearly than RGB $k$-means clustering, measured by the total variation between their label distributions (Figure~9(c) of the main paper).

\begin{figure*}[t]
\centering
\includegraphics[width=\textwidth]{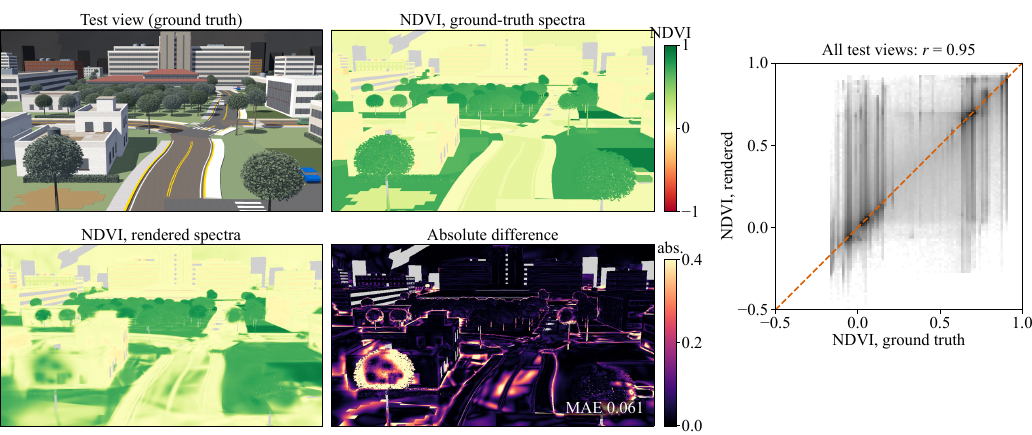}
\vspace{-6mm}
\caption{\textbf{Simulated multispectral sensor on City.} NDVI from MicaSense bands synthesized from ground-truth and rendered spectra of a test view; right, joint density over all test views.}
\label{fig:app_sensor}
\end{figure*}

\begin{figure*}[t]
\centering
\includegraphics[width=\textwidth]{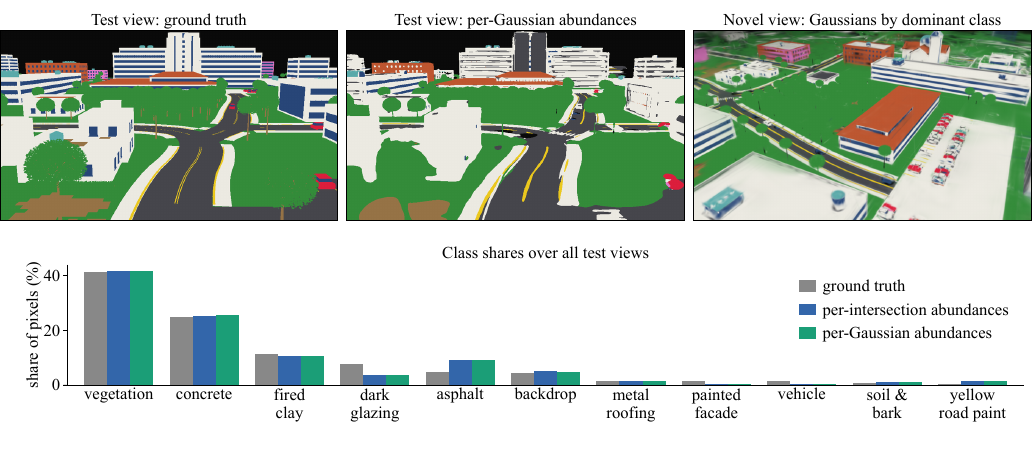}
\vspace{-6mm}
\caption{\textbf{Material statistics from per-Gaussian abundances on City.} Dominant classes in a test view and on the Gaussians in a novel view; bottom, class shares over all test views.}
\label{fig:app_stats}
\end{figure*}

\subsection{Simulated Multispectral Sensors}\label{sec:app_sensor}
\noindent Each band of a target sensor, here a MicaSense RedEdge-MX and Sentinel-2, is fitted as a non-negative combination of the 64 band responses of \bench, applied to rendered spectra and divided by the known irradiance. NDVI, the normalized difference of near-infrared and red reflectance~\citep{tucker1979ndvi}, computed from the rendered spectra closely follows the ground truth (Figure~\ref{fig:app_sensor}), and NDRE, which replaces the red band with a band on the red edge of vegetation reflectance~\citep{horler1983rededge}, behaves similarly.

\noindent\textbf{Insight.} The rendered spectra are physically meaningful beyond the bands used for training: a sensor that was never observed can be simulated from any camera by integration alone. Deviations concentrate on rare classes and material boundaries, where the abundances themselves are least reliable, so index accuracy follows identification accuracy.

\subsection{3D Material Statistics}\label{sec:app_stats}
\noindent As for the per-Gaussian mean spectrum of Sec.~\ref{sec:supp_fpq}, each Gaussian receives one abundance vector, the visibility-weighted mean of its per-intersection abundances over all training rays, which the endmember-to-class map of the main evaluation turns into class shares. Composited in the test views, these shares remain close to the ground-truth composition for the major classes, and material identification is nearly unchanged relative to per-intersection decoding (Figure~\ref{fig:app_stats}).

\noindent\textbf{Insight.} Because abundances are attached to explicit primitives, composition can be aggregated, selected or exported in 3D and remains consistent across views. A single vector per Gaussian loses little, indicating that the intensity $s_i$ absorbs most shading variation and that abundances vary little within a primitive. Classes that the model confuses, such as dark glazing and asphalt, are misestimated in the same way as in the full render, so the statistics inherit identification errors rather than adding new ones.

\begin{figure*}[t]
\centering
\includegraphics[width=\textwidth]{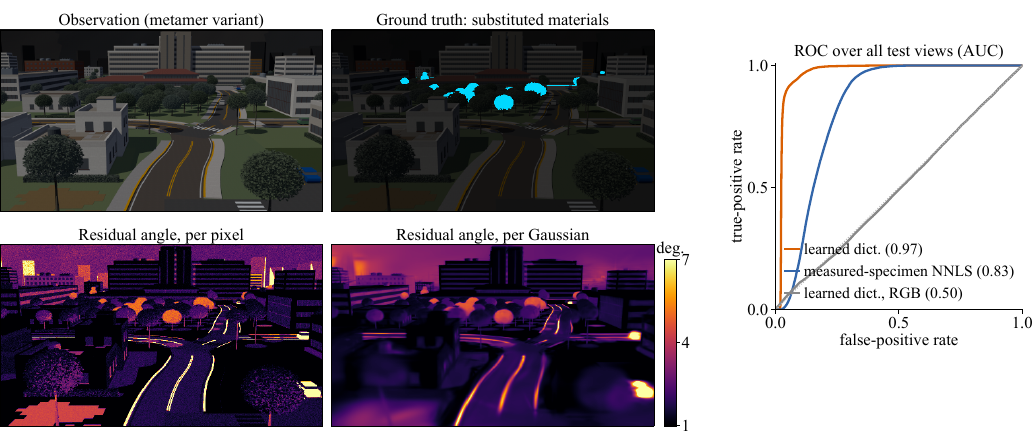}
\vspace{-6mm}
\caption{\textbf{Detection of spectra unexplained by the dictionary.} Residual of a fit to the learned endmembers on a test view of the City metamer variant, per pixel and per Gaussian, with the substituted materials in cyan; right, ROC curves over all test views for the learned dictionary, measured-specimen NNLS and the learned dictionary projected to RGB.}
\label{fig:app_anomaly}
\end{figure*}

\begin{figure*}[t]
\centering
\includegraphics[width=\textwidth]{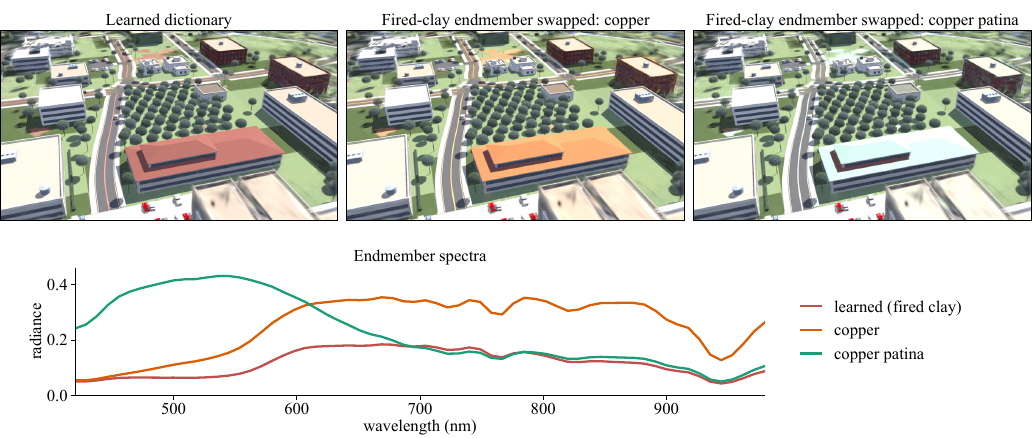}
\vspace{-6mm}
\caption{\textbf{Material editing by swapping one endmember.} A novel view of City rendered with the learned dictionary and with the fired-clay endmember replaced by ECOSTRESS copper or copper patina; abundances and intensities are unchanged; bottom, endmember spectra.}
\label{fig:app_edit}
\end{figure*}

\subsection{Spectra Unexplained by the Dictionary}\label{sec:app_anomaly}
\noindent In the metamer variant of City (Sec.~\ref{sec:supp_worlds}), some tree crowns and car finishes are replaced by materials of matching colour but different spectrum. Each observed spectrum is scored by the angle to its non-negative fit from the dictionary learned on the unaltered world, and the score can be averaged on the Gaussians over training views. This score separates the substituted materials clearly from the original ones, whereas the same fit in RGB does not separate them (Figure~\ref{fig:app_anomaly}).

\noindent\textbf{Insight.} The shared dictionary acts as a compact spectral model of the scene, so what it cannot explain marks materials absent during training. In RGB, a few non-negative endmembers already reproduce almost any colour, so metamers leave no residual. The score measures distance from the dictionary rather than novelty: dry leaves, which remain close to vegetation, are harder to detect, and materials that the dictionary represents poorly also score high.

\subsection{Material Editing}\label{sec:app_edit}
\noindent The endmember carrying the fired-clay roofs is multiplied bandwise by the ratio of a new library reflectance to the reflectance it currently represents, with abundances and intensities unchanged. For both copper and copper patina~\citep{ecostress}, the edited render remains close to a ground-truth edited world rendered with the forward model of \bench, with a larger deviation for the patina, whose spectrum departs further from the original tile (Figure~\ref{fig:app_edit}).

\noindent\textbf{Insight.} Since spectra are factored into abundances and endmembers, editing one endmember changes a material in every view while geometry, shading and mixing stay consistent. The edit follows the learned assignment: where an endmember also contributes to other classes, part of the change leaks to them, so editing precision depends on how cleanly the dictionary separates materials.

\section{Additional Visualizations}\label{sec:supp_gallery}
\noindent\textbf{Cross-view evaluation.} Figures~\ref{fig:swr_hsi_1} and~\ref{fig:swr_hsi_2} illustrate the hyperspectral observations. Figures~\ref{fig:swr_multiview_1}--\ref{fig:swr_multiview_4} compare ground truth and SWR predictions at three test cameras per world, using the main-paper models and the endmember-to-class maps fixed on training views. Camera separation from a reference view determines the examples independently of prediction quality.

\begin{figure*}[tp]
\centering
\includegraphics[width=\textwidth]{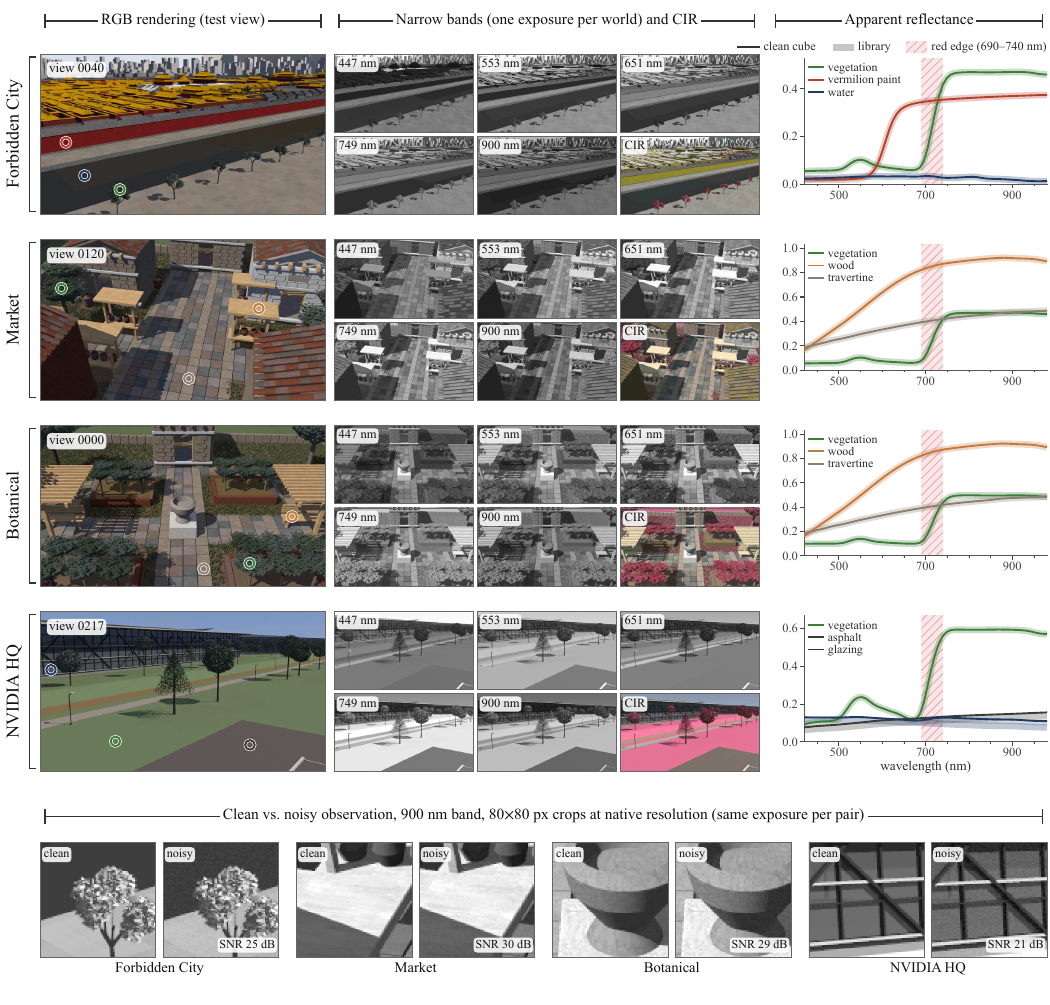}
\caption{\textbf{Simulated hyperspectral observations: Forbidden City, Market, Botanical and NVIDIA HQ.} Narrow bands share one exposure per world, so brightness differences between bands follow the radiance; the colour-infrared composite (R, G, B = 847, 651, 553\,nm) renders vegetation red. Apparent reflectance divides the clean radiance by the white-reference radiance under each pixel's illumination (median and 10--90\,\% range, over the library reflectance of the assigned specimen, drawn as a translucent grey band; hatched pink: red edge, 690--740\,nm). The apparent reflectance reproduces the library reflectance, including chlorophyll absorption near 670\,nm and the red edge. Bottom: clean and noisy crops of the 900\,nm band with the signal-to-noise ratio of the noisy crop.}
\label{fig:swr_hsi_1}
\end{figure*}

\begin{figure*}[tp]
\centering
\includegraphics[width=\textwidth]{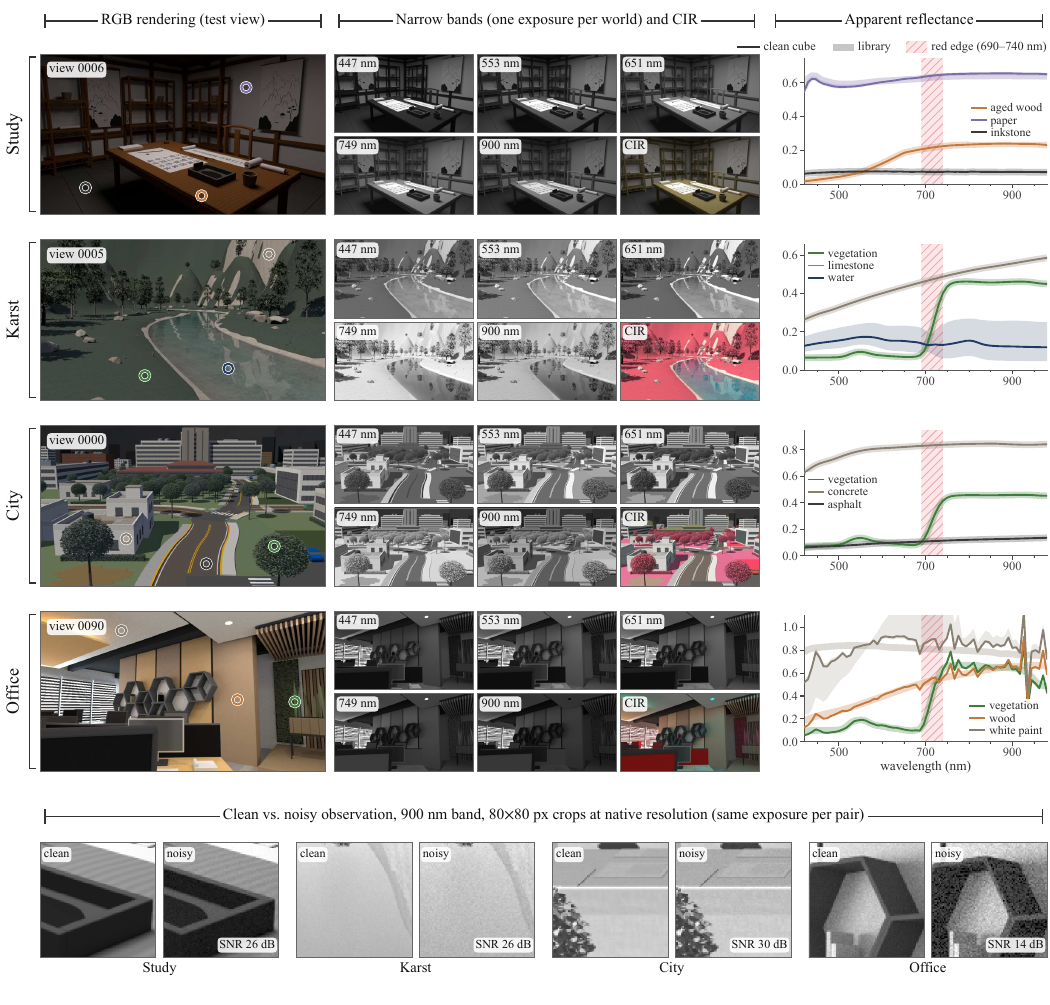}
\caption{\textbf{Simulated hyperspectral observations: Study, Karst, City and Office.} Layout as in Figure~\ref{fig:swr_hsi_1}. Indoor worlds are lit by white sources, so their bands differ less; the Office lights emit no near-infrared, which comes only from daylight through the blinds. Path-traced interiors deviate through interreflection and residual Monte Carlo variance.}
\label{fig:swr_hsi_2}
\end{figure*}

\begin{figure*}[tp]
\centering
\includegraphics[width=\textwidth,height=0.78\textheight,keepaspectratio]{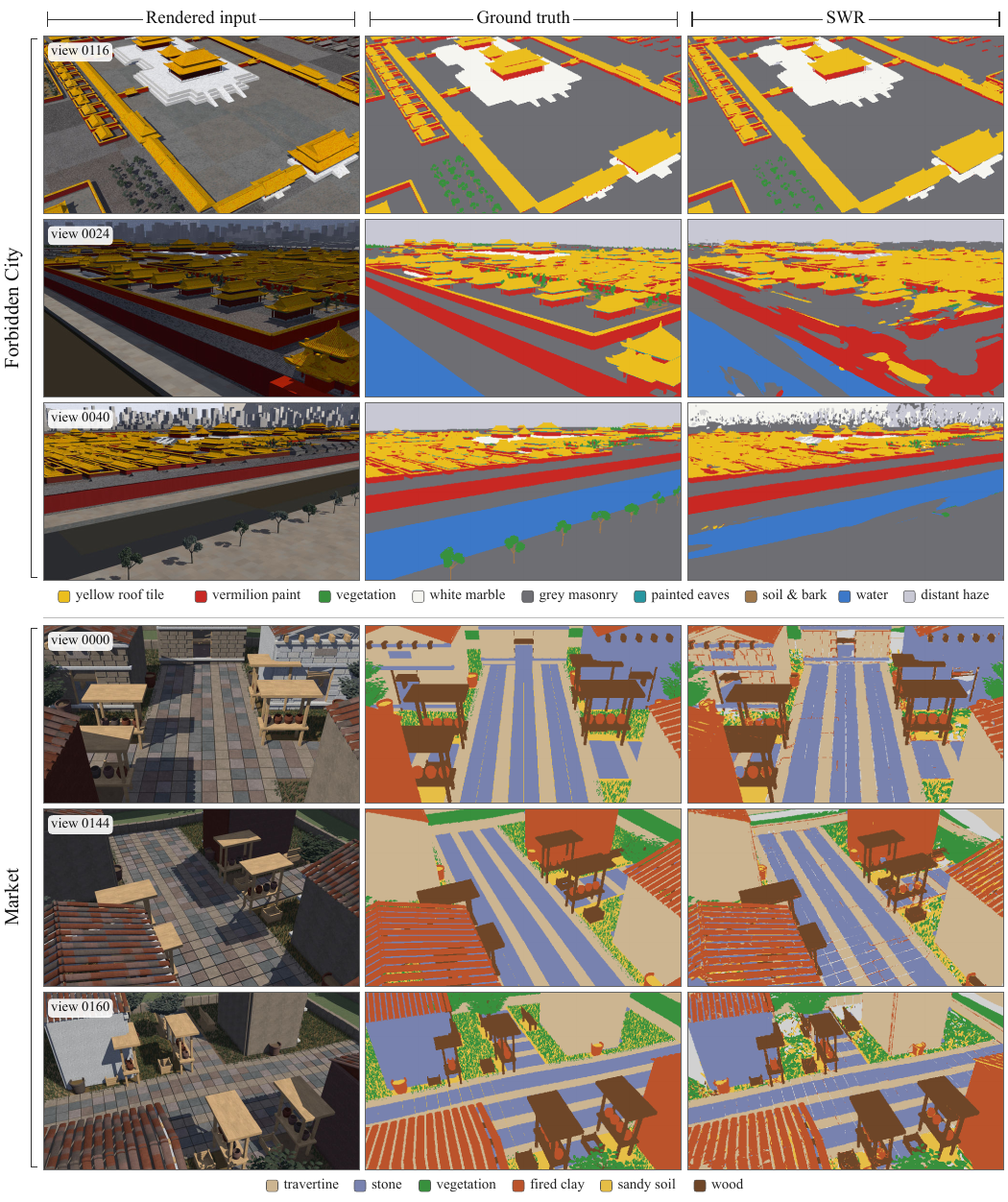}
\caption{\textbf{Additional test views: Forbidden City and Market.} Three separated test cameras per world, shown as full frames.}
\label{fig:swr_multiview_1}
\end{figure*}

\begin{figure*}[tp]
\centering
\includegraphics[width=\textwidth,height=0.84\textheight,keepaspectratio]{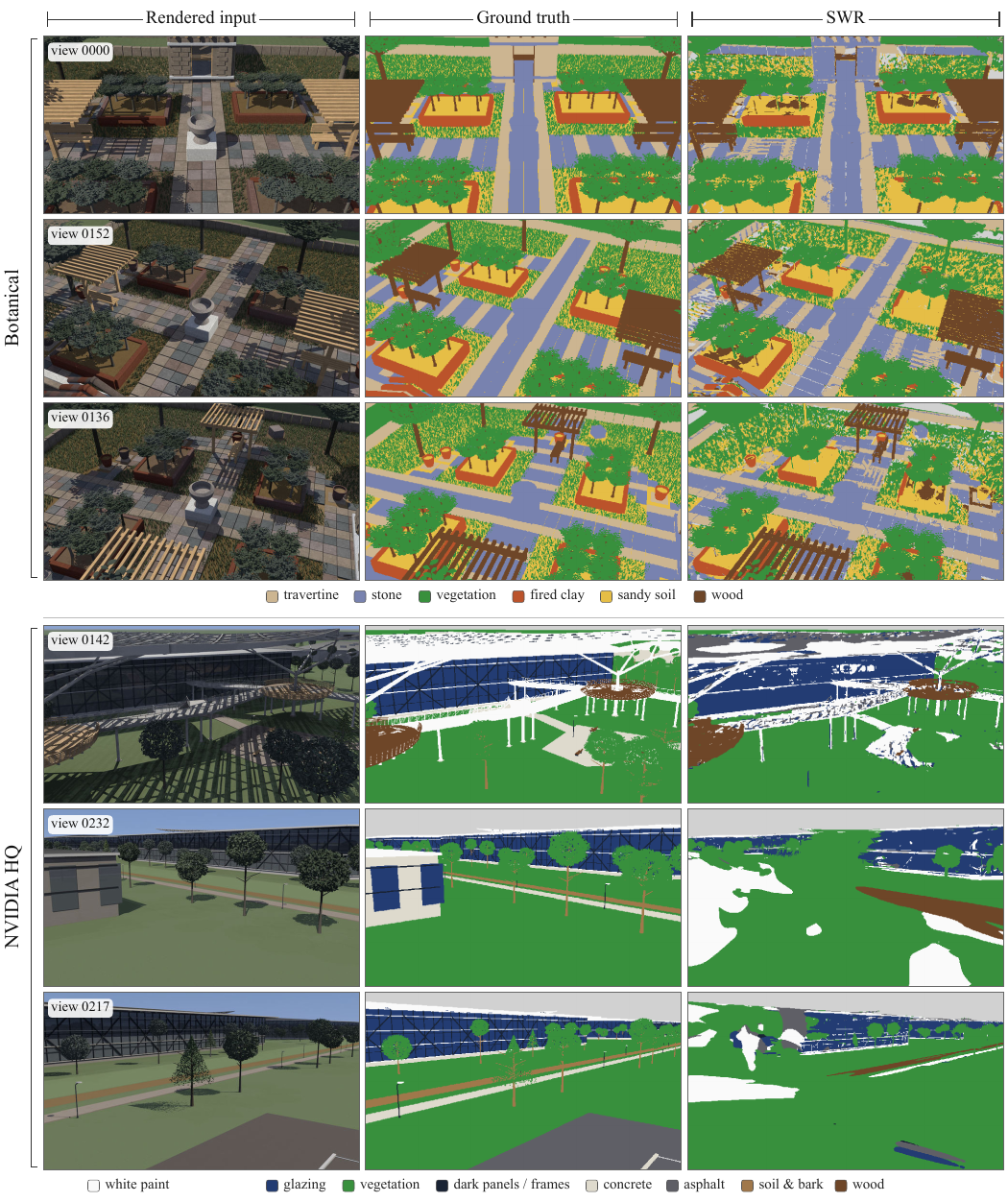}
\caption{\textbf{Additional test views: Botanical and NVIDIA HQ.} Three separated test cameras per world, shown as full frames.}
\label{fig:swr_multiview_2}
\end{figure*}

\begin{figure*}[tp]
\centering
\includegraphics[width=\textwidth,height=0.84\textheight,keepaspectratio]{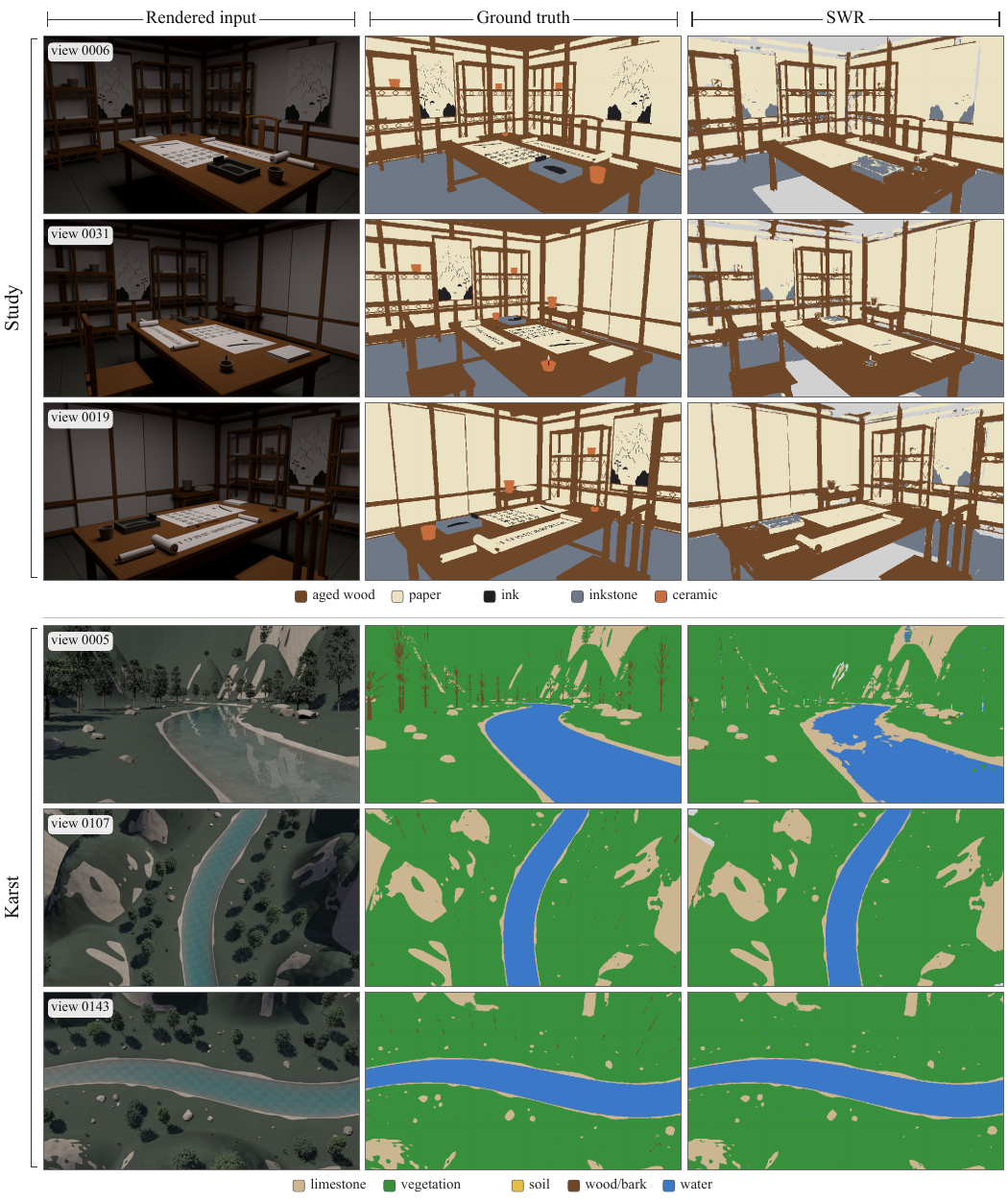}
\caption{\textbf{Additional test views: Study and Karst.} Three separated test cameras per world, shown as full frames.}
\label{fig:swr_multiview_3}
\end{figure*}

\begin{figure*}[tp]
\centering
\includegraphics[width=\textwidth,height=0.84\textheight,keepaspectratio]{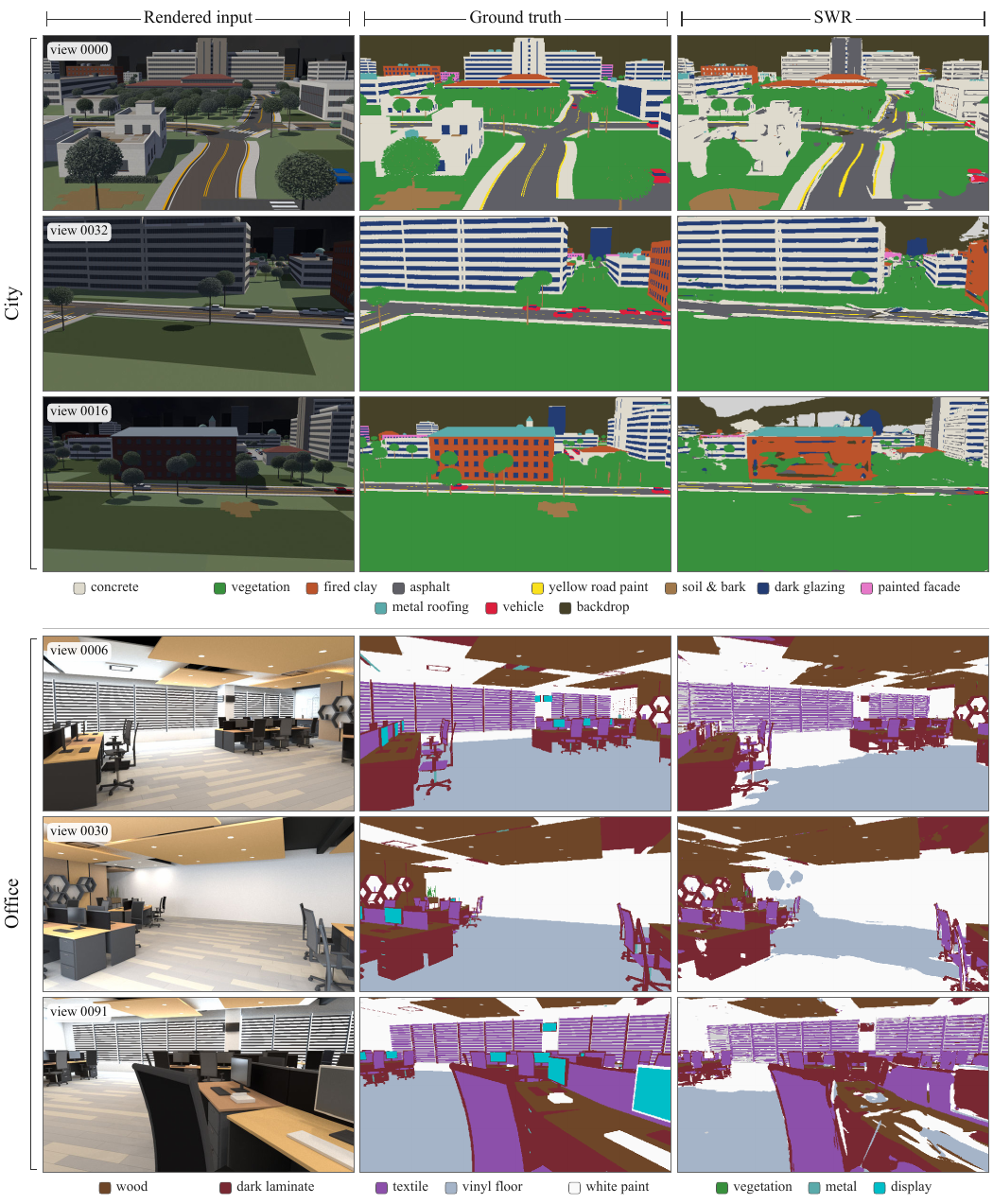}
\caption{\textbf{Additional test views: City and Office.} Three separated test cameras per world, shown as full frames.}
\label{fig:swr_multiview_4}
\end{figure*}

\twocolumn

\fi
\end{document}